\documentclass[journal]{IEEEtran}
\usepackage{amsmath}
\usepackage{algorithm}
\usepackage{multirow}
\usepackage{graphicx}
\usepackage{subcaption}
\newtheorem{definition}{Definition}

\ifCLASSINFOpdf
\else
\fi
\usepackage{etoolbox}
\makeatletter
\patchcmd{\@makecaption}
  {\scshape}
  {}
  {}
  {}
\makeatother
\usepackage{algorithm,float}
\usepackage{amssymb}
\usepackage{algorithmic}
\usepackage{lipsum}
\usepackage{amsmath}
\usepackage{multirow}
\usepackage{graphicx}
\usepackage{subcaption}

\usepackage{algorithm}
\usepackage{amsmath}
\renewcommand{\algorithmicrequire}{\textbf{Input:}}  % Use Input in the format of Algorithm
\renewcommand{\algorithmicensure}{\textbf{Output:}} % Use Output in the format of Algorithm
\usepackage{lineno,hyperref,multirow,graphicx,enumerate,color,amsmath, epstopdf}
\usepackage{booktabs}
\usepackage[justification=centering]{caption}
\usepackage[noadjust]{cite}
\usepackage{stfloats}
\usepackage[utf8]{inputenc}
\usepackage[greek,russian,spanish,english]{babel}
\usepackage[OT2,T1, T2A,OT1]{fontenc}
\usepackage{stfloats}
\usepackage{hyperref}
\begin{document}

\title{GBCT: An Efficient and Adaptive Granular-Ball Clustering Algorithm for Complex Data}

\author{Shuyin~Xia,
	    Bolun~Shi,
	    Yifan~Wang,
        Jiang~Xie$^{*}$,
        Guoyin~Wang$^{*}$,
        Xinbo~Gao
        
\thanks{Shuyin Xia, Jiang Xie, Guoyin Wang and Xinbo Gao are with the Chongqing Key Laboratory of Computational Intelligence, the Key Laboratory of Cyberspace Big Data Intelligent Security, Ministry of Education, and the Key Laboratory of Big Data Intelligent Computing, Chongqing University of Posts and Telecommunications, 400065, Chongqing, China. E-mail: xiasy@cqupt.edu.cn, xiejiang@cqupt.edu.cn (corresponding author), wanggy@cqupt.edu.cn (corresponding author), gaoxb@cqupt.edu.cn.}}

\maketitle

\begin{abstract}
Traditional clustering algorithms often focus on the most fine-grained information and achieve clustering by calculating the distance between each pair of data points or implementing other calculations based on points. This way is not inconsistent with the cognitive mechanism of~\lq\lq~global precedence~\rq\rq~in human brain, resulting in those methods\rq~bad performance in efficiency, generalization ability and robustness. To address this problem, we propose a new clustering algorithm called granular-ball clustering (GBCT) via granular-ball computing. 
Firstly, GBCT generates a smaller number of granular-balls to represent the original data, and forms clusters according to the relationship between granular-balls, instead of the traditional point relationship. At the same time, its coarse-grained characteristics are not susceptible to noise, and the algorithm is efficient and robust;
besides, as granular-balls can fit various complex data, GBCT performs much better in non-spherical data sets than other traditional clustering methods. The completely new coarse granularity representation method of GBCT and cluster formation mode can also used to improve other traditional methods. All codes can be available from the following  \href{https://github.com/wylbdthxbw/GBC}{link}.
	
%Traditional clustering algorithms often focus on the finest-grained information, achieving clustering by computing the distance between every pair of data points. However, this approach significantly increases computational costs, thereby compromising algorithm performance and generalization capabilities. Inspired by the efficient adaptive differentiation process of granular-ball computation, we propose a novel fundamental clustering algorithm. This algorithm replaces the computation of distances between a large number of data points with the calculation of distances between a fewer number of granular-balls, enabling accurate and efficient identification of complex datasets. Experimental results demonstrate that, compared to widely used clustering algorithms, our approach not only exhibits superior clustering performance but also achieves significant improvements in computational speed.
\end{abstract}


\IEEEpeerreviewmaketitle

\section{Introduction}

\IEEEPARstart{C}{luster} 
analysis is an unsupervised data analysis method used to comprehend complex unlabeled data by partitioning objects into different clusters based on their similarities, aiming to ensure high similarity within clusters and high dissimilarity between different clusters \cite{1,2}. This enables easy analysis of data characteristics. Cluster analysis finds widespread applications in various domains such as image segmentation\cite{3}, information retrieval \cite{4}, data compression \cite{5}, and bioinformatics \cite{6}.

Present clustering algorithms are broadly categorized into fundamental and enhanced algorithms. Fundamental clustering methods include partition-based clustering \cite{7}, hierarchical clustering \cite{8}, density-based clustering \cite{9}, and graph-based clustering \cite{10}. Among them, the K-Means algorithm \cite{11} is a widely used partition-based method, particularly suitable for datasets with convex shapes but ineffective for non-spherical cluster structures. Density-based clustering algorithms like DBSCAN \cite{12} and DP \cite{13} are effective in handling irregular or curved clusters with good noise resistance, but their clustering performance deteriorates in the presence of clusters with varying densities and they are sensitive to parameter selection \cite{14,15}.

Spectral clustering \cite{16,17} is a method based on graph theory. Its basic idea is to divide weighted undirected graphs into two or more optimal subgraphs, so that the internal subgraphs are as similar as possible, and the distance between subgraphs is as far as possible, so as to achieve the purpose of common clustering. It is suitable for processing data sets with complex structures, but the calculation cost is high. In addition, graph learning method \cite{18,19} is a model-based clustering method, which captures the internal structure of data through steps such as constructing similarity graph and calculating weight matrix, and then optimizes the algorithm in graph theory to obtain clustering results. However, these operations require high computational complexity.

The aforementioned analysis indicates insufficient generalization ability in existing fundamental algorithms, while enhanced algorithms aim to enhance the performance of fundamental ones. When dealing with nonlinear separable data, most enhanced algorithms adopt two main approaches: using non-Euclidean distance metrics or employing kernel tricks to map the original data into high-dimensional spaces. However, these operations often incur additional costs such as increased computational complexity or decreased accuracy. The performance of fundamental algorithms directly affects the effectiveness of enhanced algorithms. Therefore, proposing a novel fundamental clustering algorithm with good generalization ability is crucial.

Thus, a core question arises: How to achieve adaptive and efficient clustering in the face of increasingly abundant and complex data distributions? Based on this, we have developed a fundamental clustering algorithm with good learning ability for complex nonlinear data. 
The main advantage of the proposed algorithm is that the number of granular-balls is far less than the original sample points to achieve efficient and flexible representation of the sample space. At the same time, the coarse-grained characteristics of granular-balls effectively resist the interference of fine-grained noise, and the algorithm is robust. Since this algorithm is a fundamental clustering approach, we mainly compare it with classical clustering algorithms to verify its superiority in fundamental algorithms. The main contributions are as follows:

\setlength{\hangindent}{2em}
1) Inspired by the human brain's~\lq\lq global precedence\rq\rq~cognitive mechanism of the human brain and the granular-ball splitting mechanism, we propose a novel fundamental clustering algorithm called Granular-Ball Clustering (GBCT), which represents the original data by splitting granular-balls from coarse to fine adaptively to achieve clustering results.

\hangafter=1
\setlength{\hangindent}{2em}
2) Our method covers and represents unlabeled data using granular-balls of different granular sizes, demonstrating good data fitting and representation capabilities. Experimental results on multiple datasets show that GBCT achieves better clustering performance.

\hangafter=1
\setlength{\hangindent}{2em}
3) By using granular-ball distances instead of sample distances and having fewer granular-balls than sample points, computational costs are reduced. Experimental results also demonstrate the efficiency of GBCT.

\hangafter=1
\setlength{\hangindent}{2em}
4) Our method does not consider the influence of isolated granular-balls during the clustering process, effectively smoothing out noise points, thus exhibiting robustness.

The remainder of this paper is organized as follows: In \autoref{s2}, we review the research progress of current clustering algorithms and the development of granular computing. In \autoref{s3}, we provide the fundamental definition and process of our proposed algorithm, GBCT, highlighting some inherent challenges. In \autoref{s4}, We verify the effectiveness, efficiency and robustness of the GBCT algorithm by comparing it with other algorithms on synthetic and artificial data sets. Finally, in \autoref{s5}, we summarize the strengths and weaknesses of the GBCT algorithm and outline potential directions for future research.

\section{Related Work}\label{s2}
\subsection{Related Clustering Work}
Clustering algorithms are broadly categorized into foundational and advanced methods. Within foundational algorithms, partition-based K-Means\cite{20}, known for its simplicity and efficiency, iteratively optimizes cluster centers to partition data. However, its sensitivity to initial values has led to enhancements like Tzortzis and Likas's MinMax k-means, which utilizes a weighted approach to improve initial center quality\cite{21}. Feiping Nie et al. \cite{22}developed a K-Multiple Means (KMM) method to address the limitation of K-Means in modeling each category with a single center by grouping data points with multiple sub-cluster means into k specified clusters. Density-based algorithms like DBSCAN \cite{12} and DP \cite{13}  gain attention for identifying clusters of varied shapes and noise robustness, yet they are parameter-sensitive, particularly in variable density clusters. Daszykowski et al. \cite{23}enhanced DBSCAN by estimating $\epsilon$ values using the minpts parameter, simplifying parameterization. For DP, efficiency advancements led to distributed versions like LSH-DDP\cite{24}, partitioning data with Locality-Sensitive Hashing and aggregating local information under MapReduce for approximate results. 
The FastDPeak algorithm proposed by Cheng et al.\cite{25} achieves the purpose of acceleration by introducing K-nearest neighbor to reduce the calculation amount of density and $c$-distance. GB-DP\cite{26} improves the efficiency of unsupervised learning by integrating multi-granularity ideas into DP algorithm.

Model-based clustering methods often combine a variety of methods, and the design is relatively complex. Graph learning method \cite{18,19}, as a major branch, can capture the complex structure of data under nonlinear or noisy conditions. For example, Wang et al.\cite{27}Using the generative adversarial network framework for graph representation learning provides a new idea and method for graph representation learning.

Hierarchical clustering is also a commonly used clustering algorithm. It starts by treating each sample in the dataset as a separate cluster, and gradually merges these clusters together to form larger clusters, until all samples are aggregated into a larger cluster or reach user-defined termination conditions. The key to cohesive clustering is how to define the~\lq\lq distance\rq\rq~or~\lq\lq similarity\rq\rq~between clusters. Common methods include single link (nearest neighbor), full link (farthest neighbor), average link, and Ward method. Agglomerative clustering is the most common form of hierarchical clustering. An Efficient Hierarchical Clustering Algorithm for Large Datasets\cite{45} and Fast and Accurate Hierarchical Clustering Based
on Growing Multiplayer Topology Training\cite{46} are proposed to solve large-scale clustering problem, but it still cannot recognize non spherical datasets. Overall similar with other clustering methods, hierarchical clustering is also designed based on a point instead of coarse granularity because it treated each data point as a cluster in the beginning of cluster former process. In this paper, motivated by granular-ball computing, we want to propose a completely new fundamental clustering algorithm in which clusters are formed based on a coarse granularity instead of a point. Consequently, in comparison with the traditional clustering algorithms, the proposed method can exhibit better performance in efficiency, robustness and the ability of solving non spherical data. 

\subsection{Granular-ball Computing}
\begin{figure}[h]
	\centering
	\includegraphics[width=2in]{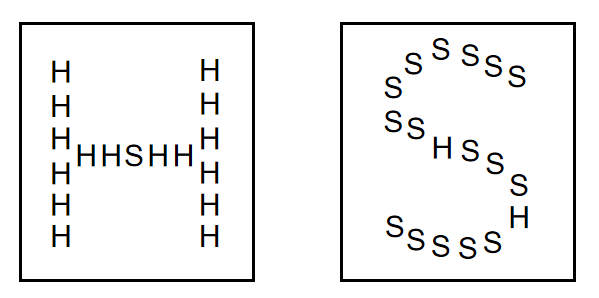}
	\caption{The Cognitive Mechanism of ~\lq\lq global precedence\rq\rq~ in Human Brain}\label{t1}
	%\Description{A woman and a girl in white dresses sit in an open car.}
\end{figure}

Chen (1982), in a study published in Sciences, demonstrated that human cognition exhibits a ~\lq\lq global precedence\rq\rq~ characteristic\cite{28}. This means that when people observe objects, they first notice the larger-scale features before attending to finer details. For example, in \autoref{t1}, individuals initially perceive the large letters ~\lq\lq H\rq\rq~ and ~\lq\lq S\rq\rq~ before noticing the smaller letters ~\lq\lq S\rq\rq~ and ~\lq\lq H\rq\rq~ contained within them. This cognitive process is adaptive. The human brain's cognitive mechanism places greater emphasis on global topological properties rather than directly processing the finest-grained raw data. Therefore, the human visual system tends to observe objects from a coarse granularity to a finer one, aligning with the principles of human cognition.
Based on the "global precedence" cognitive mechanism of the human brain, Wang \cite{29} initially introduced multi-granularity cognitive computation, and subsequently, Xia \cite{30}further proposed Granular-ball computing. This method initially treats the entire dataset as one granular-ball and continuously divides it until each granular-ball meets certain quality threshold conditions, ensuring clear decision boundaries. Granular-ball computing employs granular-balls of varying sizes to cover the sample space and conducts subsequent computations based on these granular-balls. In tasks such as classification, purity serves as a critical criterion for assessing granular-ball quality. Xia \cite{31} chose granular-ball as the foundation for multi-granularity feature representation because granular-balls offer a concise and standardized mathematical model representation. As shown in \textit{Definition} \autoref{d1} .

\begin{definition}\label{d1}
	Given a dataset \(D = \{P_i | i=1,2,3,...,n\} \in \mathbb{R}^d\), where \(n\) and \(d\) respectively represent the number of data points and dimensions, granular-balls can be abbreviated as \(GB\). The \(GB_j\) (for \(j=1,2,...,m\)) are used to cover and represent the dataset \(D\) with the goal of minimizing information loss. Here, \(|GB_j|\) denotes the number of objects contained within the \(j^{th}\) granular-ball. Consequently, the mathematical model of granular-ball computing can be succinctly described as follows:
	\begin{equation}\label{1}
		\begin{aligned}
			&~~~~~~{f(x,\vec\alpha)\longrightarrow g(GB,\vec\beta)},\\  %f(x)\to g(GB_j)
			&s.t.~~~~min~~~~    \frac{n}{\sum_{j=1}^{m}{|{GB}_j|}}+m,\\
			\quad~~~~~&~~~~~~~~quality{(GB}_j) \ge \mu.
		\end{aligned}
	\end{equation}
\end{definition}

In the formula mentioned above, $f(x,\vec{\alpha})$ represents the existing learning models with point $x$ as input, where $\vec{\alpha}$ denotes the model parameters. In the objective of the granular-ball computing model, $g(GB,\vec{\beta})$ signifies the granular-ball computing learning models with $GB$ as input, and $\vec{\beta}$ are the model parameters. When the input of the learning model shifts from the traditional point $x$ to the $GB$, the computational model transitions from the traditional $f$ to $g$. As the number of generated $GB$ is often significantly smaller than the number of sample points $x$, granular-ball computing proves to be more efficient.  

Notably, the initial term $\frac{n}{\sum_{j=1}^{m}{|{GB}_j|}}$ signifies the reciprocal of the sample coverage rate, where a higher coverage rate is preferable to minimize information loss. The term $m$ represents the number of granular-balls, emphasizing that a reduced number of granular-balls under the constraints leads to more efficient and robust computation. The constraint $quality{(GB}_j) \ge \mu$ introduces an adaptive rule, where appropriate quality metrics are developed to evaluate the quality of granular-balls and $\mu$ represents the threshold mass of the granular-ball. This is predicated on the notion that, for an enhanced representation of the original data, the quality of granular-balls should remain within a reasonable range, ensuring each granular-ball is compact and effective for clustering tasks. In granular-ball computing, the splitting process of ~\lq\lq global precedence\rq\rq~ is a standard heuristic method to optimize the constraints. The fundamental concept of this method involves initially treating the entire dataset as a granular-ball, which is subsequently split into smaller granular-balls. When the qualities of these granular-balls fail to meet the given quality threshold $T$, it becomes imperative to further split them into smaller granular-balls until the quality of each granular-ball meets the threshold $\mu$.

Although granular-ball computing theory has not been around for long, it has already given rise to various granular-ball generation methods and models in multiple fields, including granular-ball neighborhood rough sets\cite{32}, classifiers\cite{33}, spectral clustering\cite{34}, sampling\cite{35}, rough sets\cite{36}, neural networks\cite{37}, evolutionary computation\cite{38}, support vector machines\cite{39}, etc.  Compared to traditional fine-grained input approaches, granular-ball computing offers advantages in terms of efficiency, robustness, and interpretability.

\section{Main Methods}\label{s3}
In this section, the mathematical model representation of the GBCT method is first given, and the whole algorithm process is divided into two orders: granular-ball splitting and cluster formation. Then, in the splitting phase, we theoretically analyzed and proved the limitations of the weighted splitting method of sub-spheres adopted in $\rm{GBCT\_old}$ \cite{40} (the complete process is shown in the Supplementary\_Material), and indirectly verified the importance of the central consistency measure considering the internal distribution of granular-balls through this analysis. Finally, the time complexity of the algorithm was analyzed in detail.

\subsection{GBCT Methods}

\begin{figure}[H]
	\vspace{0em}
	\includegraphics[width=0.48\textwidth]{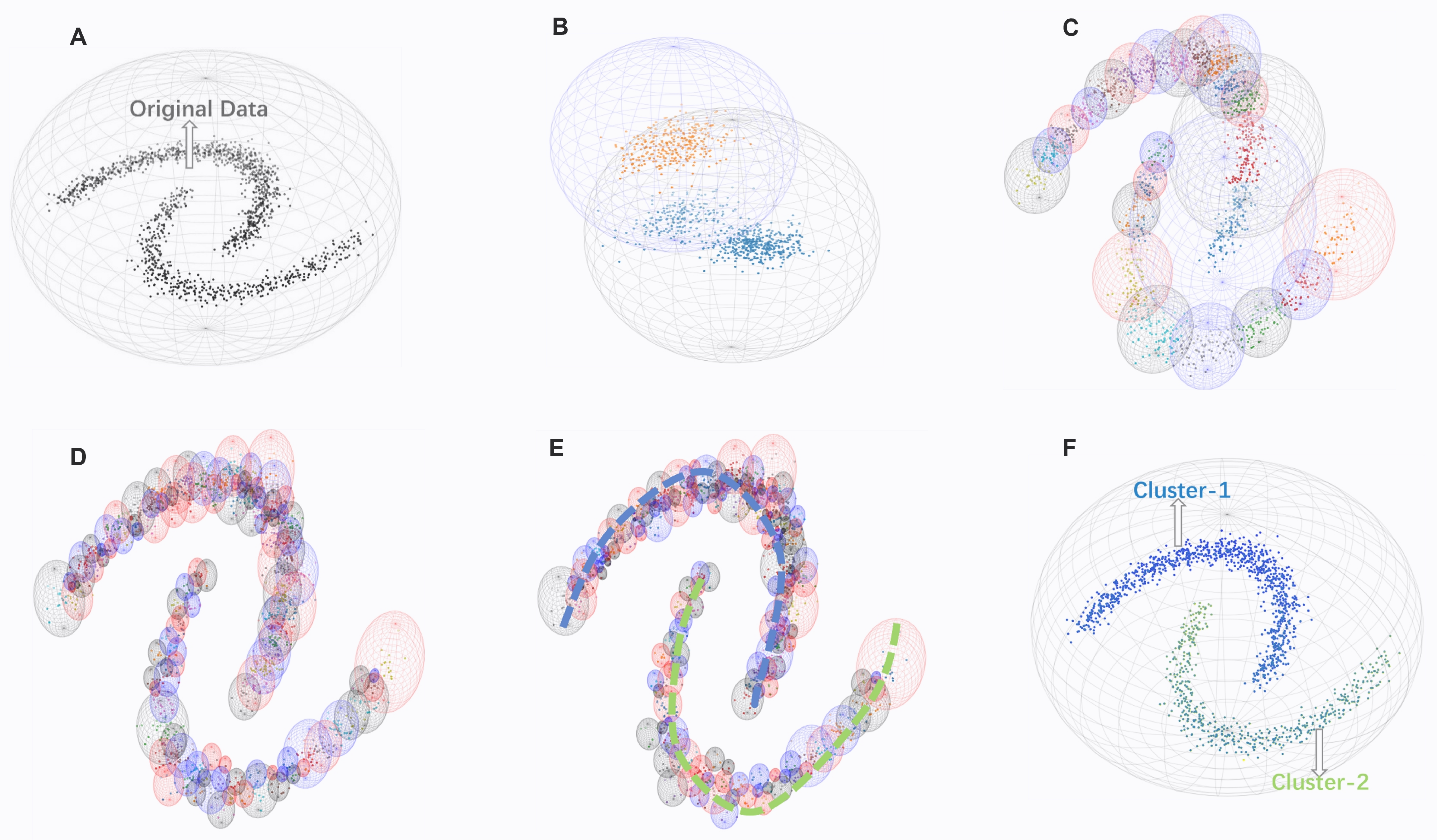}
	\captionsetup{name={Fig.},labelsep=period,singlelinecheck=off,font={small},justification=raggedright}
	\caption{The process of granular-ball clustering.  (A) Point distribution for the Moon dataset. (B)-(D) The granular-ball division process. (E) The granular-ball differentiation process. (F)  The clustering results.}\label{t2}
	\vspace{0em}
\end{figure}

\begin{figure*}[h]
	\centering
	\includegraphics[width=\linewidth]{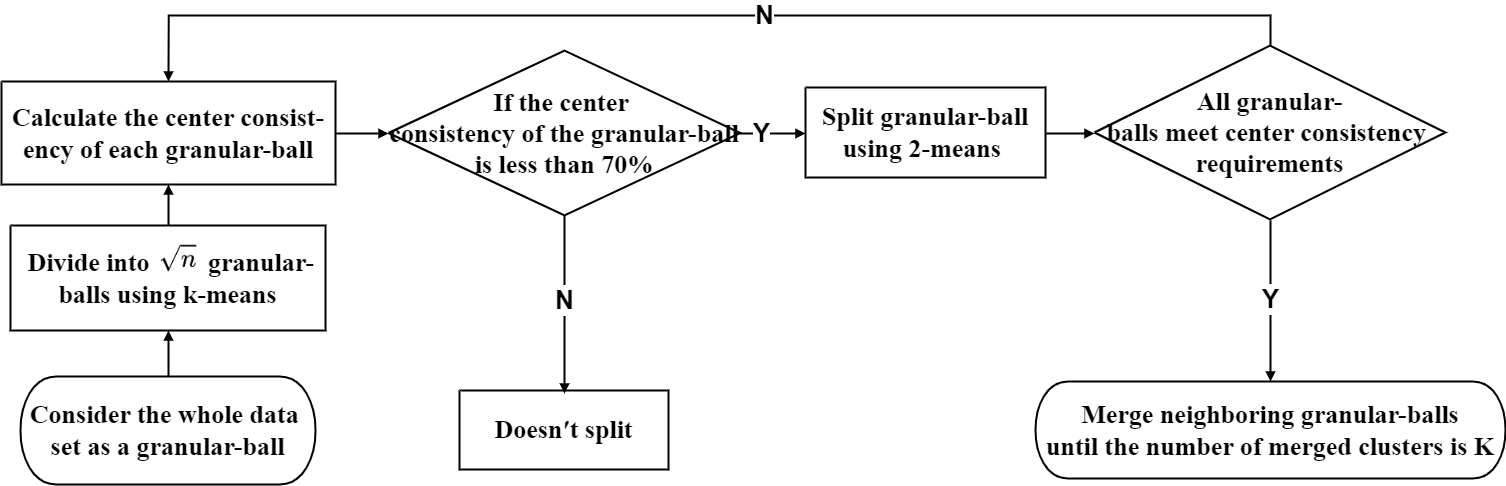}
	\caption{The Flow Chart of GBCT}\label{t3}
	%\Description{A woman and a girl in white dresses sit in an open car.}
\end{figure*}

The model of GBCT is similar with \autoref{1}, and the splitting process of the ~\lq\lq global precedence\rq\rq~ is used for solve the model. The clustering process of GBCT is divided into two phases: granular-ball generation and cluster formation. \autoref{t2} provides a simplified illustration of the GBCT method. Following the ~\lq\lq global precedence\rq\rq~ cognitive mechanism, GBCT initially treats the entire original dataset as the coarsest GB, where the granular-ball cannot fit complex data. Thus, it necessitates the adaptive splitting into GBs of different granularities by designing quality metrics and splitting stop conditions, achieving an adaptive, multi-granular, and flexible coverage and representation of the original data distribution. The clustering process consisting of granular-ball division and differentiation is shown in \autoref{t2} (A-F). The data set in \autoref{t2} can be seen as two clusters. In \autoref{t2} (A), the data set is observed as a single granular-ball. In \autoref{t2} (B-C), as the granular-balls divide, the clusters grow and form. In \autoref{t2} (D), two clusters have taken shape and matured. In \autoref{t2} (E), granular-balls that are close together form an cluster in \autoref{t2} (F). \autoref{t3} shows a more detailed process description of GBCT.

In the granular-ball generation process, it is necessary to use a specific metric to measure the quality of a granular-ball to judge whether it is needed to be splitting or not. However, it is very hard to make a specific metric effective when the data distribution of a granular-ball is two complex. So, the splitting process consists of two parts including a coarse division and a fine division of binary splits. The coarse division transforms the whole data into  \( \sqrt{n} \)  granular-balls in which the data is approximately spherical distributed, enabling a metric can measure the quality of a granular-ball more accurate and effective. \( n \) denotes the number of data points in a data set. In the fine division, a granular-ball is splitting to be two sub-granular-balls if its quality does not meet the requirement condition. In the coarse division, setting the number to $ \sqrt{n}$ is to balance the time complexity and data distribution of granular-balls. Larger the number of granular-balls, simpler the data distribution in a granular-ball, which makes a specific metric more effective; however, too large number may lead to over division for some granular-balls and a high time complexity. The number of $ \sqrt{n}$ can make time complexity limited to $O(n^{1.5})$, which is analyzed in the next section.

\subsection{Adaptive Splitting}
After establishing the fundamental mathematical model and algorithmic concept of GBCT, it's necessary to proceed with a secondary adaptive splitting of the initially spherical cluster-distributed data, based on the granular-balls obtained from the coarse division. At this juncture, an appropriate granular-ball quality metric and precise splitting stop conditions are crucial for determining the final fitting results of the granular-balls and the clustering outcome. In $\rm{GBCT\_old}$, the quality of a granular-ball was measured by its average radius. Therefore, $\rm{GBCT\_old}$ tends to split out too many small granular-balls, and needs other extra conditions to avoid to generate too many smallest granular-ball, i.e., a point. Specifically, in $\rm{GBCT\_old}$, a granular-ball does not split when its sub-granular-ball contains only one point. To address this problem, we gauge the quality of a granular-ball by the compactness of sample distribution within it, as reflected by the point density in it, which is defined as follows.
\begin{definition}\label{d2} For a $GB$, $GB's$ density, the point density in $GB$ ($\rho_{GB}$), is defined as follows:
\begin{equation}\label{2}
	\rho_{GB} =\frac{|GB|}{r^d},
\end{equation}
where \( |GB| \) denotes the number of objects in $GB$, $d$ denotes the data dimensionality, and \(r\) and \(c\) denotes its radius and center which are shown as follows:
\begin{equation}\label{3}
	r={\rm max}\parallel P_i-c\parallel, P_i \in GB,
\end{equation}

\begin{equation}\label{4}
	c = \frac{1}{|GB|} \sum_{i=1}^{|GB|} P_i, \quad P_i \in GB,
\end{equation}

where $\left \| \cdot  \right \| $ represents the $L_2$ norm, and the radius $r$ of $GB$  is defined as the maximum distance from the sample points in the granular-ball to its center. 
\end{definition}	

For a granular-ball, if its sub-granular-balls' density increased, it indicates the sparse region of the parent granular-ball is partially eliminated after splitting, in which case it should continue to split; otherwise, it should not continue to split.

In $\rm{GBCT\_old}$, splitting stops when the parent ball's quality exceeds the weighted quality of its sub-balls. The effectiveness of this weighted sub-ball splitting method has been thoroughly theoretically verified, with the proof detailed in the Supplementary\_Material. The results indicate that whether using the average radius or density as a measure of quality, the weighted sub-ball splitting approach may lead to over-splitting or premature convergence, where granular-balls that should continue splitting remain intact, while smaller granular-balls split prematurely into isolated balls. This finding is necessary and indirectly validates the rationality and effectiveness of the consistency measurement method we propose next. To determine whether a granular-ball should split by measuring the distribution within it, we designed a new granular-ball quality metric—center consistency measurement, as shown in \textit{Definition} \autoref{d3} .
\begin{definition}\label{d3}
For a $GB$, the densities of the granular-ball at the maximum radius and the average radius are respectively denoted as follows:
\begin{equation}\label{6}
	\mathrm{ave}\rho_{GB}=\frac{|\{x_i | \parallel x_i-c\parallel \le r_{ave}\}|}{r_{ave}^d}, 
\end{equation}
\begin{equation}
	r_{ave} = \frac{1}{|GB|} \sum_{i=1}^{|GB|} \parallel P_i-c\parallel, P_i \in GB,
\end{equation}
\begin{equation}\label{7}
	\mathrm{max}\rho_{GB}=\frac{|GB|}{r^d}.
\end{equation}
It is worth noting that the sample points within the average radius and the sample points inside the granular-ball are selected for the calculation of the average density and average radius of the granular-ball, respectively. $GB's$ center consistency metric ($\mathrm{con}_{GB}$) is defined as follows:
\begin{equation}\label{5}
	\mathrm{con}_{GB}=\frac{Min(ave\rho GB,max\rho GB)}{Max(ave\rho GB,max\rho GB)}.
\end{equation}

Center consistency metric characterizes the similarity between the central region and overall density. If the density within the central region is closer to the overall density of the granular-ball, $\mathrm{con}_{GB_{j}}$ approaches 1, indicating a more uniform distribution of sample points within the granular-ball, hence better granular-ball quality. Conversely, a large disparity between the center-range density and overall density suggests an uneven distribution, indicating poorer granular-ball quality. However, it is not necessary to set $\mathrm{con}_{GB_{j}}$ to 1 to assess granular-ball quality; this condition is too stringent for the granular-ball splitting generation. We consider an acceptable range for $\mathrm{con}_{GB_{j}}$ to be between 70\% and 1. If it is below 70\%, we consider it a low-quality granular-ball that needs to continue splitting. Factually, 70\% for center consistency metric is not a sensitive value, and effective for most data sets. The specific algorithm design is shown in \textbf{Algorithm} \autoref{A1} .

\end{definition}

\begin{algorithm}
	\caption{Granular-Ball Splitting}\label{A1}
	\begin{algorithmic}[1]
		\renewcommand{\algorithmicrequire}{\textbf{Input:}}
		\renewcommand{\algorithmicensure}{\textbf{Output:}}
		
		\REQUIRE Dataset $X = \{x_1, x_2, \ldots, x_n \}$.        %\in \mathbb{R}^d
		\ENSURE $GB$.
		\STATE Initialize $k$ = the integer part of $\sqrt{n}$;
		\STATE Perform $k$-means on $X$ to generate initial granular-balls $GB=\{GB_1, GB_2, \ldots, GB_{k}\}$;
		\REPEAT
		\FOR{each $GB_i$ in $GB$}
		\STATE Calculate ${\rm con}_{GB_i}$ by Eq. \ref{5} ;
		\IF{${\rm con}_{GB_i} < 70\% $}
		\STATE Employ $2$-division algorithm to split $GB_i$ into two sub-balls $GB_{i_1}$, $GB_{i_2}$;
		\STATE Compute the density of $GB_i$, $GB_{i_1}$ and $GB_{i_2}$ by Eq. \ref{2} ;
		\IF{$|GB_{i_1}|,|GB_{i_2}| \geq 2$ and {${\rm con}_{GB_{i_1}} ,{\rm con}_{GB_{i_2}} \geq 70\%$}}
		\STATE $GB = \{GB \setminus \{GB_i\}\} \cup \{GB_{i_1}, GB_{i_2}\}$.
		\ENDIF
		\ENDIF
		\ENDFOR
		\UNTIL{$|GB|$ does not increase.}
	\end{algorithmic}
\end{algorithm}

\subsection{Cluster Formation}
To facilitate the merging process of granular-balls, we introduce a distance metric between granular-balls, namely "boundary distance". For any two granular-balls, the boundary distance is defined as the centroid distance minus the sum of their maximum radius. This is formally defined as follows:
\begin{definition}\label{d4}
For any two granular-balls $GB_i$ and $GB_j$, where $c_i$ and $c_j$ are their centroids, and $r_i$ and $r_j$ represent the maximum radius of granular-balls $GB_i$ and $GB_j$ respectively, the boundary distance between any two granular-balls is defined as:
\begin{equation} \label{8}
	\mathrm{Dist_{raw}}(GB_i,GB_j)=\begin{Vmatrix}c_i-c_j\end{Vmatrix}-(r_i+r_j).
\end{equation}

Through \textit{Definition} \autoref{d4}, we calculate the raw boundary distance for all pairs of granular-balls, which serves as a measure of similarity between granular-balls. A smaller boundary distance indicates greater similarity between granular-balls. The boundary distance may be negative, indicating that the distance between the centers of two granular-balls is less than the sum of their radius. This is harmful for calculating similarity. Therefore, in order to eliminate negative values, the boundary distance is defined by adding twice the minimum negative absolute value, $\delta$, to the distance matrix as follows:

\begin{equation} \label{8_2}
	\mathrm{Dis}t(GB_i,GB_j)=\begin{Vmatrix}c_i-c_j\end{Vmatrix}-(r_i+r_j)+\delta.
\end{equation}

\begin{equation} \label{9}
	\mathrm{Sim}(GB_i,GB_j)=\begin{cases}
	\frac{1}{{\rm Dist}(GB_i,GB_j)}, & i \neq j \\
	0,                           & i=j
	\end{cases}.
\end{equation}

\autoref{8_2} and \autoref{9} is used to describe \autoref{8} from perspective of similarity. Factually, to find the highest similar granular-ball for a given granular-ball is equivalent to the granular-ball which has the minimal value of \autoref{8}, which could be a negative value. After two or more granular-balls are merged into one cluster, the distance between two clusters is defined as the distance between the two granular-balls with the maximum similarity within each cluster. In other words, the distance between two clusters is equal to the distance between the nearest granular-balls in the two clusters. The formula for calculating the similarity between two clusters is defined as:

\begin{equation} \label{10}
	\begin{gathered}
		\mathrm{Sim}(Cluster_i,Cluster_j)=max(\mathrm{Sim}(GB_p,GB_q)), \\
	s.t.\quad	GB_p\in Cluster_i,\\
	     \quad   	GB_q\in Cluster_j.
	\end{gathered}
\end{equation}

\end{definition}

\begin{algorithm}[!h]
	\caption{Merging Process}\label{A2}
	\label{alg:Merging}
	\renewcommand{\algorithmicrequire}{\textbf{Input:}}
	\renewcommand{\algorithmicensure}{\textbf{Output:}}	
	\begin{algorithmic}[1]
		\REQUIRE  The granular-ball splitting result $GB^{\prime}=\{GB_1, GB_2, \ldots, GB_m\}$, where $m$ denotes the number of granular-balls in $GB^{\prime}$. The number of final clusters is $K$.   %%input
		\ENSURE $K$ clusters.    %%output
		\STATE Calculate the average density of all granular-ball according to \autoref{7}, and then select granular-balls with a density less than 0.2 times the average density as noise granular-balls that will not be merged temporarily.
		\STATE Each granular-ball in $GB^{\prime}$ is treated as a cluster, and merge all the clusters in $GB^{\prime}$ that are most similar to each granular-ball and update $GB^{\prime}$;
		\STATE  Merge all the clusters in $GB^{\prime}$ again that are most similar to each granular-ball and update $GB^{\prime}$;
		\WHILE{$|GB^{\prime}| > K$}
		\STATE Calculate the similarity matrix $S$ of $GB^{\prime}$.
%		\IF {epoch $<$ gamma}
%		\STATE Merge $|GB^{\ast}|$ clusters that are most similar to each cluster.
%		\ELSE
		\STATE Merge only the $top_{|GB^{\prime}| - K}$ clusters that are most similar to each cluster.
%		\ENDIF
		\STATE Merge clusters containing the same sub-clusters.
		
%		\STATE epoch$++$.
		\ENDWHILE
		\STATE Set the labels of the noise granular-balls to the label of its closest cluster.
		\RETURN $K$ clusters
	\end{algorithmic}
\end{algorithm}

\begin{figure*}[h]
	\centering
	\includegraphics[width=\linewidth]{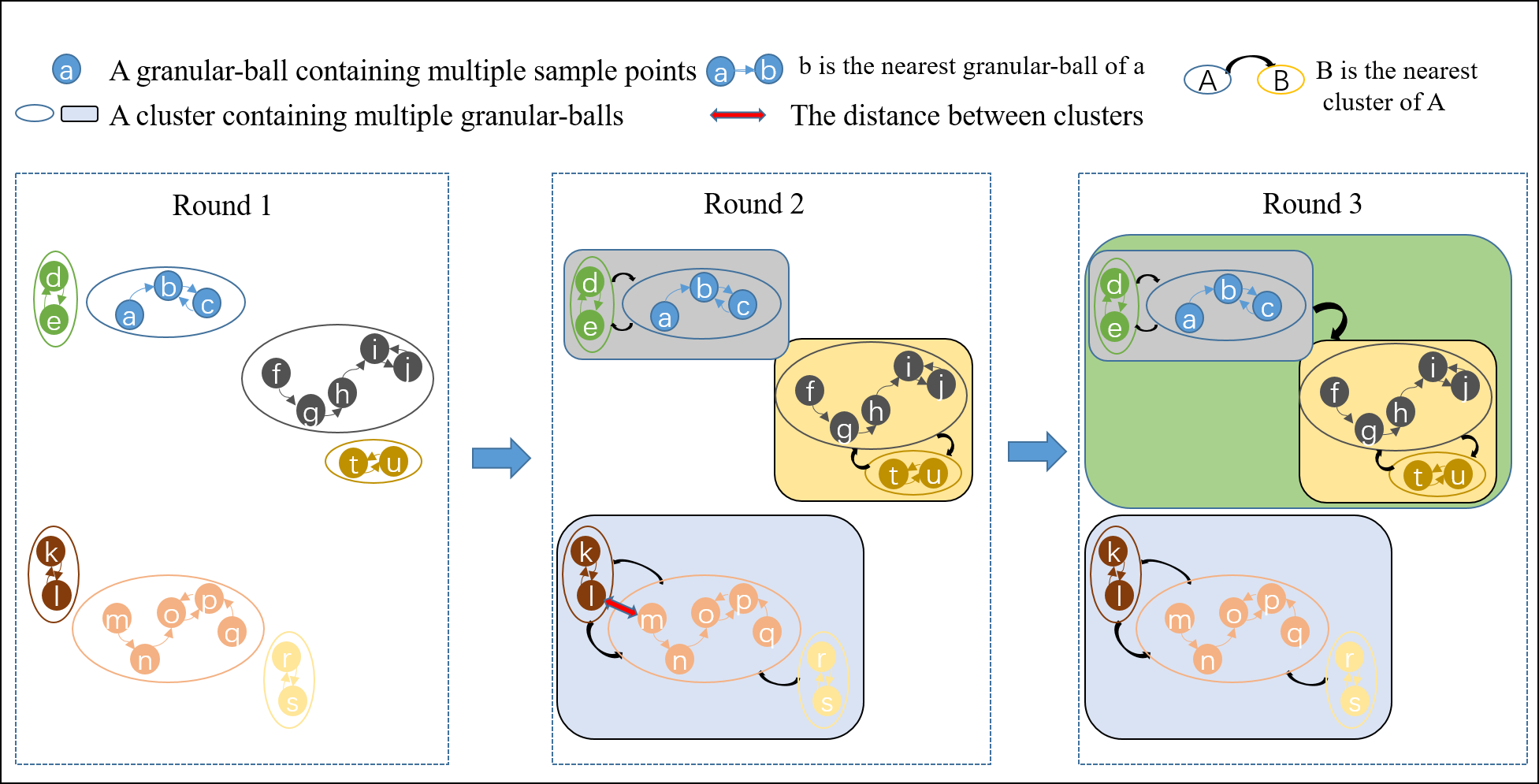}
	\caption{Cluster Formation Process}\label{t4}
\end{figure*}

The visualization of the merging algorithm is shown in \autoref{t4}. In the scenario depicted in \autoref{t4}, considering a clustering objective of $K=2$, the initial stage generates $a$ to $u$ granular-balls, each containing multiple sample points. The clustering process is divided into three rounds:

1) In Round 1, when there are a large number of granular-balls, each granular-ball seeks its closest granular-ball, with distances between granular-balls calculated according to \textit{Definition} \autoref{d4}. When nearest relationship is established between granular-balls, they form a cluster, represented by ellipses in the \autoref{t4}~. 

2) In Round 2, all clusters are still involved in finding the nearest cluster for merging(There are 7 clusters, so there are 7 corresponding black arrows). The similarity between two clusters is calculated according to \autoref{10}. So far, there are three remaining clusters.

3) In Round 3, not all clusters participate in finding the nearest cluster for merging. Instead, only $M-K$ clusters (where $M$ is the total number of clusters in the current round) search for their nearest clusters to merge with. When merging in the first and second rounds, we adopt a strategy of merging all clusters to avoid the influence of noise. In the third round and beyond, we use to merge only the first $M-K$ clusters that are most similar, where $M$ is the current number of clusters and $K$ is the final number of clusters. There are two reasons for using $M-K$: Firstly, to avoid clusters that have already been merged from being further merged into other clusters; Secondly, to consider the efficiency of the algorithm. If the current number of clusters is $M$, it is necessary to merge at least $M-K$ times to reach $K$ clusters (the clusters with the highest similarity need to be merged twice). The final 2 clusters are formed to meet the target.

\textbf{Discussion of solving overlapping problem}: In the merging phase, $\rm{GBCT\_old}$ directly treats the overlapping granular-balls as a cluster. Therefore, it cannot solve the clustering problem for overlapping datasets well because all the granular-balls may be overlapped. The merging of GBCT is based on the distance between granular-balls, which can divide the overlapping data into different clusters with closer distances, cleverly solving the clustering problem on overlapping datasets. This is also very beneficial for solving high-dimensional clustering problem because the overlapping problem is much common in high-dimensional data sets. The experimental results in the next section validate this advantage. The algorithm design of cluster merging process is shown in \textbf{Algorithm} \autoref{A2} .

\subsection{Time Complexity Analysis}
The granular-ball splitting algorithm and cluster merging algorithm are shown in \textbf{Algorithm} \autoref{A1} and \textbf{Algorithm} \autoref{A2} respectively. For \textbf{Algorithm} \autoref{A1}, in step 2, to generate a good coarse division result, we use complete k-means algorithm; however, in many cases, k-division can be used for a higher efficiency.

Let $n$ denote the number of samples in the dataset, $k$ represent the number of clusters, and the number of granular-balls is $m$. The $m$ is much smaller than the sample size $n$ in the dataset. The time complexity derivation of GBCT is as follows:

\textbf{Splitting Stage}: Coarse partitioning is performed using $K$-means, with the number of iterations set to $t$ (a constant). Therefore, the time complexity of coarse partitioning is $O(n*\sqrt{n}*t)$. Subsequently, center-consistent splitting is carried out, where each splitting operation involves traversing all granular-balls. For each granular-ball, a center-consistency check is conducted, and if the condition is satisfied, a $2$-means split is performed. The time complexity of each granular-ball split is $O(2*(n/m))$. In the average scenario, ending after $m * \log{\sqrt{n}}$ iterations, the average time complexity of center-consistent splitting is $O(2 * n *\log{\sqrt{n}})$. Therefore, the overall complexity of splitting is $O(n^{\frac{3}{2}})$.

\textbf{Merge Stage}: It is necessary to compute the "boundary distance" between each pair of granular-balls, with a time complexity of $O({m}^2)$. In the average scenario, merging must be performed $\log{m}$ times. Therefore, the time complexity of the merging process is calculated as $O({m}^2 * \log{m})$.

In summary, the overall time complexity of the GBCT algorithm is approximately $O(n^{\frac{3}{2}} + {m}^2 * \log{m})$. The \autoref{table1} depicts a comparison of the time complexity of GBCT with other algorithms(The explanation of the comparison algorithm is detailed in \autoref{s4}). In the KM algorithm, $k$ is the number of clusters and $t$ is the number of iteration rounds. The $\lambda$ in the HCDC algorithm is a value much smaller than $n$. 

\begin{table}[]
	\centering
	\caption{Time Complexity Comparison}\label{table1}
	\begin{tabular}{cc}
		\hline
		Method   & Time Complexity                    \\ \hline
		KM       & $O(n*k*t)$                             \\
		DBSCAN   & $O(n^2)$           \\
		HCDC     & $O({(\lambda / 2 + 1)} * n^2)$           \\
		DP       & $O(n^2)$           \\
		AC      & $O(n^3)$            \\
		SC       & $O(n^3)$           \\
		GBDP     & $O(n * \log{\sqrt{n}})$  \\
		GBSC     & $O({m}^3)$            \\
		GBCT\_old & $O(n^2)$            \\ 
		GBCT      &  $O(n^{\frac{3}{2}} + {m}^2 * \log{m})$ \\ \hline
	\end{tabular}
\end{table}

\section{experiment}\label{s4}
In this section, we demonstrate the power of GBCT in terms of effectiveness, efficiency, and robustness through both synthetic and real datasets.
\subsection{Experimental Setup}
\subsubsection{ Experimental Environment}

Given that this paper involves the performance evaluation of algorithms, the experimental environment may have a certain impact on the final results. Therefore, the fairness of the experiment was thoroughly considered during the experimental process. The experiments were conducted within a Python environment. All experiments in this paper were carried out on a computer equipped with 32GB DDR4 memory and an 11th Gen Intel(R) Core(TM) i7-11700 @ 2.50GHz processor. The computer's operating system is Windows 10 Professional, version 22H2. To ensure fairness in comparison of execution times, all algorithms were implemented in Python.

\subsubsection{ Datasets}
To validate the clustering performance of the GBCT algorithm in the experiments, the experiments were conducted on 15 synthetic datasets, 3 UCI \cite{40} real datasets, 4 overlapping datasets, and 6 noise datasets. The detailed information for the synthetic datasets, real datasets, overlapping datasets, and noise datasets is presented in \autoref{table2} and \autoref{table_merging} respectively.

\begin{table*}[]
	\centering
	\caption{Information of the Synthetic Datasets}\label{table2}
	\begin{tabular}{cccccccccccccccc}
		\hline
		Datasets  & A    & B    & C    & D    & E    & F    & G    & H   & I    & J    & K   & L    & M    & N    & O    \\ \hline
		Instances & 1735 & 7679 & 7200 & 6800 & 1438 & 1641 & 1039 & 212 & 1043 & 1020 & 567 & 3603 & 1427 & 1016 & 6699 \\
		Clusters  & 6    & 7    & 6    & 9    & 3    & 3    & 4    & 3   & 2    & 3    & 2   & 3    & 4    & 4    & 5    \\ \hline
	\end{tabular}
\end{table*}

\begin{table}[]
	\centering
	\caption{Information of the Real Datasets, Overlapping Datasets and Noise Datasets}\label{table_merging}
	\begin{tabular}{ccccc}
		\hline
		Datasets & Instances & Clusters & Dimensions & Type        \\ \hline
		cell     & 8680      & 20       & 2          & real        \\
		soybean  & 682       & 3        & 35         & real        \\
		mushroom & 8123      & 2        & 22         & real        \\
		overlap1 & 788       & 7        & 2          & overlapping \\
		overlap2 & 1680      & 3        & 2          & overlapping \\
		overlap3 & 1500      & 3        & 2          & overlapping \\
		overlap4 & 1400      & 2        & 2          & overlapping \\
		noise1   & 2082      & 6        & 2          & noise       \\
		noise2   & 9214      & 7        & 2          & noise       \\
		noise3   & 2002      & 2        & 2          & noise       \\
		noise4   & 4804      & 3        & 2          & noise       \\
		noise5   & 8000      & 6        & 2          & noise       \\
		noise6   & 8000      & 9        & 2          & noise       \\ \hline
	\end{tabular}
\end{table}

%
%\begin{table}[ht]
%	\centering
%	\caption{Information of the Real Datasets}\label{table3}
%	\begin{tabular}{@{}cccc@{}} % Four columns, all centered (change c to l or r for left or right alignment)
%		\toprule
%		Datasets & Instances & Clusters & Dimensions \\ 
%		\midrule
%		cell & 8680 & 20 & 2 \\
%		soybean & 682 & 3 & 35 \\
%		mushroom & 8123 & 2 & 22 \\
%		\bottomrule
%	\end{tabular}
%\end{table}
%
%\begin{table}[ht]
%\centering
%\caption{Information of the Overlapping Datasets}\label{table4}
%\begin{tabular}{lccc}
%\hline
%Datasets & Instances & Clusters & Dimensions \\ \hline
%overlap1 & 788       & 7        & 2          \\
%overlap2 & 1680      & 3        & 2          \\
%overlap3 & 1500      & 3        & 2          \\
%overlap4 & 1400      & 2        & 2          \\ \hline
%\end{tabular}
%\end{table}
%
%
%\begin{table}[ht]
%\centering
%\caption{Information of the Noise Datasets}\label{table5}
%\begin{tabular}{cccc}
%\hline
%\multicolumn{1}{l}{Datasets} & Instances & Clusters & Dimensions \\ \hline
%noise1                       & 2082      & 6        & 2          \\
%noise2                       & 9214      & 7        & 2          \\
%noise3                       & 2002      & 2        & 2          \\
%noise4                       & 4804      & 3        & 2          \\
%noise5                       & 8000      & 6        & 2          \\
%noise6                       & 8000      & 9        & 2          \\ \hline
%\end{tabular}
%\end{table}

\subsubsection{ Baselines}
We compared our proposed GBCT method with five widely used basic clustering algorithms and four powerful clustering algorithms as follows:

K-means clustering algorithm, abbreviated as KM.
Density-based spatial clustering of applications with noise, abbreviated as DBSCAN.
Clustering by fast search and find of density peaks, abbreviated as DP.
Spectral clustering, abbreviated as SC.
Agglomerative hierarchical clustering algorithm-a\cite{44}, abbreviated as AC.
A novel hierarchical clustering algorithm based on density-distance cores for datasets with varying density\cite{41}, abbreviated as HCDC.
An efficient and adaptive clustering algorithm based on Granular-ball, abbreviated as  $\rm{GBCT\_old}$.
A Fast Granular-ball Based Density Peaks Clustering Algorithm for Large-Scale Data\cite{26}, abbreviated as GBDP.
An Efficient Spectral Clustering Algorithm Based on Granular-Ball\cite{43}, abbreviated as GBSC.

\subsubsection{Evaluating indicator}
The clustering results were evaluated using two external criteria: clustering accuracy (ACC) and normalized mutual information (NMI), which can be calculated
according to \cite{18}. Higher values of ACC and NMI indicate better clustering performance. Additionally, as a complement to the external evaluation criteria, the synthetic datasets were also visualized.

\subsection{Effectiveness}
We demonstrate the effectiveness of GBCT from three aspects: synthetic dataset, real dataset, and overlapping dataset.
\subsubsection{Clustering on Synthetic Datasets}
\begin{figure}[htp]
	\centering
	
	% 第一行
	\begin{minipage}[b]{0.32\linewidth}
		\centerline{\includegraphics[width=\textwidth]{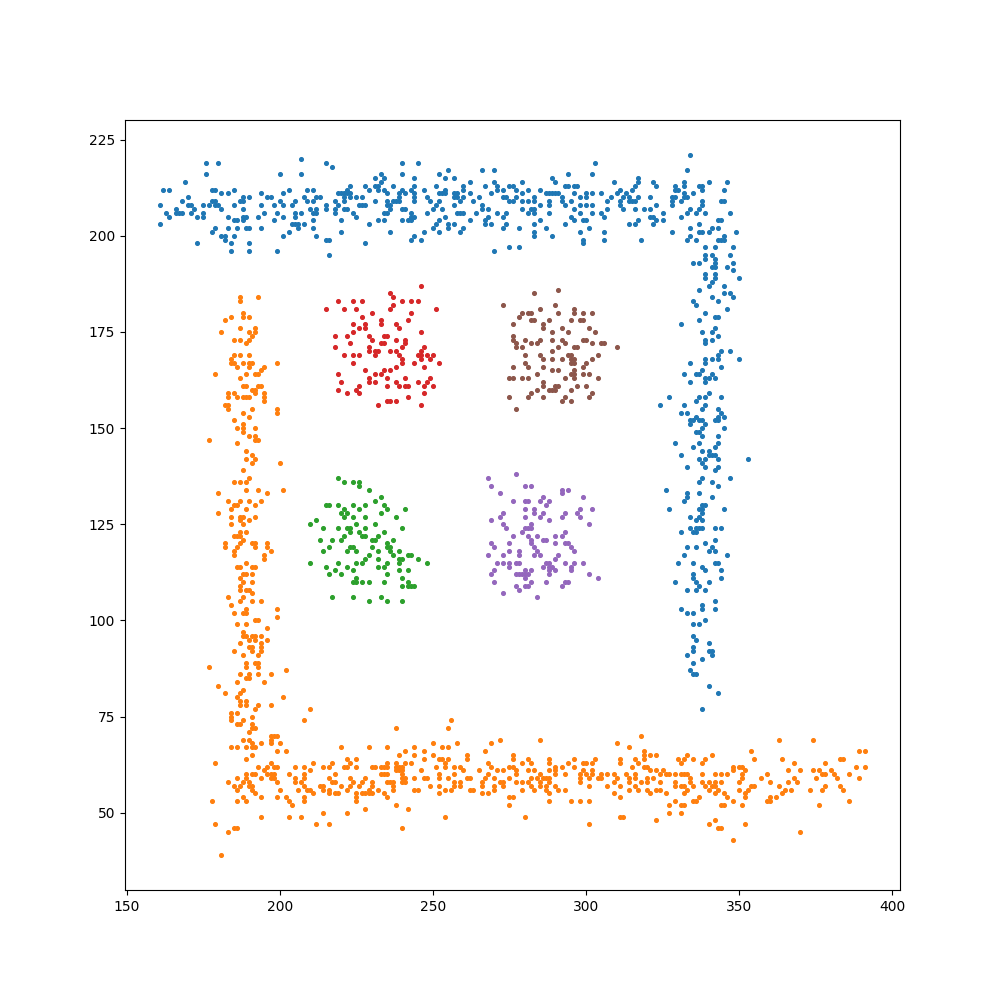}}
		\centerline{A}
		\label{fig:A}
	\end{minipage}
	% \hfill % 这个命令在子图之间添加了一些水平空间
	% 重复上面的代码块，更改图片路径、caption和label来添加其他图片
	\begin{minipage}[b]{0.32\linewidth}
		\centerline{\includegraphics[width=\textwidth]{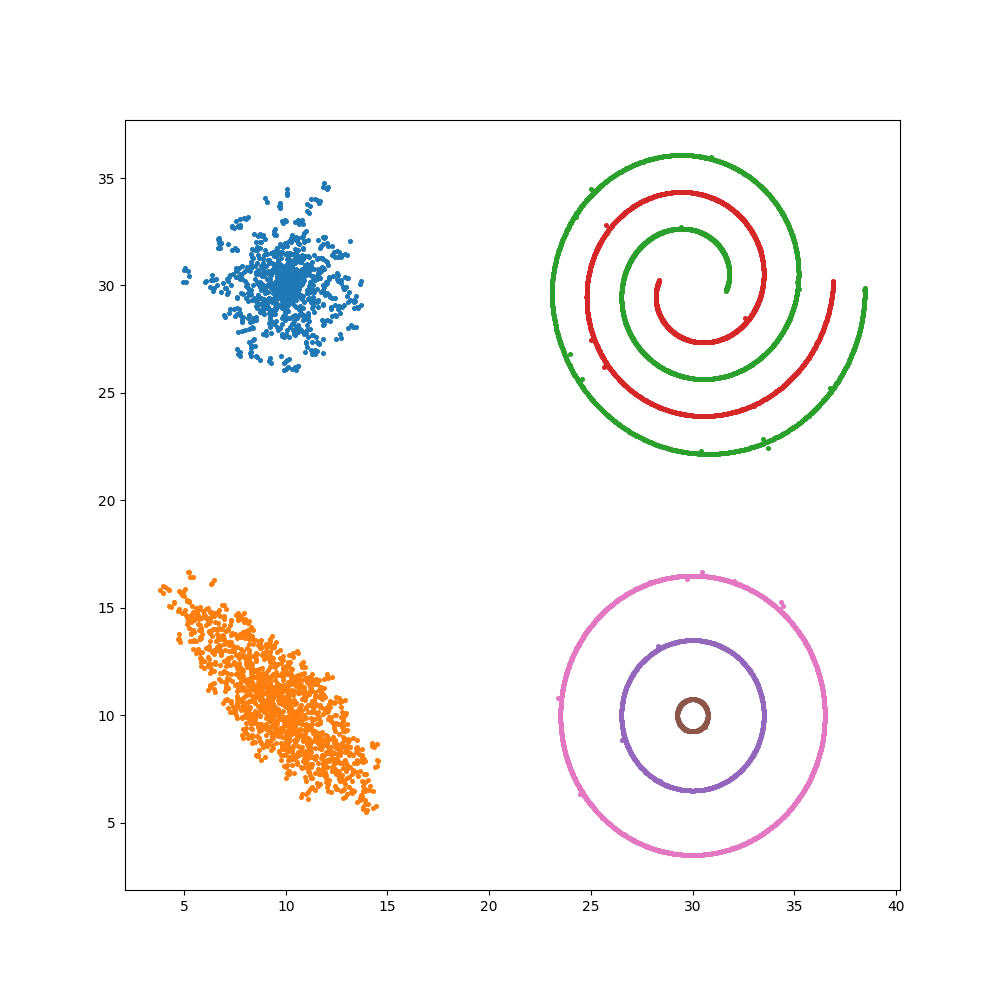}}
		\centerline{B}
		\label{fig:A}
	\end{minipage}
	% \hfill % 这个命令在子图之间添加了一些水平空间
	\begin{minipage}[b]{0.32\linewidth}
		\centerline{\includegraphics[width=\textwidth]{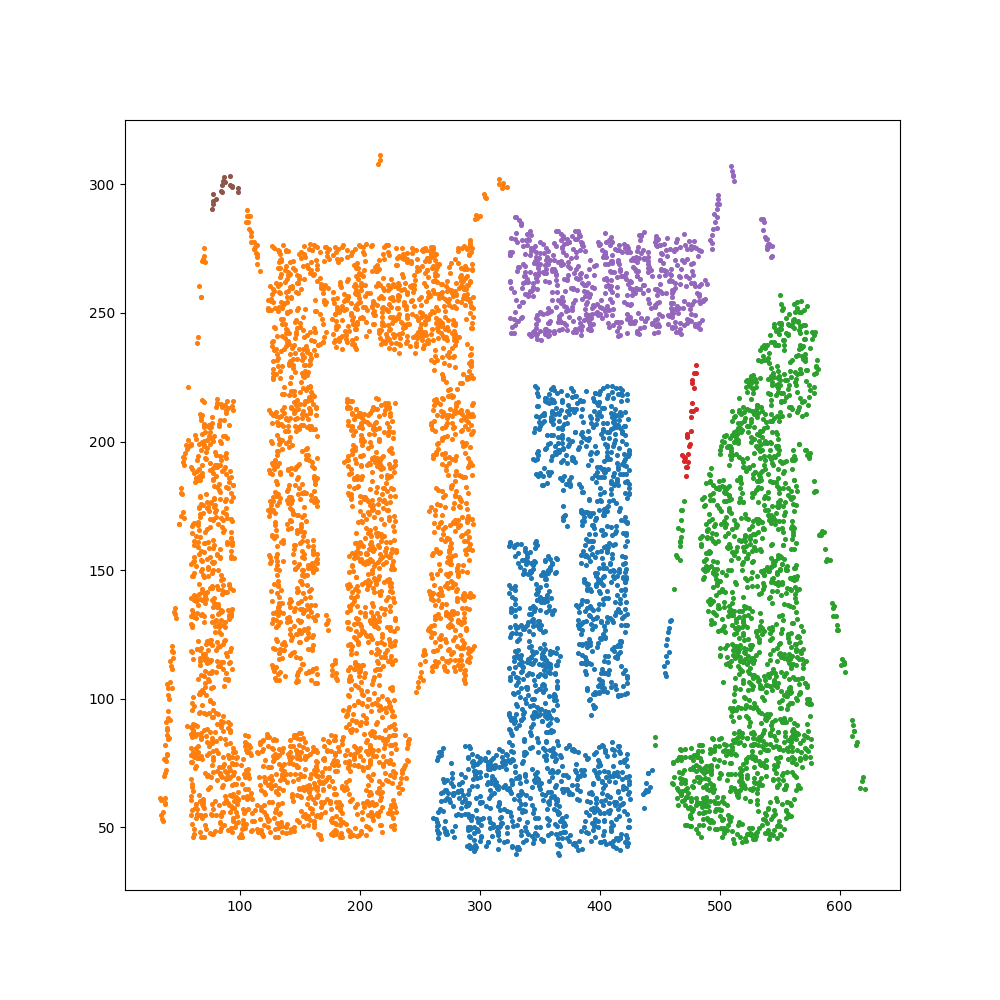}}
		\centerline{C}
		\label{fig:A}
	\end{minipage}
	% \hfill % 这个命令在子图之间添加了一些水平空间
	
	% 第一行
	\begin{minipage}[b]{0.32\linewidth}
		\centerline{\includegraphics[width=\textwidth]{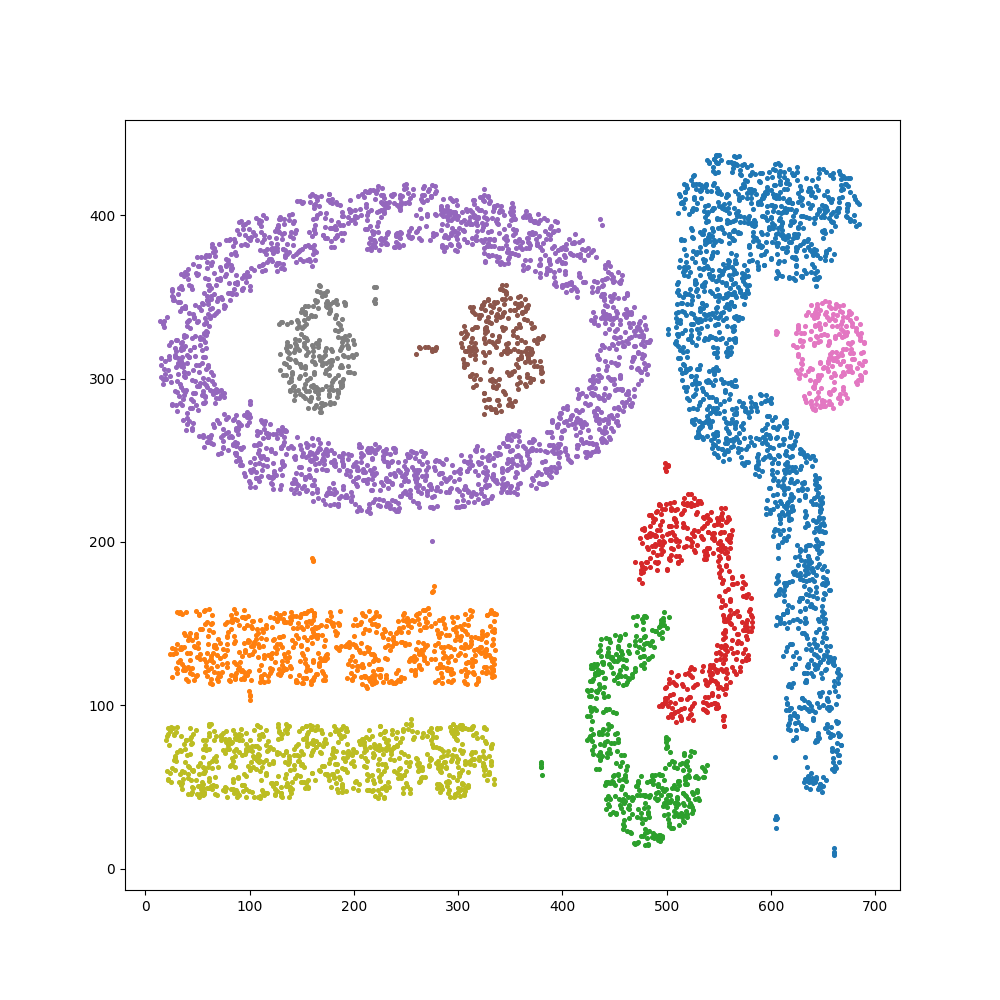}}
		\centerline{D}
		\label{fig:A}
	\end{minipage}
	% \hfill % 这个命令在子图之间添加了一些水平空间
	% 重复上面的代码块，更改图片路径、caption和label来添加其他图片
	\begin{minipage}[b]{0.32\linewidth}
		\centerline{\includegraphics[width=\textwidth]{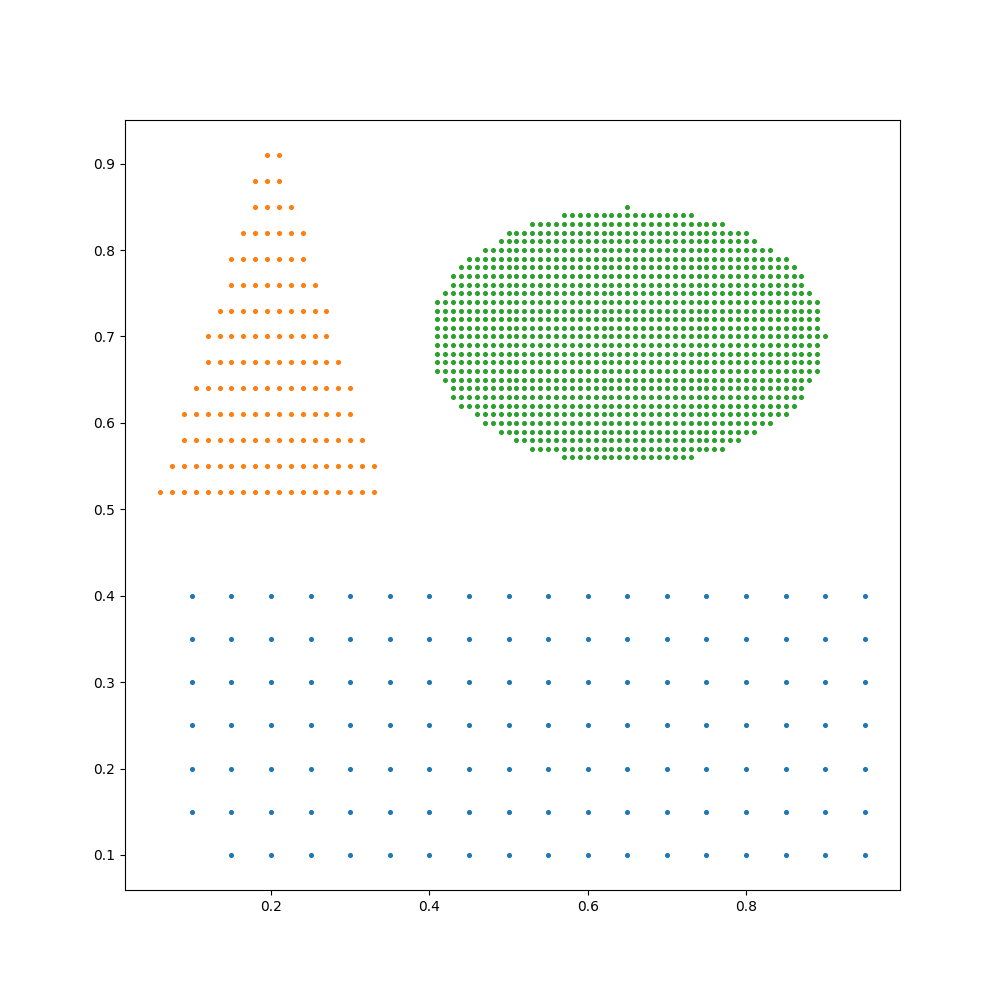}}
		\centerline{E}
		\label{fig:A}
	\end{minipage}
	% \hfill % 这个命令在子图之间添加了一些水平空间
	\begin{minipage}[b]{0.32\linewidth}
		\centerline{\includegraphics[width=\textwidth]{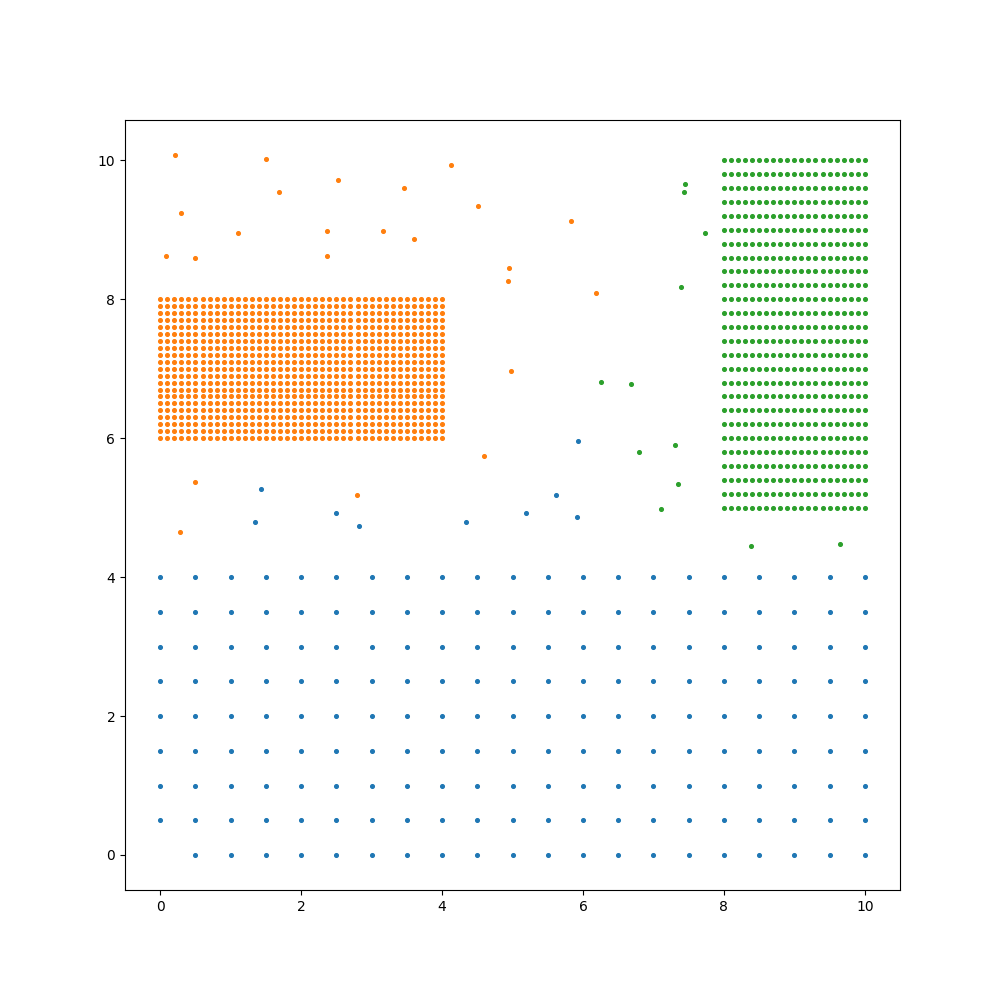}}
		\centerline{F}
		\label{fig:A}
	\end{minipage}
	% \hfill % 这个命令在子图之间添加了一些水平空间
	
	% 第一行
	\begin{minipage}[b]{0.32\linewidth}
		\centerline{\includegraphics[width=\textwidth]{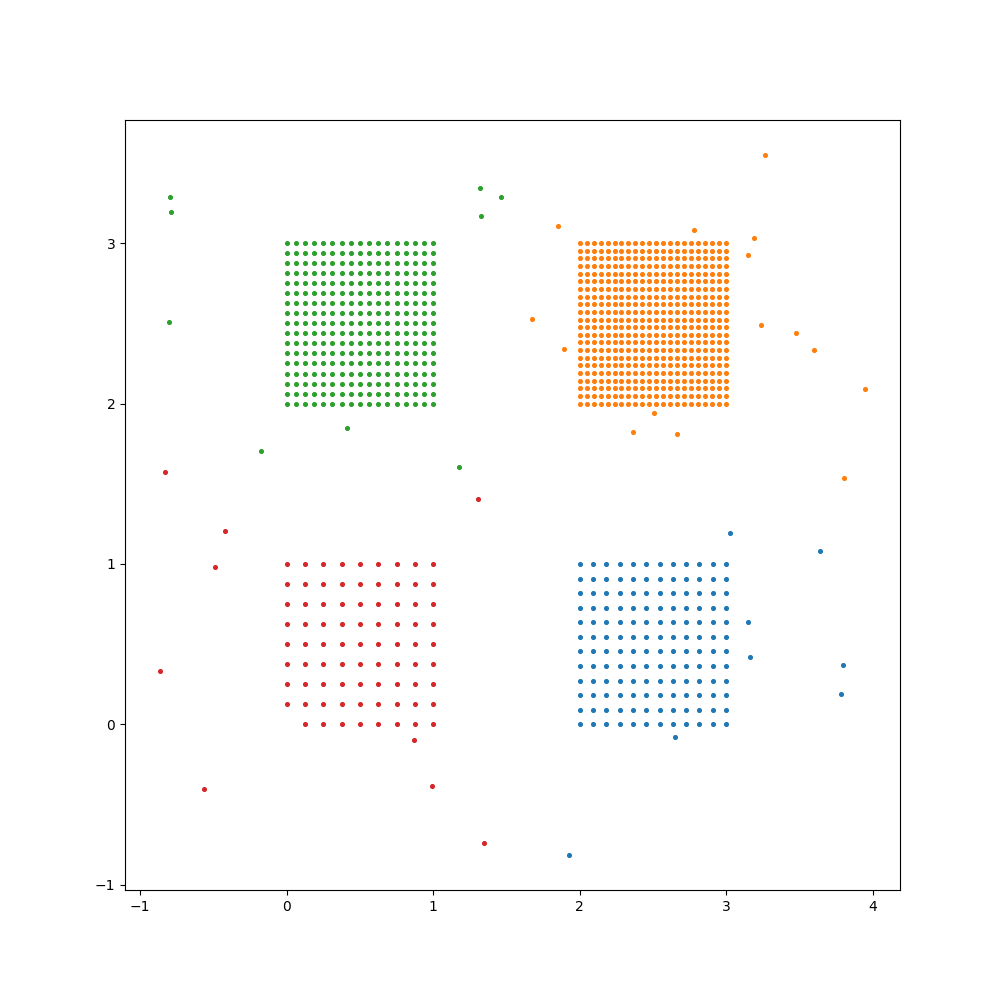}}
		\centerline{G}
		\label{fig:A}
	\end{minipage}
	% \hfill % 这个命令在子图之间添加了一些水平空间
	% 重复上面的代码块，更改图片路径、caption和label来添加其他图片
	\begin{minipage}[b]{0.32\linewidth}
		\centerline{\includegraphics[width=\textwidth]{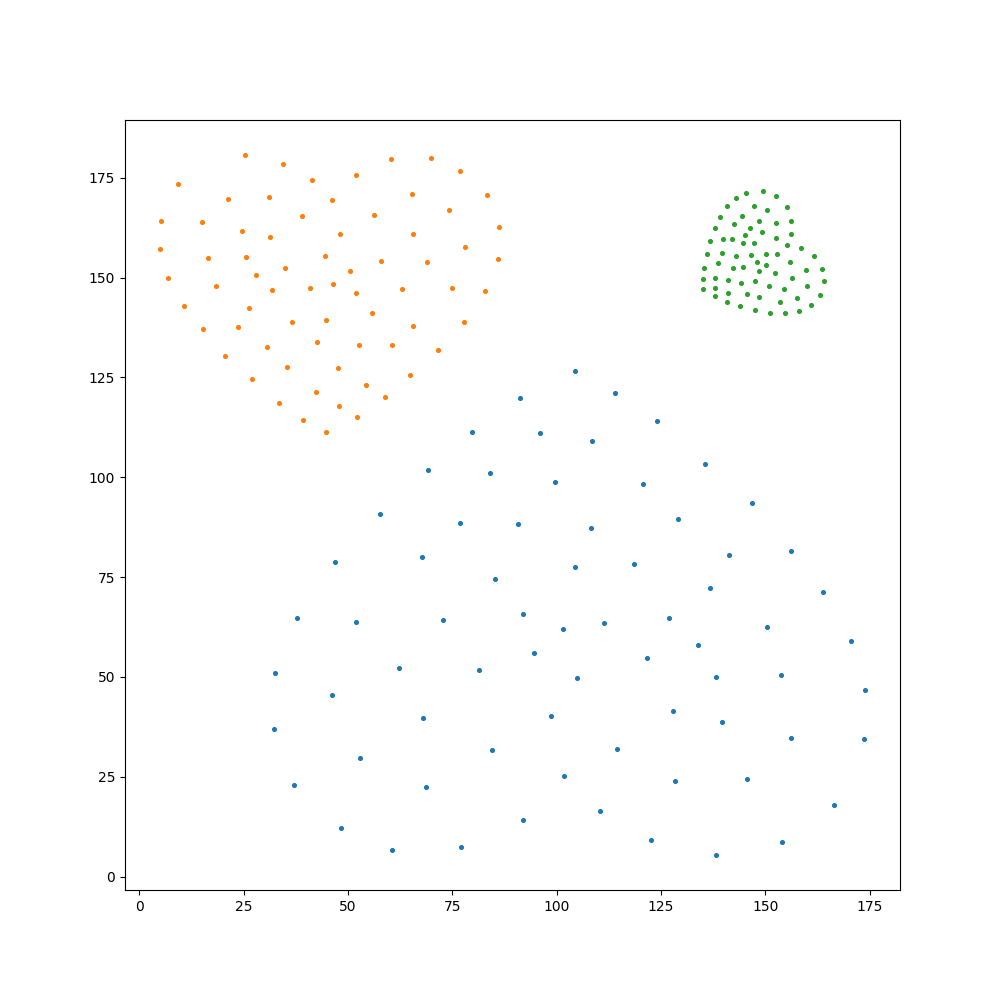}}
		\centerline{H}
		\label{fig:A}
	\end{minipage}
	% \hfill % 这个命令在子图之间添加了一些水平空间
	\begin{minipage}[b]{0.32\linewidth}
		\centerline{\includegraphics[width=\textwidth]{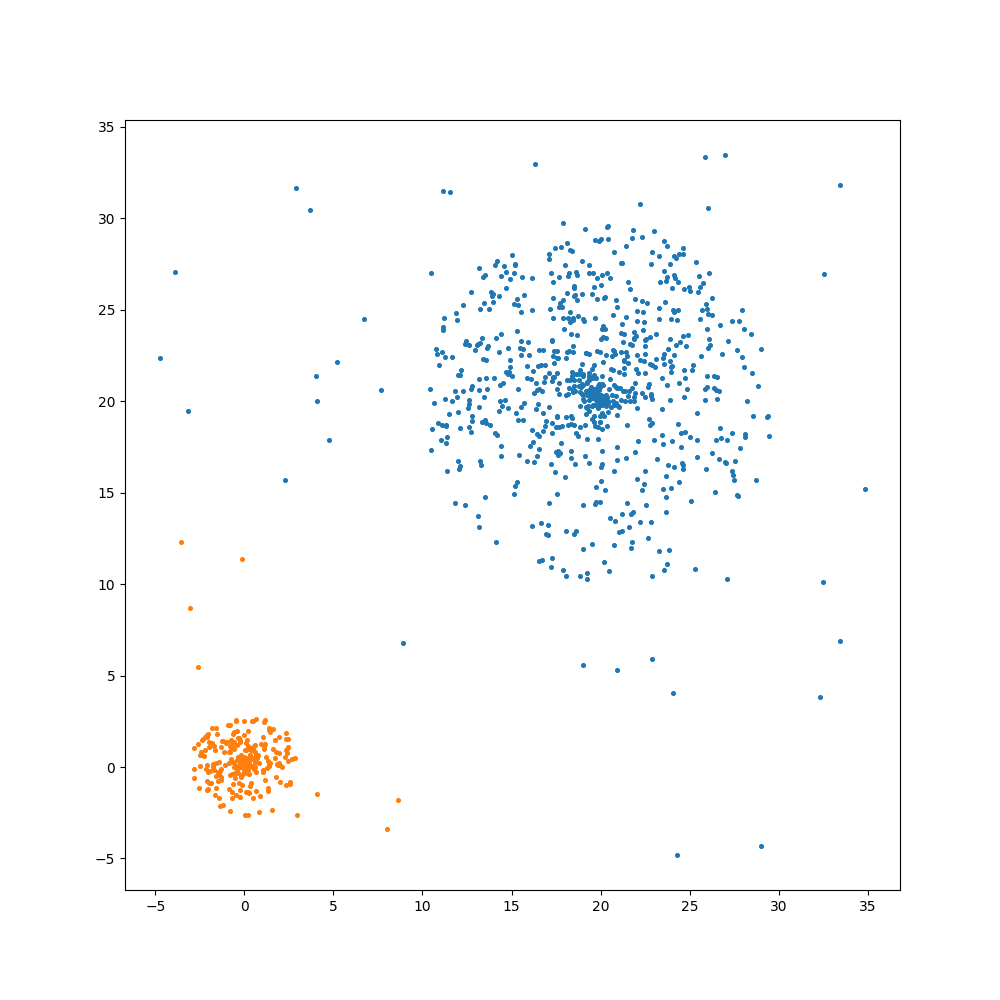}}
		\centerline{I}
		\label{fig:A}
	\end{minipage}
	% \hfill % 这个命令在子图之间添加了一些水平空间
	
	% 第一行
	\begin{minipage}[b]{0.32\linewidth}
		\centerline{\includegraphics[width=\textwidth]{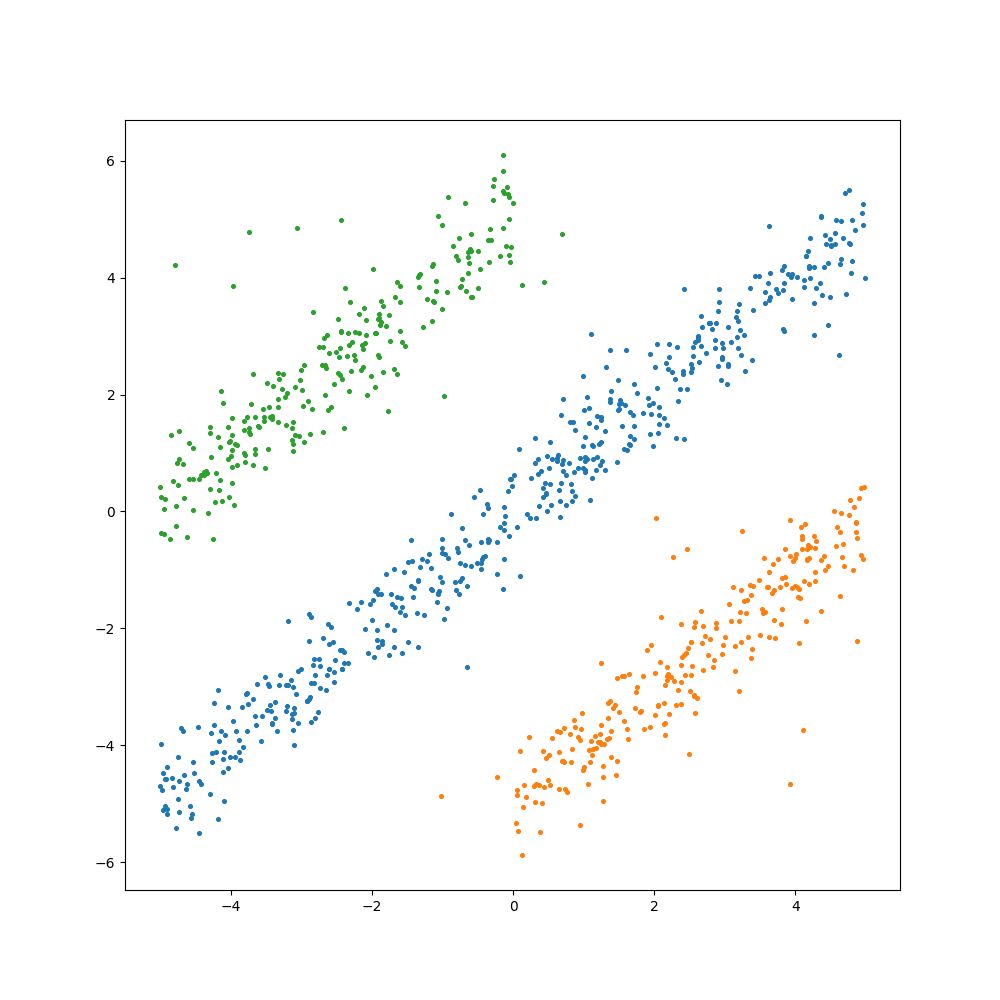}}
		\centerline{J}
		\label{fig:A}
	\end{minipage}
	% \hfill % 这个命令在子图之间添加了一些水平空间
	% 重复上面的代码块，更改图片路径、caption和label来添加其他图片
	\begin{minipage}[b]{0.32\linewidth}
		\centerline{\includegraphics[width=\textwidth]{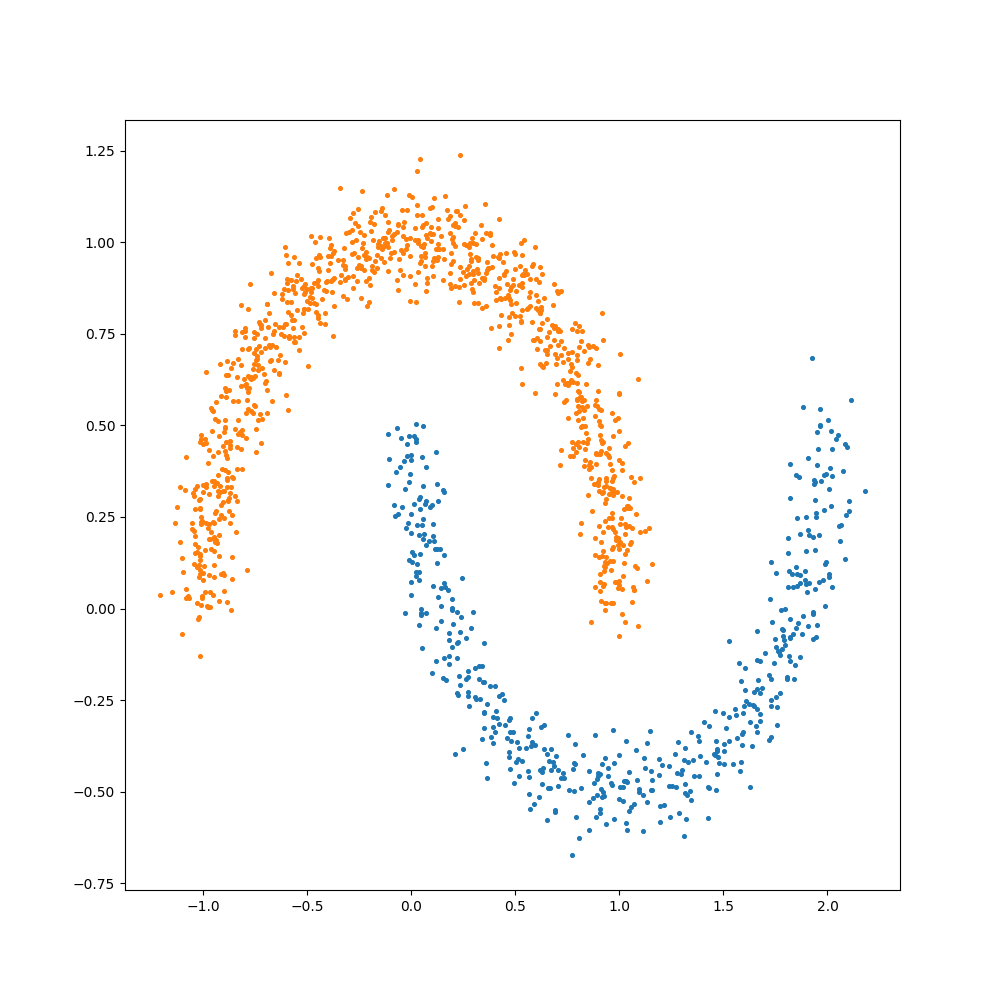}}
		\centerline{K}
		\label{fig:A}
	\end{minipage}
	% \hfill % 这个命令在子图之间添加了一些水平空间
	\begin{minipage}[b]{0.32\linewidth}
		\centerline{\includegraphics[width=\textwidth]{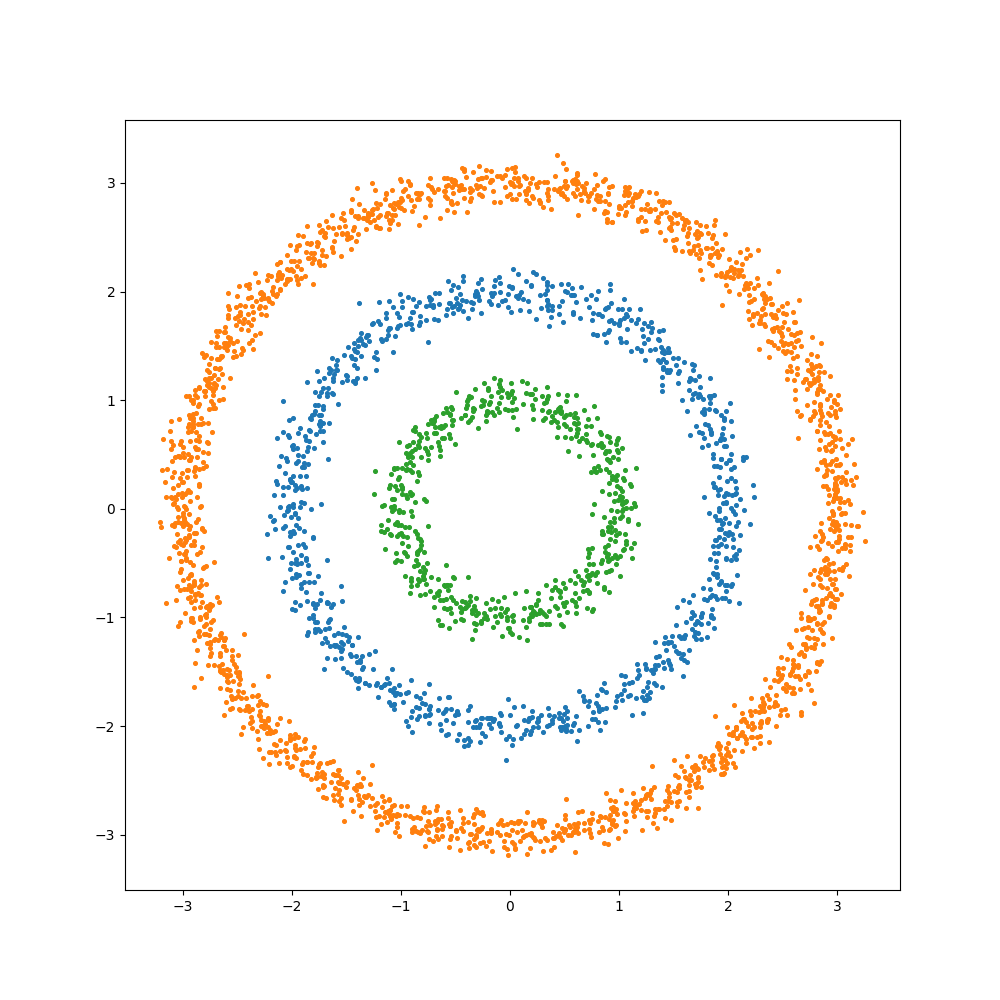}}
		\centerline{L}
		\label{fig:A}
	\end{minipage}
	% \hfill % 这个命令在子图之间添加了一些水平空间
	
	% 第一行
	\begin{minipage}[b]{0.32\linewidth}
		\centerline{\includegraphics[width=\textwidth]{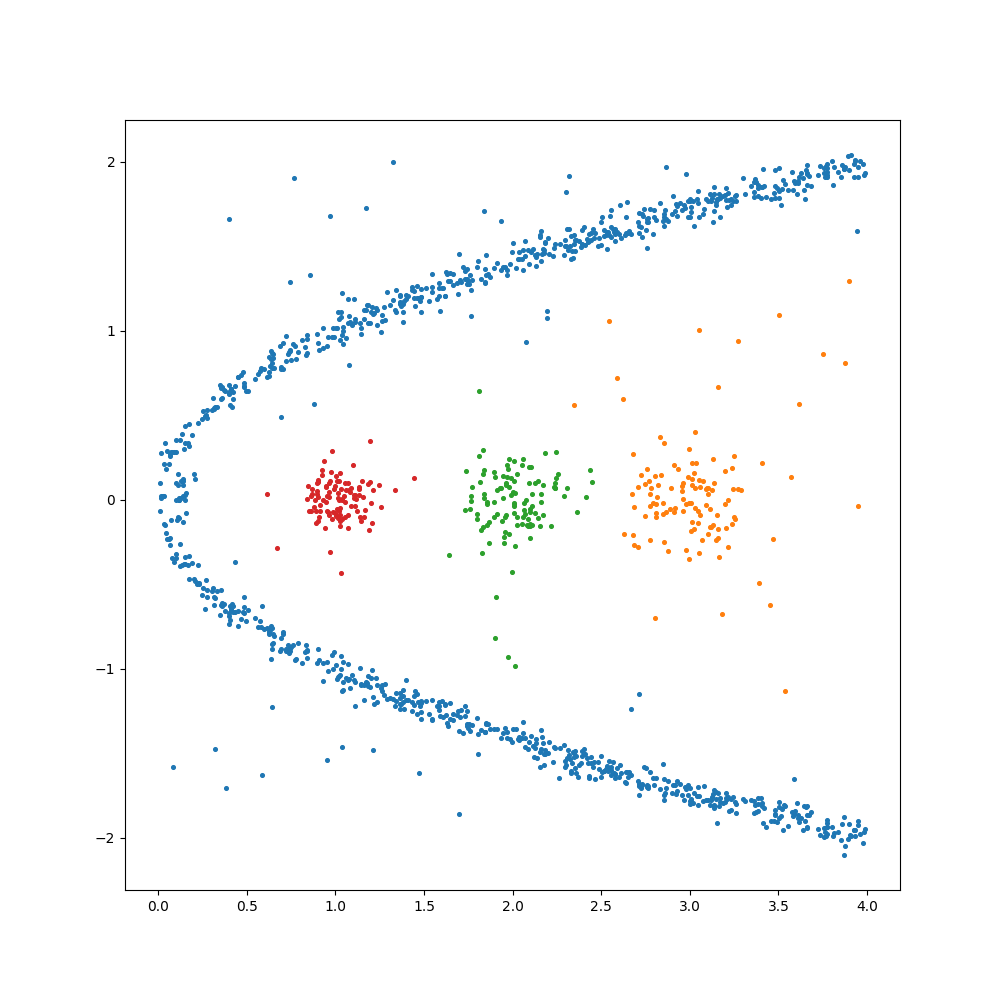}}
		\centerline{M}
		\label{fig:A}
	\end{minipage}
	% \hfill % 这个命令在子图之间添加了一些水平空间
	% 重复上面的代码块，更改图片路径、caption和label来添加其他图片
	\begin{minipage}[b]{0.32\linewidth}
		\centerline{\includegraphics[width=\textwidth]{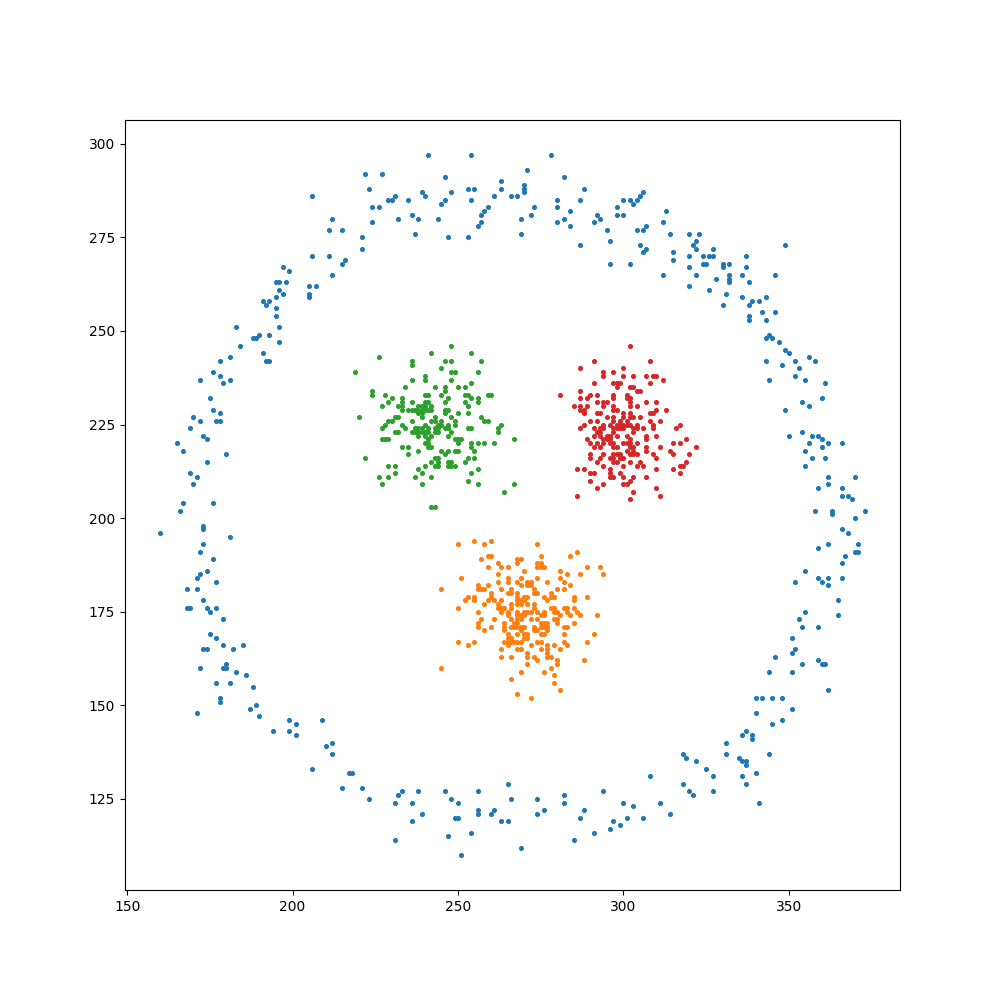}}
		\centerline{N}
		\label{fig:A}
	\end{minipage}
	% \hfill % 这个命令在子图之间添加了一些水平空间
	\begin{minipage}[b]{0.32\linewidth}
		\centerline{\includegraphics[width=\textwidth]{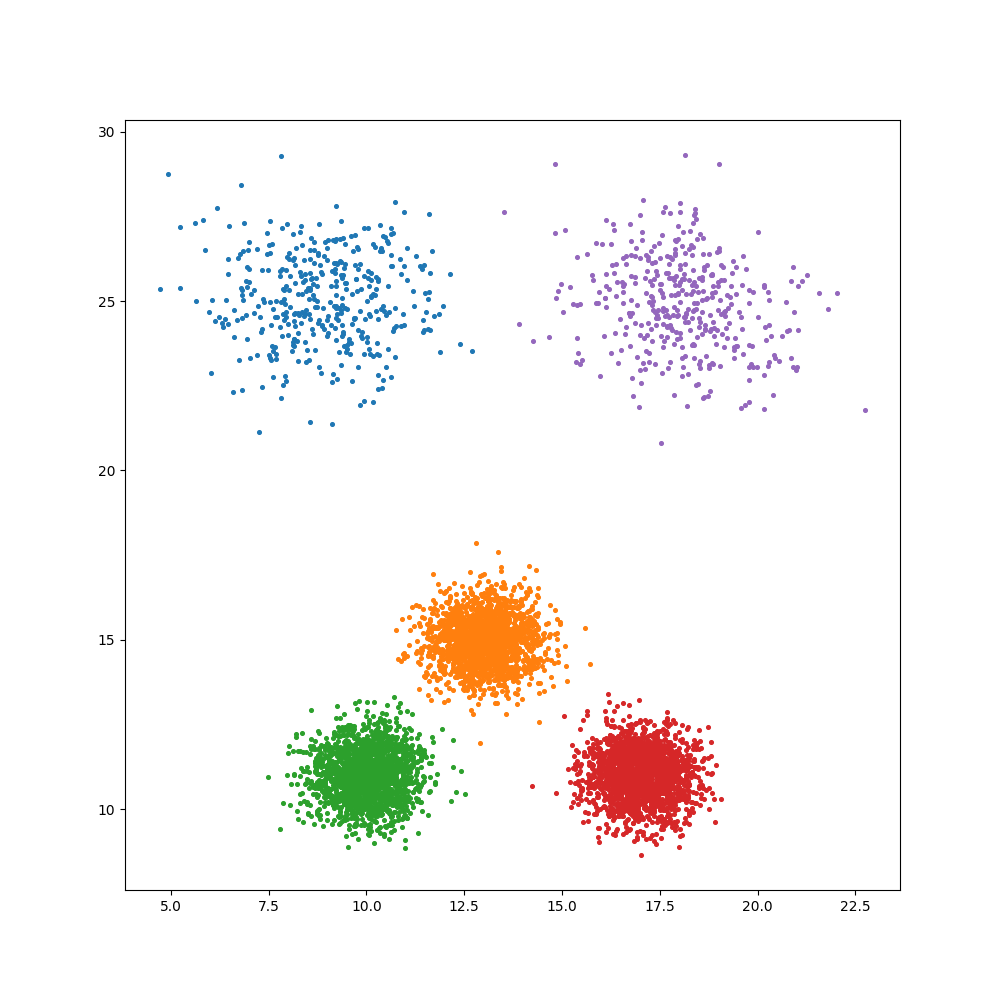}}
		\centerline{O}
		\label{fig:A}
	\end{minipage}
	% \hfill % 这个命令在子图之间添加了一些水平空间
	
	\caption{The Performance of GBCT on Synthetic Datasets}
	\label{t5}
\end{figure}

The ACC, NMI, and runtime tables for 15 synthetic datasets are shown in \autoref{table6}. The parameters of GBCT and all comparison algorithms are shown in \autoref{table7}. In the experiment, as a single point can affect the clustering effect, only granular-balls containing at least two sample points are considered in the clustering process.  
In \autoref{t5} (A), we consider a synthetic dataset composed of two L-shaped clusters and four spherical clusters. Our method correctly identifies the clustering structure of the dataset. In \autoref{t5} (B), the dataset consists of three concentric circular clusters, two intertwined spiral clusters, and two spherical clusters. The clustering results depicted in  \autoref{t5} (B) demonstrate that our method achieves accurate clustering outcomes. In \autoref{t5} (C), the dataset comprises intersecting U-shaped, T-shaped, and L-shaped clusters, along with some noise around and between clusters. Our method accurately identifies all clusters without being influenced by the noise. In  \autoref{t5} (D), the dataset is composed of O-shaped, C-shaped, S-shaped, bar-shaped, and spherical clusters, with some noise around the clusters as well. As expected, our method successfully identifies all clusters. Extensive and comprehensive experiments on another 11 synthetic datasets further corroborate the effectiveness and applicability of our algorithm. These datasets vary in density, number of clusters, and the mixture of spherical and non-spherical shapes. We also visualize the granularity partition and clustering results for each dataset. The results emphasize the generality, effectiveness, and certain robustness of our algorithm in identifying non-spherical data. At the same time, the article does not reflect but in the algorithm testing phase we verified a conclusion: When the dimension increases, the accuracy of many algorithms will decrease, while the clustering accuracy of the method supported by granular-ball computing is relatively stable when the dimension changes.

% \usepackage{multirow}
\begin{table*}[]
	\centering
	\caption{ACC, NMI and TIME(Second) Comparison on Synthetic Datasets in Python 3}\label{table6}
	\begin{tabular}{ccccccccccc}
		\hline
		Datasets                 &      & KM             & DBSCAN         & DP             & SC             & AC             & HCDC           & GBDP           & GBSC           & GBCT            \\ \hline
		\multirow{3}{*}{A}       & ACC  & 0.470          & 0.995          & 0.705          & 0.552          & \textbf{1.000} & \textbf{1.000} & 0.410          & 0.499          & \textbf{1.000} \\
		& NMI  & 0.503          & 0.980          & 0.684          & 0.641          & \textbf{1.000} & \textbf{1.000} & 0.479          & 0.137          & \textbf{1.000} \\
		& TIME & 0.423          & 10.324         & 0.745          & 10.077         & 45.696         & 6.151          & \textbf{0.173} & 1.015          & 1.768          \\
		\multirow{3}{*}{B}       & ACC  & 0.535          & \textbf{1.000} & 0.812          & 0.453          & \textbf{1.000} & \textbf{1.000} & 0.692          & 0.633          & \textbf{1.000} \\
		& NMI  & 0.694          & 0.999          & 0.822          & 0.650          & \textbf{1.000} & \textbf{1.000} & 0.812          & 0.755          & \textbf{1.000} \\
		& TIME & 2.947          & 194.579        & 16.577         & 369.773        & 3253.452       & 185.315        & \textbf{0.418} & 27.096         & 14.257         \\
		\multirow{3}{*}{C}       & ACC  & 0.620          & \textbf{0.975}          & 0.779          & 0.855          & 0.251          & 0.684          & 0.458          & 0.727          & 0.685 \\
		& NMI  & 0.637          & \textbf{0.912}          & 0.841         & 0.846          & 0.011          & 0.786          & 0.516          & 0.659          & 0.830 \\
		& TIME & 3.156          & 163.753        & 14.528         & 304.810        & 2400.672       & 110.455        & \textbf{0.523} & 24.029         & 12.678         \\
		\multirow{3}{*}{D}       & ACC  & 0.456          & 0.989          & 0.466          & 0.543          & 0.417          & 0.886          & 0.428          & 0.623          & \textbf{1.000} \\
		& NMI  & 0.614          & 0.963          & 0.670          & 0.729          & 0.431          & 0.919          & 0.588          & 0.716          & \textbf{1.000} \\
		& TIME & 3.580          & 144.286        & 12.634         & 250.134        & 1890.292       & 98.716         & \textbf{0.512} & 21.831         & 9.118          \\
		\multirow{3}{*}{E}       & ACC  & 0.812          & 0.939          & 0.864          & \textbf{1.000} & \textbf{1.000} & \textbf{1.000} & 0.610          & 0.769          & \textbf{1.000} \\
		& NMI  & 0.548          & 0.718          & 0.611          & \textbf{1.000} & \textbf{1.000} & \textbf{1.000} & 0.279          & 0.478          & \textbf{1.000} \\
		& TIME & 0.258          & 6.816          & 0.597          & 6.389          & 24.428         & 8.174          & \textbf{0.244} & 0.803          & 0.827          \\
		\multirow{3}{*}{F}       & ACC  & 0.979          & 0.991          & 0.944          & 0.815          & 0.537          & 0.863          & 0.634          & 0.800          & \textbf{1.000} \\
		& NMI  & 0.904          & 0.948          & 0.840          & 0.630          & 0.002          & 0.789          & 0.346          & 0.434          & \textbf{1.000} \\
		& TIME & \textbf{0.199} & 8.875          & 0.693          & 8.273          & 38.779         & 5.880          & 0.249          & 0.997          & 1.184          \\
		\multirow{3}{*}{G}       & ACC  & 0.996          & 0.991          & \textbf{1.000} & 0.996          & 0.482          & 0.999          & 0.767          & 0.868          & \textbf{1.000} \\
		& NMI  & 0.980          & 0.955          & \textbf{1.000} & 0.980          & 0.010          & 0.995          & 0.697          & 0.732          & \textbf{1.000} \\
		& TIME & \textbf{0.104} & 4.084          & 0.324          & 3.256          & 17.037         & 2.262          & 0.210          & 0.545          & 0.909          \\
		\multirow{3}{*}{H}       & ACC  & 0.962          & 0.991          & 0.934          & \textbf{1.000} & \textbf{1.000} & \textbf{1.000} & 0.660          & \textbf{1.000} & \textbf{1.000} \\
		& NMI  & 0.867          & 0.961          & 0.842          & \textbf{1.000} & \textbf{1.000} & \textbf{1.000} & 0.732          & \textbf{1.000} & \textbf{1.000} \\
		& TIME & \textbf{0.020} & 0.605          & 0.024          & 0.148          & 0.332          & 0.264          & 0.122          & 0.030          & 0.195          \\
		\multirow{3}{*}{I}       & ACC  & 0.995          & 0.991          & \textbf{1.000} & 0.999          & 0.751          & 0.997          & 0.996          & 0.999          & \textbf{1.000} \\
		& NMI  & 0.953          & 0.919          & \textbf{1.000} & 0.988          & 0.002          & 0.966          & 0.957          & 0.988          & \textbf{1.000} \\
		& TIME & \textbf{0.039} & 3.734          & 0.343          & 3.079          & 22.221         & 2.251          & 0.229          & 0.671          & 0.947          \\
		\multirow{3}{*}{J}       & ACC  & 0.683          & 0.996          & 0.668          & \textbf{1.000} & 0.750          & 0.999          & 0.631          & \textbf{1.000} & \textbf{1.000} \\
		& NMI  & 0.536          & 0.975          & 0.540          & \textbf{1.000} & 0.700          & 0.994          & 0.641          & \textbf{1.000} & \textbf{1.000} \\
		& TIME & \textbf{0.076} & 3.487          & 0.319          & 3.010          & 10.402         & 2.132          & 0.215          & 0.621          & 0.943          \\
		\multirow{3}{*}{K}       & ACC  & 0.832          & \textbf{1.000} & 0.768          & \textbf{1.000} & \textbf{1.000} & \textbf{1.000} & 0.550          & \textbf{1.000} & \textbf{1.000} \\
		& NMI  & 0.343          & \textbf{1.000} & 0.241          & \textbf{1.000} & \textbf{1.000} & \textbf{1.000} & 0.102          & \textbf{1.000} & \textbf{1.000} \\
		& TIME & \textbf{0.072} & 8.046          & 0.642          & 6.445          & 27.896         & 4.429          & 0.277          & 1.213          & 1.474          \\
		\multirow{3}{*}{L}       & ACC  & 0.335          & 0.999          & 0.498          & 0.556          & \textbf{1.000} & \textbf{1.000} & 0.377          & \textbf{1.000} & \textbf{1.000} \\
		& NMI  & 0.000          & 0.991          & 0.157          & 0.669          & \textbf{1.000} & \textbf{1.000} & 0.095          & \textbf{1.000} & \textbf{1.000} \\
		& TIME & 0.453          & 41.227         & 3.398          & 49.791         & 287.165        & 24.867         & \textbf{0.363} & 7.194          & 4.367          \\
		\multirow{3}{*}{M}       & ACC  & 0.392          & 0.970          & 0.375          & 0.357          & 0.769          & 0.992          & 0.370          & 0.990          & \textbf{1.000} \\
		& NMI  & 0.202          & 0.824          & 0.275          & 0.250          & 0.009          & 0.957          & 0.248          & 0.939          & \textbf{1.000} \\
		& TIME & 0.294          & 6.929          & 0.599          & 6.218          & 24.613         & 4.247          & \textbf{0.248} & 2.621          & 1.248          \\
		\multirow{3}{*}{N}       & ACC  & 0.681          & \textbf{1.000} & 0.613          & 0.583          & \textbf{1.000} & \textbf{1.000} & 0.596          & 0.473          & \textbf{1.000} \\
		& NMI  & 0.568          & \textbf{1.000} & 0.562          & 0.661          & \textbf{1.000} & \textbf{1.000} & 0.541          & 0.123          & \textbf{1.000} \\
		& TIME & \textbf{0.181} & 3.231          & 0.311          & 3.632          & 9.405          & 2.416          & 0.223          & 1.636          & 1.010          \\
		\multirow{3}{*}{O}       & ACC  & \textbf{1.000} & 0.414          & \textbf{1.000} & \textbf{1.000} & 0.358          & \textbf{1.000} & 0.999          & \textbf{1.000} & \textbf{1.000} \\
		& NMI  & \textbf{1.000} & 0.465          & 0.998          & \textbf{1.000} & 0.410          & 0.998          & 0.996          & \textbf{1.000} & \textbf{1.000} \\
		& TIME & 0.879          & 139.068        & 12.408         & 251.947        & 2250.319       & 86.324         & \textbf{0.530} & 25.623         & 8.926          \\
		\multirow{3}{*}{Average} & ACC  & 0.702          & 0.949          & 0.762          & 0.781          & 0.754          & 0.961          & 0.612          & 0.825          & \textbf{0.979} \\
		& NMI  & 0.621          & 0.907          & 0.672          & 0.803          & 0.572          & 0.960          & 0.535          & 0.731          & \textbf{0.989} \\
		& TIME & 1.074          & 49.270         & 4.276          & 85.132         & 686.847        & 36.259         & \textbf{0.302} & 7.728          & 3.990         \\ \hline
	\end{tabular}
\end{table*}

As a comparison, the visualization results of KM, DBSCAN, DP, SC, AC, HCDC, and GBCT will be shown in the Supplementary\_Material.
Even though KM, DBSCAN, DP, SC, AC, HCDC, and  $\rm{GBCT\_old}$ undergo multiple optimizations with parameter tuning, the clustering results often fail to align with visual intuition; clusters that should be connected are often disconnected. In contrast, the clustering results produced by GBCT are noticeably superior and exhibit clear visual intuitiveness. The remarkable performance of GBCT can be attributed to its reliance on the nearest distance across entire clusters for clustering, enabling it to identify complex U-shaped, L-shaped, and T-shaped data patterns. Moreover, GBCT incorporates noise-resistant optimizations during merging, preventing noise from being treated as an independent cluster.

%\begin{table*}[]
%	\centering
%	\caption{Time(Second) Comparison of Different Algorithms in Python 3}\label{table6}
%	\begin{tabular}{cccccccccc}
%		\hline
%		Dataset & Instance & KM    & DBSCAN & DP    & SC      & AC        & HCDC   & GBC\_old & GBC   \\ \hline
%		A       & 1735     & 0.216 & 3.422  & 2.44  & 5.968   & 27.38     & 0.448  & 2.20     & 1.29  \\
%		B       & 7679     & 0.492 & 72.927 & 16.83 & 237.465 & 1,734.232 & 71.686 & 26.94    & 11.22 \\
%		C       & 7200     & 0.482 & 68.192 & 14.74 & 199.327 & 1,416.154 & 10.704 & 22.35    & 6.86  \\
%		D       & 6800     & 0.759 & 56.886 & 11.32 & 158.799 & 1,202.701 & 12.674 & 19.59    & 6.28  \\ \hline
%	\end{tabular}
%\end{table*}

\begin{table*}[]
	\centering
	\caption{Parameters Settings on Synthetic Datasets}\label{table7}
	\begin{tabular}{cccccccccc}
		\hline
		& KM  & DBSCAN             & DP  & SC  & AC  & HCDC & GBDP & GBSC           & GBCT         \\ \hline
		A & k=6 & eps=8,minpts=7     & k=6 & k=6 & k=6 & k=6  & k=6  & k=6,delta=0.1  & k=6 \\
		B & k=7 & eps=0.91,minpts=8  & k=7 & k=7 & k=7 & k=7  & k=7  & k=7,delta=0.1  & k=7 \\
		C & k=6 & eps=14,minpts=24   & k=6 & k=6 & k=6 & k=6  & k=6  & k=6,delta=0.1  & k=6 \\
		D & k=9 & eps=9.91,minpts=10 & k=9 & k=9 & k=9 & k=9  & k=9  & k=9,delta=0.1  & k=9 \\
		E & k=3 & eps=0.05,minpts=10 & k=3 & k=3 & k=3 & k=3  & k=3  & k=3,delta=0.1  & k=3 \\
		F & k=3 & eps=0.89,minpts=9  & k=3 & k=3 & k=3 & k=3  & k=3  & k=3,delta=0.1  & k=3 \\
		G & k=4 & eps=0.55,minpts=18 & k=4 & k=4 & k=4 & k=4  & k=4  & k=4,delta=0.1  & k=4 \\
		H & k=5 & eps=16.8,minpts=3  & k=5 & k=5 & k=5 & k=5  & k=5  & k=3,delta=0.1  & k=5 \\
		I & k=2 & eps=5.5,minpts=14  & k=2 & k=2 & k=2 & k=2  & k=2  & k=2,delta=0.1  & k=2 \\
		J & k=3 & eps=0.8,minpts=14  & k=3 & k=3 & k=3 & k=3  & k=3  & k=3,delta=0.1  & k=3 \\
		K & k=2 & eps=0.15,minpts=13 & k=2 & k=2 & k=2 & k=2  & k=2  & k=2,delta=0.1  & k=2 \\
		L & k=3 & eps=0.19,minpts=12 & k=3 & k=3 & k=3 & k=3  & k=3  & k=3,delta=0.05 & k=3 \\
		M & k=4 & eps=0.19,minpts=12 & k=4 & k=4 & k=4 & k=4  & k=4  & k=4,delta=0.05 & k=4 \\
		N & k=4 & eps=10,minpts=3    & k=4 & k=4 & k=4 & k=4  & k=4  & k=4,delta=0.05 & k=4 \\
		O & k=5 & eps=1.09,minpts=24 & k=5 & k=5 & k=5 & k=5  & k=5  & k=5,delta=0.1  & k=5 \\ \hline
	\end{tabular}
\end{table*}

The KM algorithm, while the fastest, is limited by its strategy of selecting cluster centers and then assigning points, making it unable to identify non-spherical clusters and performing poorly on datasets with significantly different cluster counts between clusters.
DBSCAN can detect clusters of arbitrary shapes, but it is parameter sensitive and has high time complexity, which is not as efficient as GBCT.
DP can discover clusters of arbitrary shapes but is relatively time-consuming in terms of computing and selecting cluster centers, with room for improvement in precision.
SC can identify clustering structures in non-convex datasets but requires the computation of similarity matrices and Laplacian matrices for all data points, resulting in high computational complexity. Additionally, its performance deteriorates with increasing data dimensionality, making it unsuitable for high-dimensional data.
AC has a high time complexity of $O(n^3)$, is sensitive to noise and outliers, and cannot effectively identify clusters in complex datasets.
HCDC and  $\rm{GBCT\_old}$  have some effectiveness in identifying non-spherical datasets, but they are less efficient.

% \usepackage{multirow}
\begin{table*}[]
	\centering
	\caption{ACC and NMI Comparison on Real Datasets}\label{table8}
	\begin{tabular}{ccccccccccc}
		\hline
		Datasets                  &     & KM    & DBSCAN & DP             & SC             & AC    & HCDC  & GBDP  & GBSC  & GBCT            \\ \hline
		\multirow{2}{*}{cell}     & ACC & 0.803 & 0.800  & \textbf{0.873} & 0.413          & 0.794 & 0.814 & 0.743 & 0.374 & 0.834          \\
		& NMI & 0.885 & 0.880  & 0.923          & 0.640          & 0.893 & 0.789 & 0.876 & 0.476 & \textbf{0.929} \\
		\multirow{2}{*}{soybean}  & ACC & 0.493 & 0.843  & 0.538          & \textbf{0.953} & 0.469 & 0.708 & 0.494 & 0.848 & 0.921          \\
		& NMI & 0.085 & 0.005  & 0.116          & \textbf{0.676} & 0.146 & 0.041 & 0.143 & 0.000 & 0.541          \\
		\multirow{2}{*}{mushroom} & ACC & 0.629 & 0.562  & \textbf{0.718} & 0.533          & 0.636 & 0.522 & 0.707 & 0.517 & 0.576          \\
		& NMI & 0.053 & 0.171  & \textbf{0.237} & 0.003          & 0.061 & 0.009 & 0.194 & 0.000 & 0.133          \\
		\multirow{2}{*}{average}  & ACC & 0.624 & 0.735  & 0.710          & 0.633          & 0.633 & 0.681 & 0.648 & 0.580 & \textbf{0.777} \\
		& NMI & 0.341 & 0.352  & 0.425          & 0.440          & 0.367 & 0.280 & 0.404 & 0.159 & \textbf{0.534} \\ \hline
	\end{tabular}
\end{table*}

%Based on the analysis above, GBC, as a fundamental algorithm, can identify complex datasets. To further validate the efficiency of the algorithm, GBC is compared with baseline methods with faster clustering times. \autoref{table6}  shows the time comparison of all algorithms and GBC on four representative datasets. For fairness, all algorithms were implemented in Python 3 and run on the same computer. It is important to note that GBC algorithm is not optimized based on any existing clustering algorithms. Therefore, in the time comparison, all compared algorithms use relatively basic versions. The results in \autoref{table6}  demonstrate that GBC algorithm significantly outperforms DP and  $\rm{GBC\_old}$  algorithms in clustering speed, with processing speeds approximately twice as fast as both. Additionally, from the data in \autoref{table6} , it can be observed that apart from the time performance of HCDC being close to GBC on dataset A, GBC's performance on other datasets is significantly better than other algorithms. Particularly on dataset B and dataset C, the speed improvement of GBC compared to AC algorithm reaches an astonishing hundredfold. This consistent performance across multiple datasets confirms that GBC algorithm not only outperforms widely used basic algorithms in clustering effectiveness but also achieves a satisfactory level of efficiency in terms of time.

In conclusion, GBCT can identify non-spherical complex data with certain noise resistance, attributable to its unique granular-ball generation and splitting strategy, enabling it to fit data distributions and achieve effective clustering purposes. It is noteworthy that while GBCT achieves high precision, it does incur a certain degree of time cost. However, GBCT significantly reduces time costs by utilizing granular-balls of different sizes to fit data and calculating distances between granular-balls instead of between points, as in baseline algorithms. This innovation enables GBCT algorithm to strike a good balance between performance and efficiency.
\\

\subsubsection{Clustering on Real Datasets}
To demonstrate the effectiveness of the GBCT algorithm, experiments were also conducted on real datasets. The clustering results on real datasets were quantitatively evaluated using ACC (Accuracy) and NMI (Normalized Mutual Information). Both ACC and NMI are clustering evaluation metrics, but they have different interpretations. ACC represents the proportion of correctly classified objects in the clustering results to the total number of objects, while NMI measures the similarity between the clustering results and the true class labels.
From \autoref{table8} , it can be observed that GBCT achieves the highest average ACC and NMI, proving the effectiveness of the algorithm.\\

\subsubsection{Clustering on Overlapping Datasets}
\begin{figure}[htp]
	\centering
	
	% 第一行
	\begin{subfigure}[b]{0.49\linewidth}
		\includegraphics[width=\textwidth]{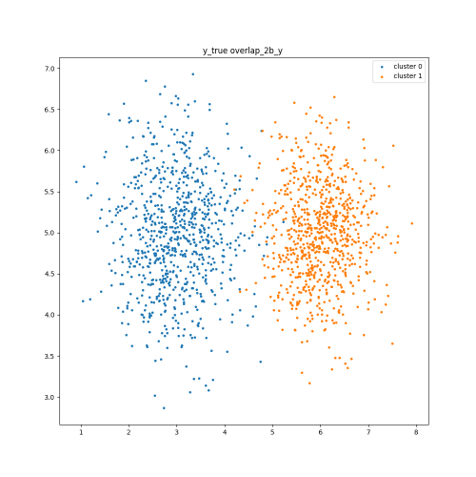}
		\caption{}
		\label{fig:A}
	\end{subfigure}
	% \hfill % 这个命令在子图之间添加了一些水平空间
	% 重复上面的代码块，更改图片路径、caption和label来添加其他图片
	\begin{subfigure}[b]{0.49\linewidth}
		\includegraphics[width=\textwidth]{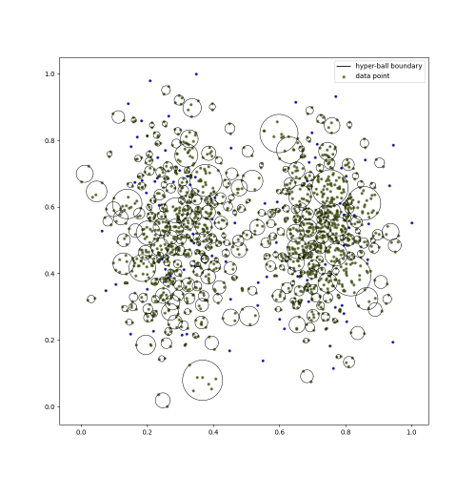}
		\caption{}
		\label{fig:A}
	\end{subfigure}
	% \hfill % 这个命令在子图之间添加了一些水平空间
	\begin{subfigure}[b]{0.49\linewidth}
		\includegraphics[width=\textwidth]{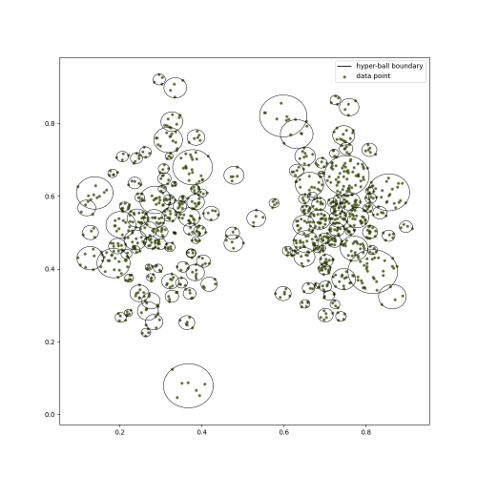}
		\caption{}
		\label{fig:A}
	\end{subfigure}
	% \hfill % 这个命令在子图之间添加了一些水平空间
	\begin{subfigure}[b]{0.49\linewidth}
		\includegraphics[width=\textwidth]{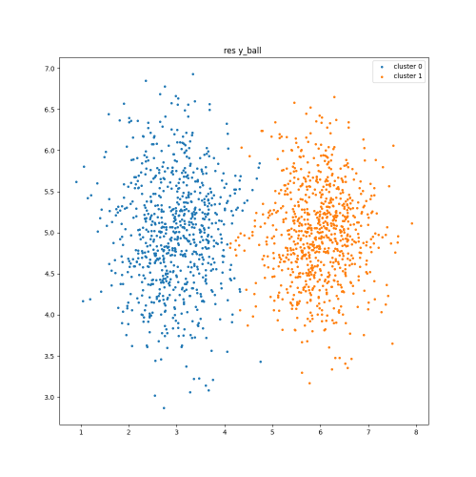}
		\caption{}
		\label{fig:A}
	\end{subfigure}
	
	\caption{The Process of GBCT Identifying Overlapping Datasets}
	\label{t6}
\end{figure}

The GBCT algorithm merges granular-balls based on the distance between the nearest sample points within each granular-ball and can utilize the features of the granular-balls for noise reduction. These two steps ensure that granular clustering can identify overlapping datasets. As illustrated in the \autoref{t6} (a), there is no clear boundary between the two clusters, with the two categories of data even overlapping and intersecting. \autoref{t6} (b) and (c), demonstrate how GBCT processes such datasets. \autoref{t6} (a) displays the original distribution of the overlapping dataset, \autoref{t6} (b) shows the result after granular-ball fitting of the dataset, \autoref{t6} (c) presents the outcome after removing noise granular-balls, and \autoref{t6} (d) depicts the prediction results of the GBCT algorithm.

\begin{figure}[htp]
	\centering
	
	% 第一行
	\begin{subfigure}[b]{0.49\linewidth}
		\includegraphics[width=\textwidth]{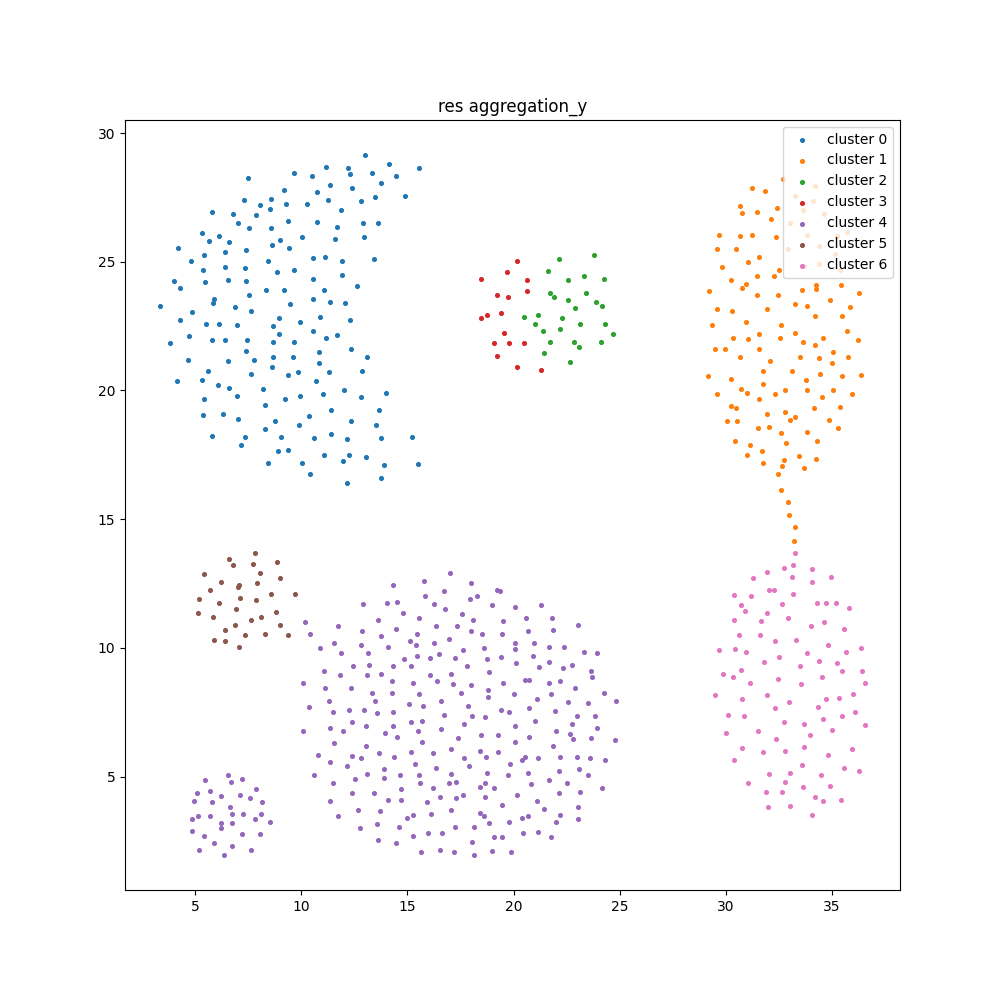}
		\caption{overlap1}
		\label{fig:A}
	\end{subfigure}
	% \hfill % 这个命令在子图之间添加了一些水平空间
	% 重复上面的代码块，更改图片路径、caption和label来添加其他图片
	\begin{subfigure}[b]{0.49\linewidth}
		\includegraphics[width=\textwidth]{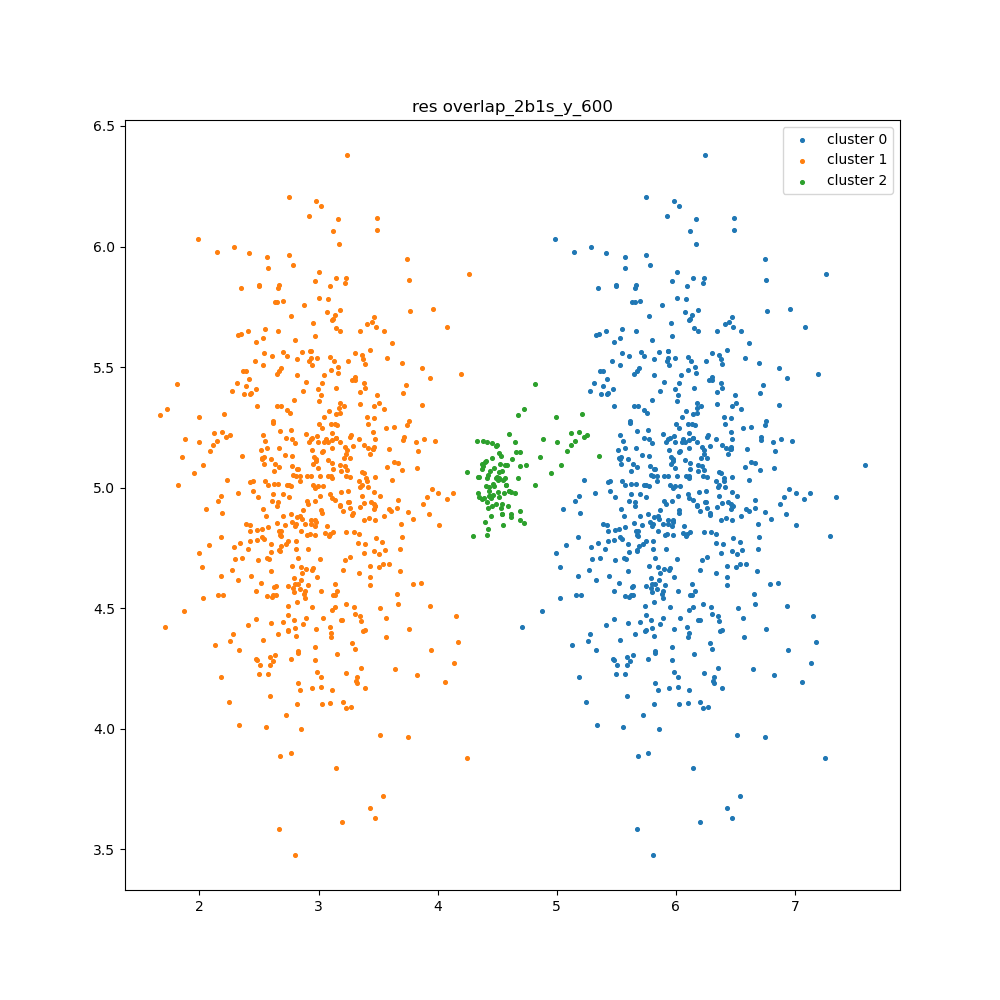}
		\caption{overlap2}
		\label{fig:A}
	\end{subfigure}
	% \hfill % 这个命令在子图之间添加了一些水平空间
	\begin{subfigure}[b]{0.49\linewidth}
		\includegraphics[width=\textwidth]{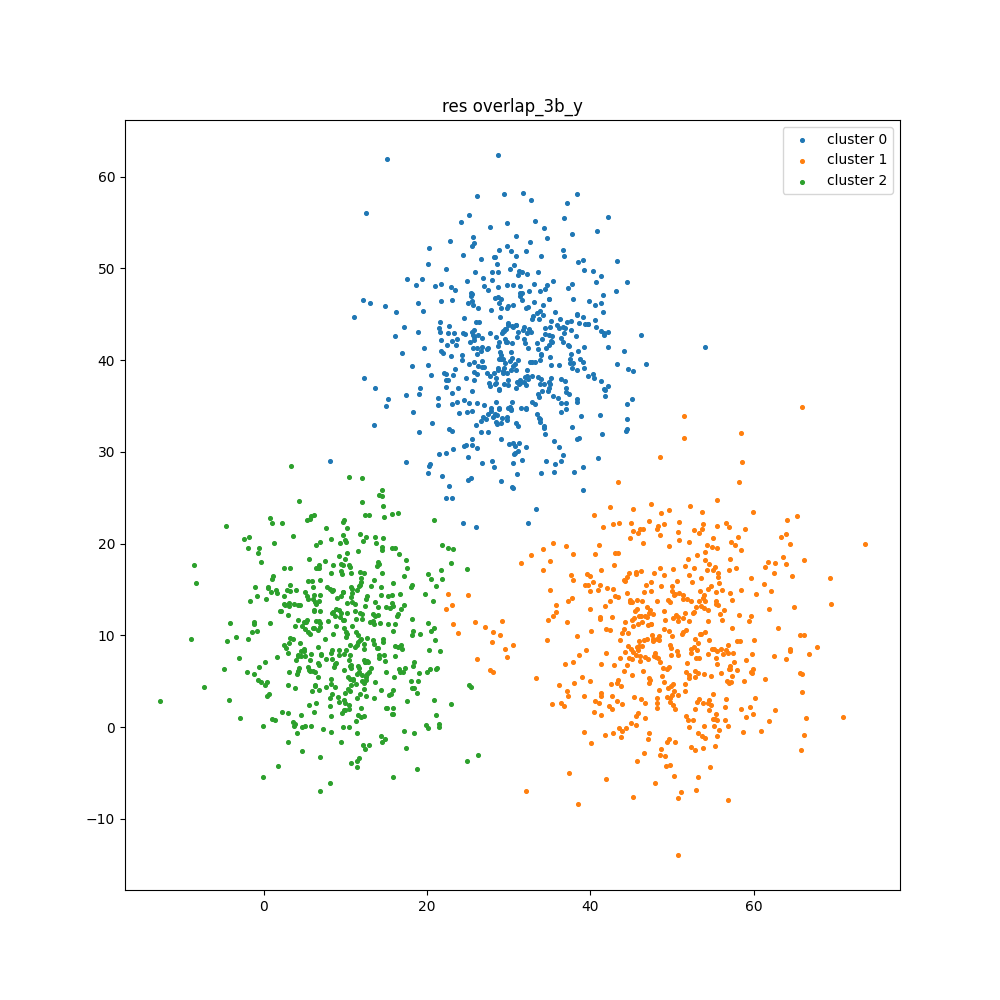}
		\caption{overlap3}
		\label{fig:A}
	\end{subfigure}
	% \hfill % 这个命令在子图之间添加了一些水平空间
	\begin{subfigure}[b]{0.49\linewidth}
		\includegraphics[width=\textwidth]{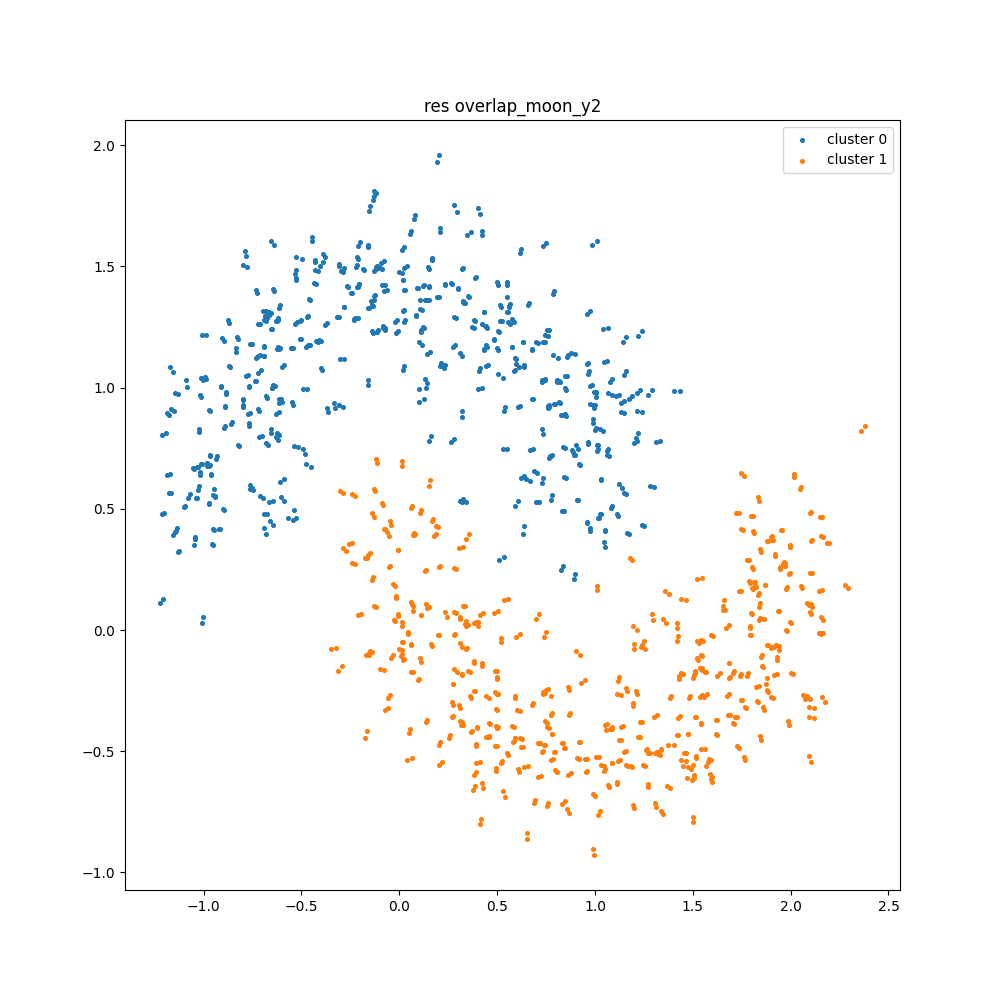}
		\caption{overlap4}
		\label{fig:A}
	\end{subfigure}
	
	\caption{The Performance of GBCT on Overlapping Datasets}
	\label{t7}
\end{figure}

To validate the universality of GBCT in solving overlapping datasets, we utilized four datasets to demonstrate that GBCT can effectively address clustering problems on overlapping datasets. As shown in \autoref{t7}, \autoref{t7} (a) contains four clusters with slight connections, \autoref{t7} (b) features three clusters with significant differences in cluster sizes, \autoref{t7} (c) includes three clusters with overlapping boundaries, and \autoref{t7} (d) represents an overlapping moon-shaped dataset. \autoref{table9} displays the ACC (Accuracy) and NMI (Normalized Mutual Information) of GBCT and all comparison algorithms on four overlapping datasets. The results indicate that GBCT is well-suited to handling overlapping datasets efficiently.

% \usepackage{multirow}
\begin{table*}[]
	\centering
	\caption{ACC and NMI Comparison on Overlapping Datasets}\label{table9}
	\begin{tabular}{cclccllcccc}
		\hline
		Datasets                  &     & \multicolumn{1}{c}{KM} & DBSCAN & DP             & \multicolumn{1}{c}{SC} & \multicolumn{1}{c}{AC} & HCDC  & GBDP           & GBSC  & GBCT            \\ \hline
		\multirow{2}{*}{overlap1} & ACC & 0.857                  & 0.983  & \textbf{0.997}          & 0.792                  & 0.845                  & 0.994 & 0.986          & 0.583 & 0.935 \\
		& NMI & 0.831                  & 0.953  & \textbf{0.992}          & 0.826                  & 0.906                  & 0.985 & 0.964          & 0.395 & 0.947 \\
		\multirow{2}{*}{overlap2} & ACC & 0.718                  & 0.956  & 0.989          & 0.751                  & 0.767                  & 0.944 & \textbf{0.993} & 0.476 & 0.991          \\
		& NMI & 0.692                  & 0.805  & 0.945          & 0.743                  & 0.722                  & 0.848 & \textbf{0.962} & 0.000 & 0.949          \\
		\multirow{2}{*}{overlap3} & ACC & \textbf{0.986}         & 0.968  & 0.984          & \textbf{0.986}         & \textbf{0.986}         & 0.651 & 0.982          & 0.984 & \textbf{0.986} \\
		& NMI & \textbf{0.927}         & 0.868  & 0.920          & \textbf{0.927}         & \textbf{0.927}         & 0.653 & 0.912          & 0.920 & \textbf{0.927} \\
		\multirow{2}{*}{overlap4} & ACC & 0.893                  & 0.955  & \textbf{0.995} & 0.931                  & 0.829                  & 0.504 & 0.787          & 0.984 & 0.993          \\
		& MMI & 0.509                  & 0.778  & \textbf{0.964} & 0.703                  & 0.47                   & 0.008 & 0.389          & 0.884 & 0.946          \\ \hline
	\end{tabular}
\end{table*}

\subsection{Efficiency}

GBCT is based on granular-ball computing, processing objects that are granular-ball rather than individual samples. Because GBCT processes granularity, it can compress originally large-scale datasets into relatively small, granulated datasets. We demonstrate the efficiency of GBCT by demonstrating its runtime performance on synthetic datasets. Since DBSCAN, SC, AC, and HCDC are significantly different from KM, DP, GBDP, and GBSC in running time, we take the running time of the running algorithm of each algorithm as the logarithm base 10 to facilitate display and comparison. At the same time, the time of the data set whose running time exceeds 600(s) is adjusted to 600(s).

\begin{figure}[htp]
	\centering	
	% 第一行
	\vspace{-0.4cm}
	\begin{subfigure}[b]{0.9\linewidth}
		\includegraphics[width=\textwidth]{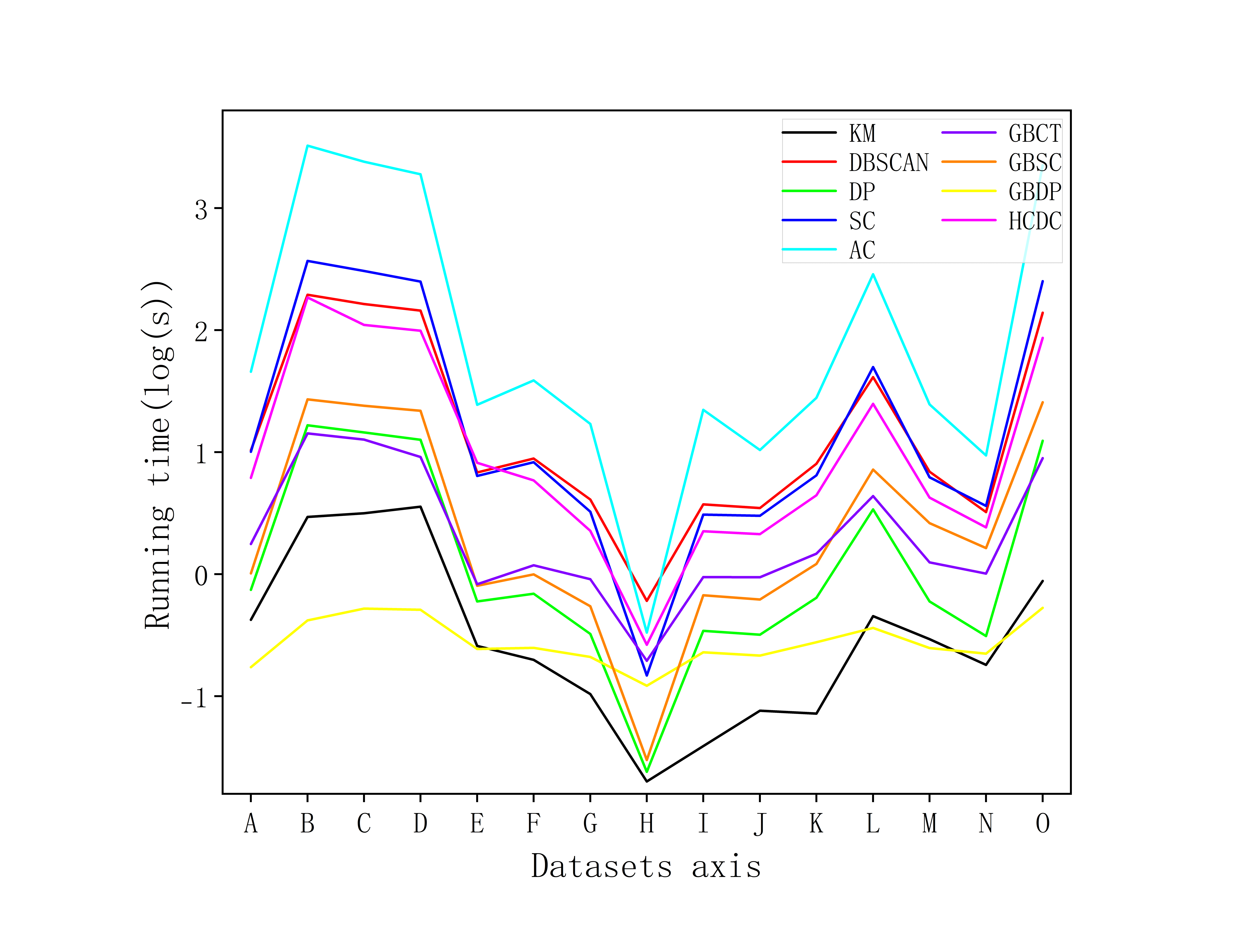}
		\label{fig:A}
	\end{subfigure}	
    \vspace{-0.4cm}
	\caption{Comparison of Time (seconds) between GBCT and DBSCAN, SC, AC, HCDC, KM, DP, GBDP, GBSC on Synthetic Datasets}
	\label{fig_larger}
\end{figure}

The comparison data is shown in \autoref{fig_larger}, and it can be seen from \autoref{fig_larger} that the efficiency of GBCT is much higher than that of AC, SC, DBSCAN, and HCDC, and even more than one hundred times ahead. Moreover, GBCT is ahead of GBSC and DP, and second only to KM and GBDP. From the experimental results, it can be seen that GBCT is more effective for granular-balls than single sample points.

\vspace{-0.4cm}

\subsection{Robustness}
We demonstrate the robustness of GBCT by its performance on noisy datasets. Unlike traditional clustering models that are based on the calculation of individual sample points, granular clustering operates on the basis of granular-balls. Each granular-ball contains multiple sample points, and the distribution of these points inherently reflects the characteristics of the granular-ball. We can compute based on the features of the granular-balls, inferring which are noise and which are samples.

\begin{figure}[htp]
    \centering
    \includegraphics[width=\linewidth]{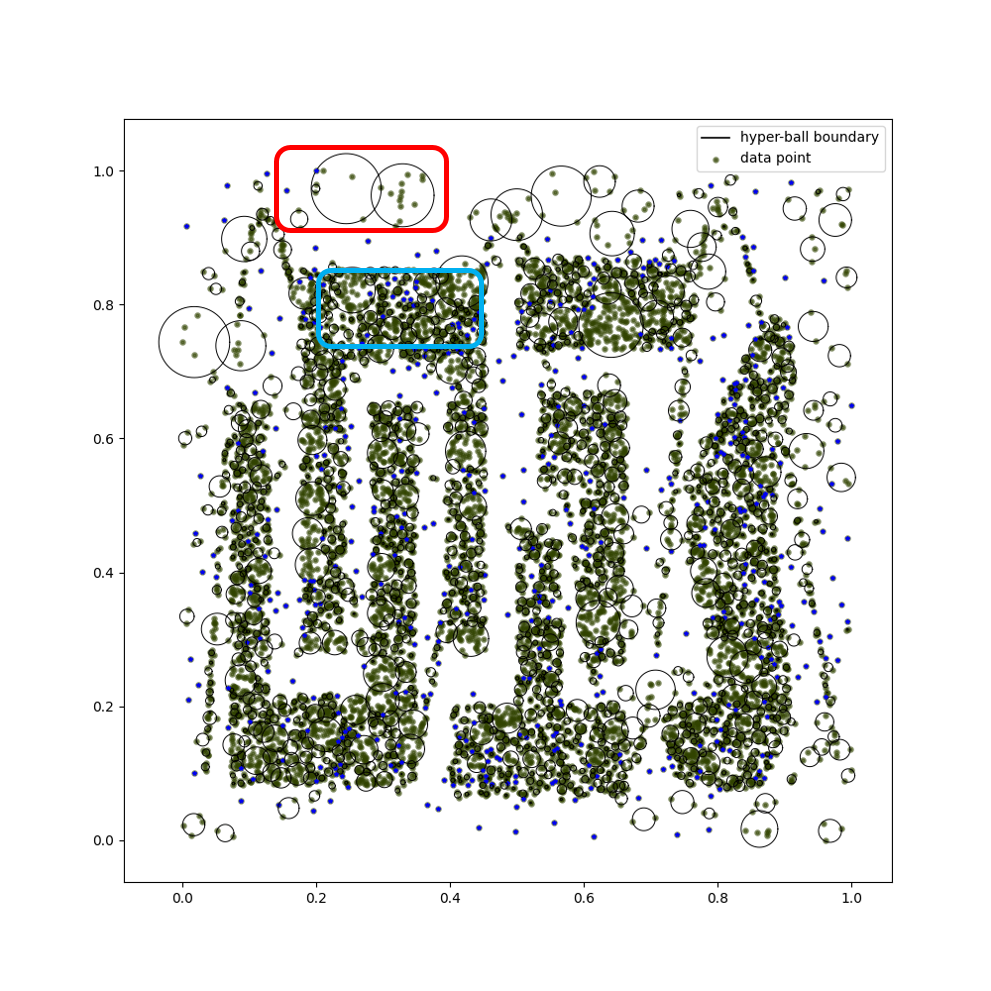}
    \caption{The Principle of GBCT's Robustness}
    \label{t8}
\end{figure}

\autoref{t8} displays the data distribution after granular-ball fitting. In the core area of the dataset (indicated by the blue box), the granular-balls fitted with samples are very dense, whereas in the peripheral regions of the dataset (indicated by the red box), the granular-balls are quite sparse. Specifically, we use the number of samples inside the granular-ball divided by the maximum radius of the granular-ball as a measure of whether the granular-ball is a noisy granular-ball. Based on this characteristic, we can conveniently identify noise granular-balls, which do not participate in the clustering process. Once the clustering is complete, noise granular-balls are assigned to the nearest cluster.

\begin{figure}[htp]
	\centering
	% 第一行
	\begin{subfigure}[b]{0.32\linewidth}
		\includegraphics[width=\textwidth]{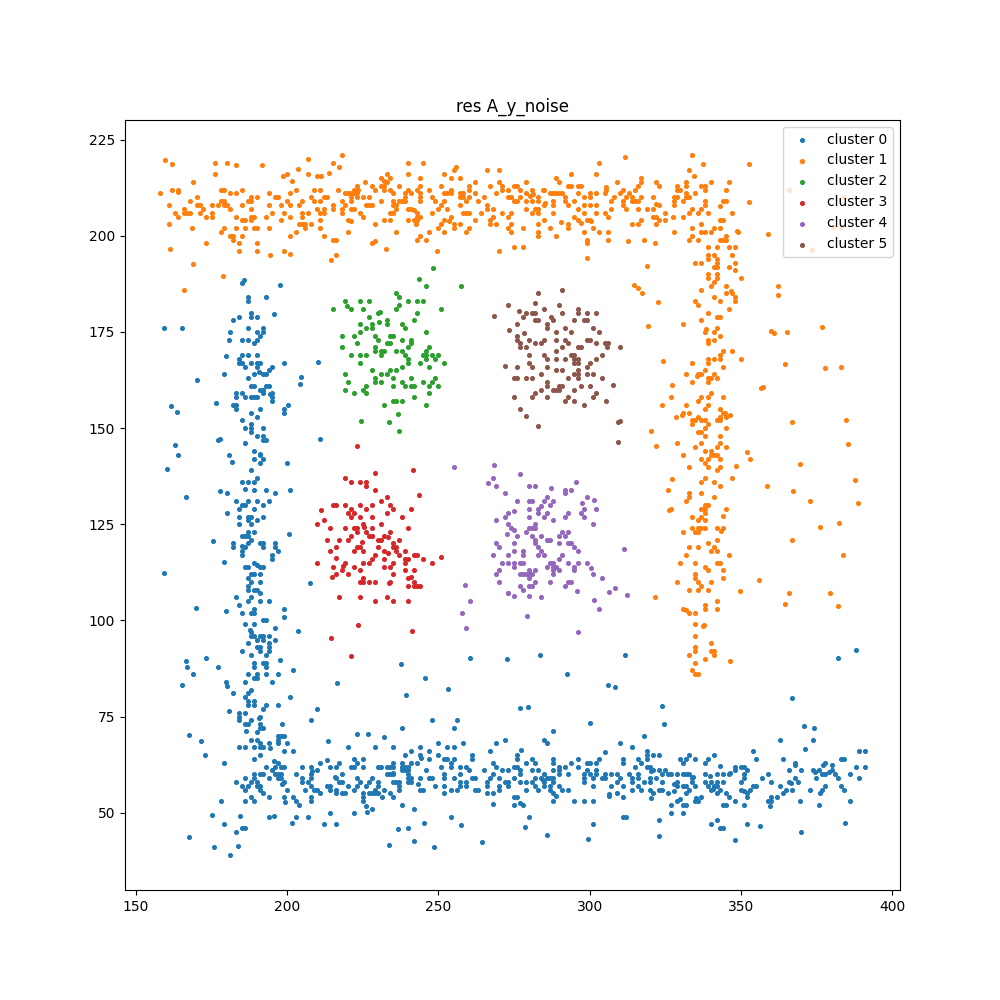}
		\caption{noise1}
		\label{fig:A}
	\end{subfigure}
	% \hfill % 这个命令在子图之间添加了一些水平空间
	% 重复上面的代码块，更改图片路径、caption和label来添加其他图片
	\begin{subfigure}[b]{0.32\linewidth}
		\includegraphics[width=\textwidth]{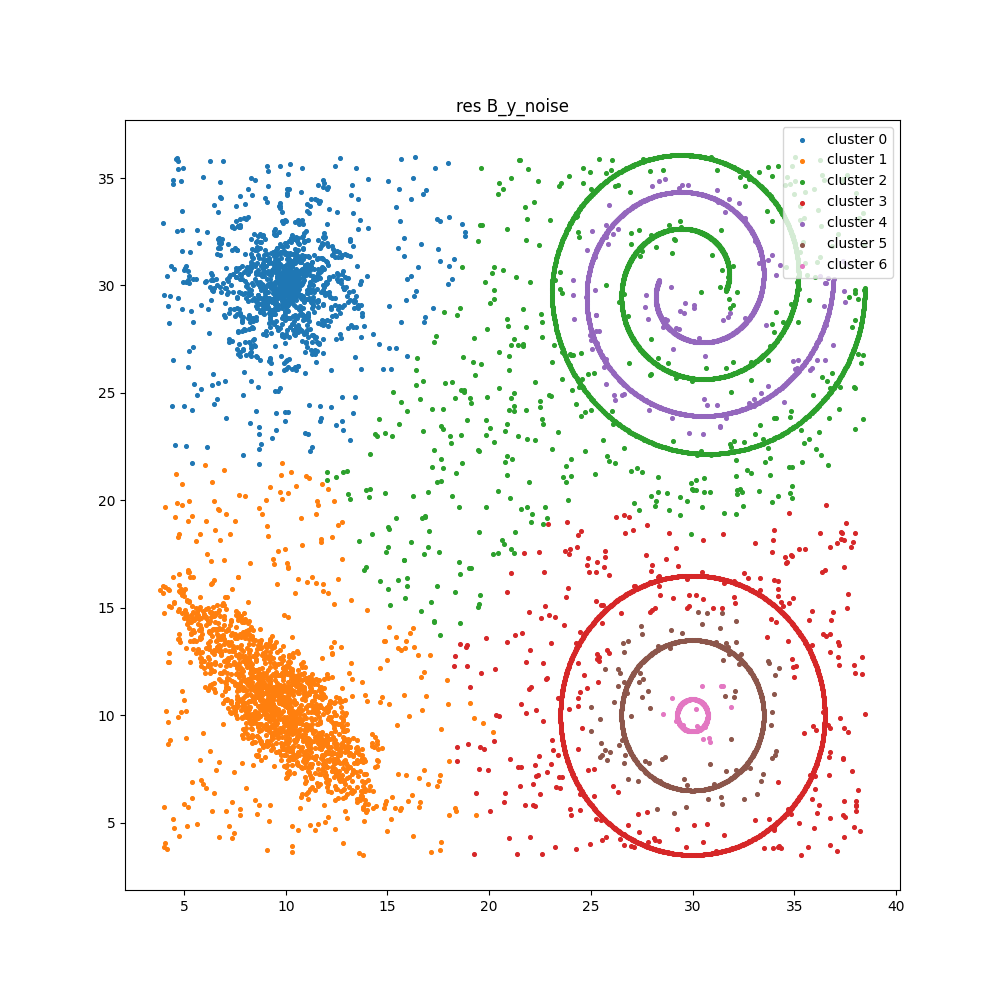}
		\caption{noise2}
		\label{fig:A}
	\end{subfigure}
	% \hfill % 这个命令在子图之间添加了一些水平空间
	\begin{subfigure}[b]{0.32\linewidth}
		\includegraphics[width=\textwidth]{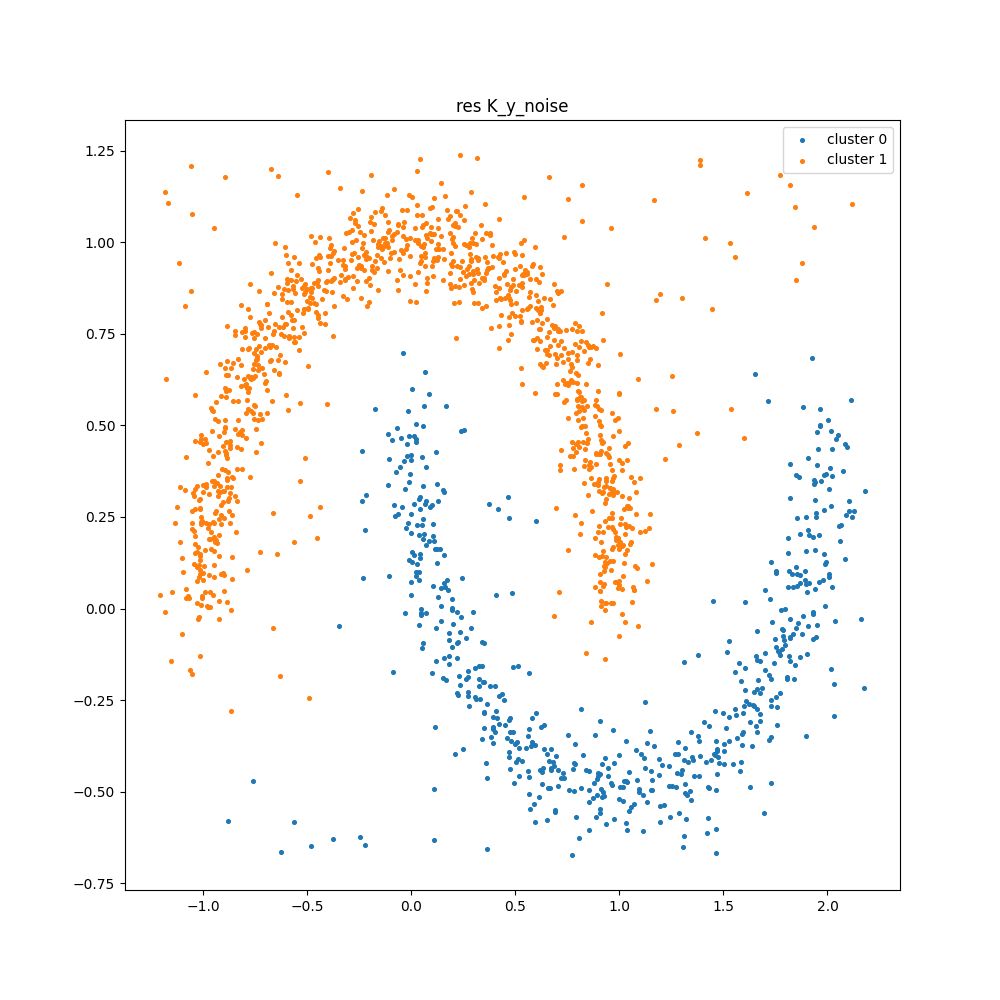}
		\caption{noise3}
		\label{fig:A}
	\end{subfigure}
	% \hfill % 这个命令在子图之间添加了一些水平空间
	
	% 第一行
	\begin{subfigure}[b]{0.32\linewidth}
		\includegraphics[width=\textwidth]{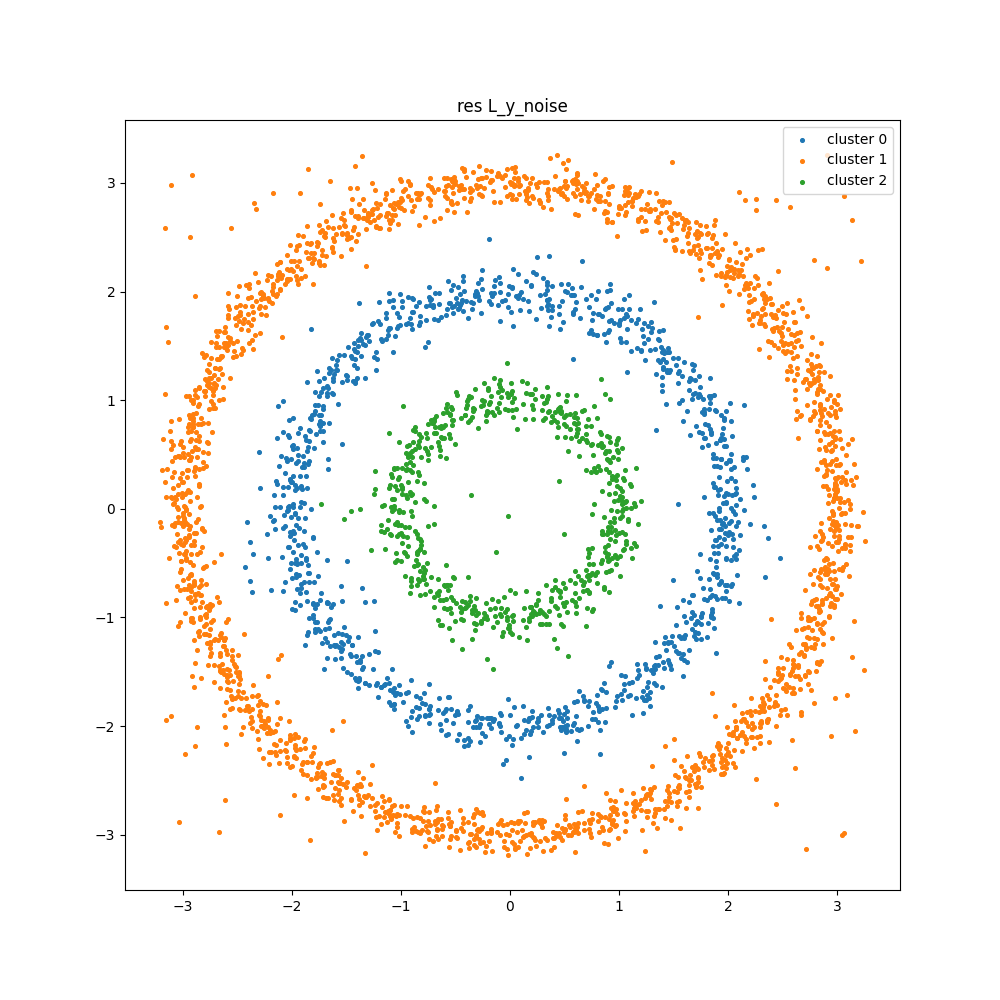}
		\caption{noise4}
		\label{fig:A}
	\end{subfigure}
	% \hfill % 这个命令在子图之间添加了一些水平空间
	% 重复上面的代码块，更改图片路径、caption和label来添加其他图片
	\begin{subfigure}[b]{0.32\linewidth}
		\includegraphics[width=\textwidth]{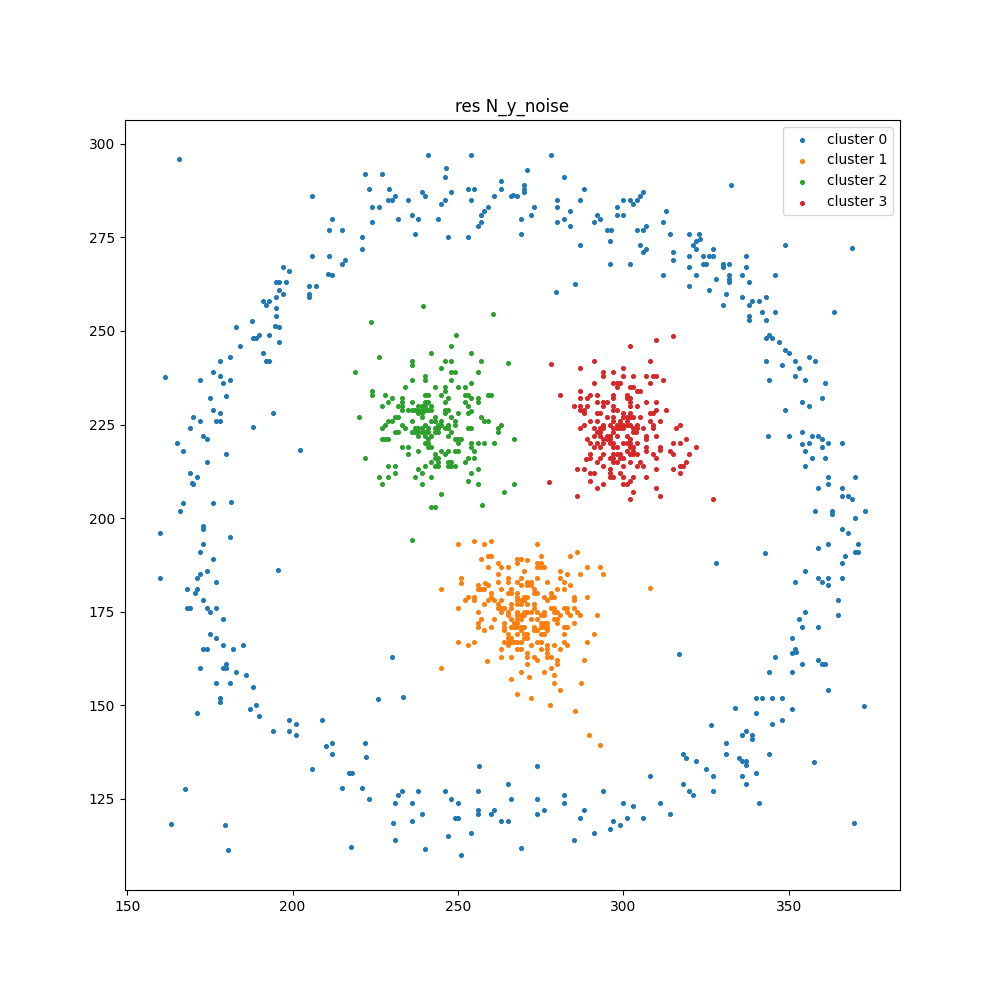}
		\caption{noise5}
		\label{fig:A}
	\end{subfigure}
	% \hfill % 这个命令在子图之间添加了一些水平空间
	\begin{subfigure}[b]{0.32\linewidth}
		\includegraphics[width=\textwidth]{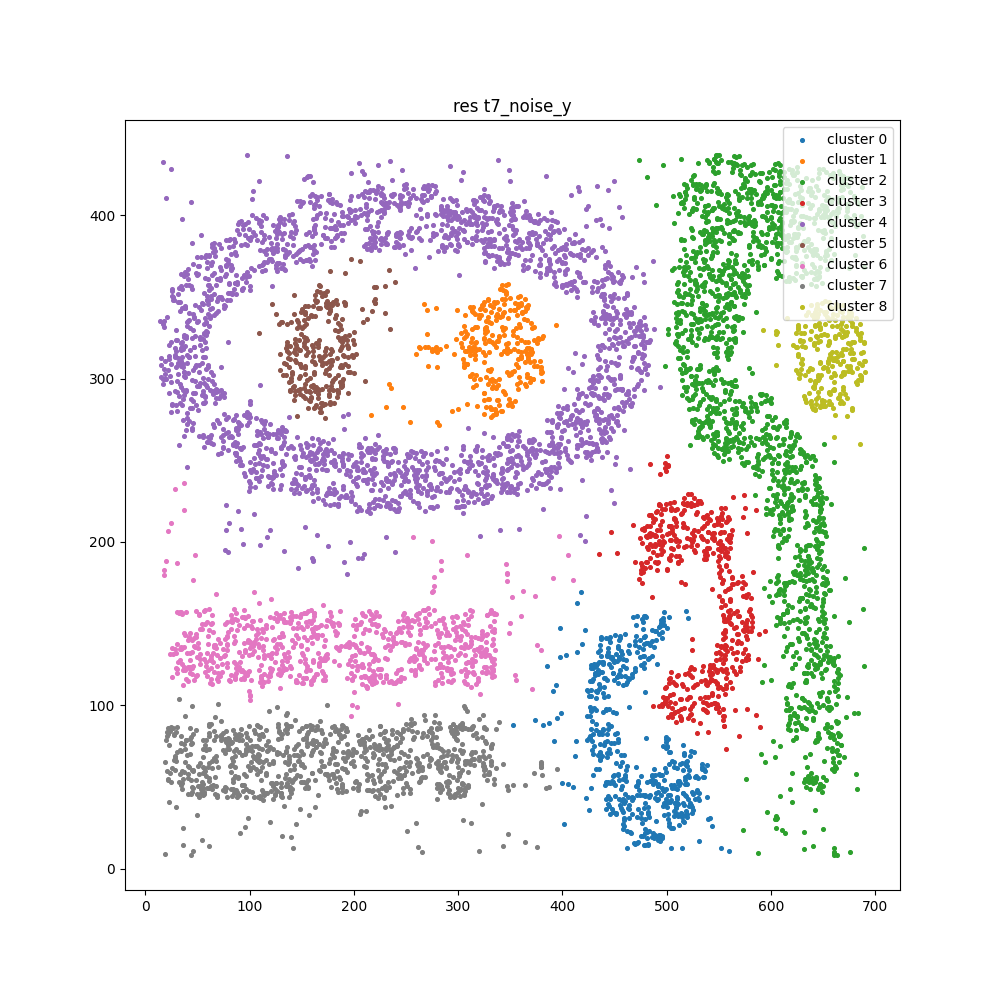}
		\caption{noise6}
		\label{fig:A}
	\end{subfigure}
 
	\caption{The Performance of GBCT on noise Datasets}
	\label{t9}
\end{figure}

To demonstrate the robustness of GBCT, experiments were conducted on six noise datasets \autoref{t9} showcases the performance of GBCT on these noise datasets. \autoref{table10} displays the ACC (Accuracy) and NMI (Normalized Mutual Information) of GBCT and all comparison algorithms on six noise datasets. The results indicate that GBCT possesses strong noise resistance capabilities.

% \usepackage{multirow}
\begin{table*}[]
	\centering
	\caption{ACC and NMI Comparison on Noise Datasets}\label{table10}
	\begin{tabular}{ccccccccccc}
		\hline
		Datasets                &     & KM    & DBSCAN         & DP    & SC    & AC    & HCDC  & GBDP  & GBSC  & GBCT            \\ \hline
		\multirow{2}{*}{noise1} & ACC & 0.488 & 0.896          & 0.629 & 0.437 & 0.515 & 0.998 & 0.515 & 0.600 & \textbf{1.000} \\
		& NMI & 0.542 & 0.801          & 0.642 & 0.590 & 0.564 & 0.992 & 0.515 & 0.347 & \textbf{1.000} \\
		\multirow{2}{*}{noise2} & ACC & 0.595 & 0.761          & 0.672 & 0.558 & 0.607 & 0.796 & 0.467 & 0.597 & \textbf{1.000} \\
		& NMI & 0.699 & 0.789          & 0.770 & 0.589 & 0.699 & 0.893 & 0.577 & 0.638 & \textbf{1.000} \\
		\multirow{2}{*}{noise3} & ACC & 0.830 & 0.844          & 0.770 & 0.830 & 0.750 & 0.665 & 0.793 & 0.885 & \textbf{1.000} \\
		& NMI & 0.338 & 0.649          & 0.223 & 0.362 & 0.201 & 0.000 & 0.418 & 0.482 & \textbf{1.000} \\
		\multirow{2}{*}{noise4} & ACC & 0.335 & 0.796          & 0.474 & 0.347 & 0.445 & 0.554 & 0.426 & 0.342 & \textbf{0.999} \\
		& MMI & 0.000 & 0.662          & 0.235 & 0.001 & 0.081 & 0.000 & 0.085 & 0.000 & \textbf{0.998} \\
		\multirow{2}{*}{noise5} & ACC & 0.581 & \textbf{0.993} & 0.573 & 0.654 & 0.590 & 0.676 & 0.700 & 0.631 & 0.933          \\
		& NMI & 0.586 & \textbf{0.980} & 0.683 & 0.671 & 0.602 & 0.722 & 0.687 & 0.609 & 0.908          \\
		\multirow{2}{*}{noise6} & ACC & 0.435 & \textbf{0.989} & 0.431 & 0.435 & 0.471 & 0.439 & 0.490 & 0.466 & 0.986          \\
		& NMI & 0.551 & \textbf{0.969} & 0.597 & 0.523 & 0.575 & 0.530 & 0.635 & 0.564 & 0.966         \\ \hline
	\end{tabular}
\end{table*}

\subsection{Adaptive GBCT}

In addition to being able to specify the number of clustering clusters, we have also developed a fully adaptive version of GBCT that requires no parameters. The adaptiveness of GBCT is akin to finding the number of clusters using the Elbow Method in KM, but unlike KM, which needs to perform clustering multiple times, GBCT requires only one clustering attempt to determine the number of clusters.

The premise for GBCT to identify the number of clusters is based on the assumption that the distance between samples within a cluster is significantly smaller than the distance between samples from different clusters. When performing clustering, GBCT initially merges granular-balls within clusters. Once all the granular-balls within clusters have been merged, inter-cluster merging begins, at which point the nearest distance between granular-balls across different clusters rapidly increases. This serves as an indication that an appropriate number of clusters has been reached.

\begin{figure}[htp]
	\centering
	% 第一行
	\begin{subfigure}[b]{0.49\linewidth}
		\includegraphics[width=\textwidth]{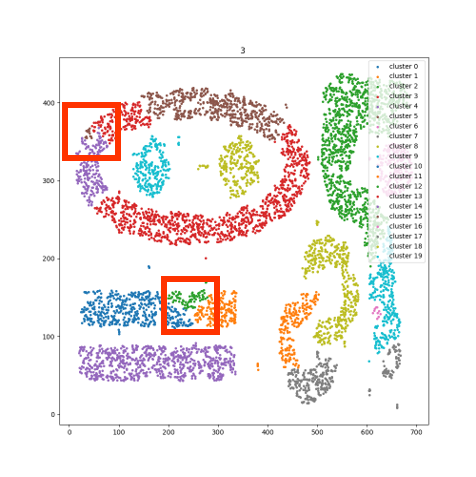}
		\caption{}
		\label{fig:A}
	\end{subfigure}
	% \hfill % 这个命令在子图之间添加了一些水平空间
	% 重复上面的代码块，更改图片路径、caption和label来添加其他图片
	\begin{subfigure}[b]{0.49\linewidth}
		\includegraphics[width=\textwidth]{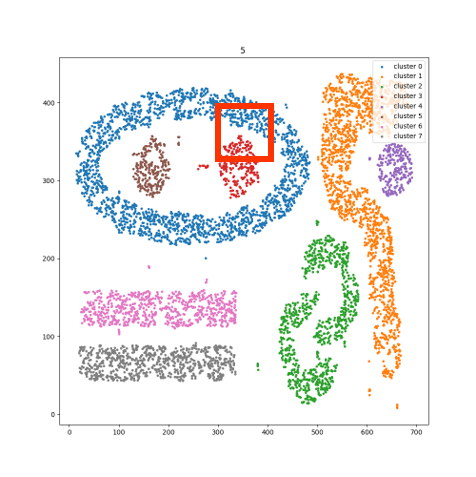}
		\caption{}
		\label{fig:A}
	\end{subfigure}
	% \hfill % 这个命令在子图之间添加了一些水平空间
	\begin{subfigure}[b]{0.49\linewidth}
		\includegraphics[width=\textwidth]{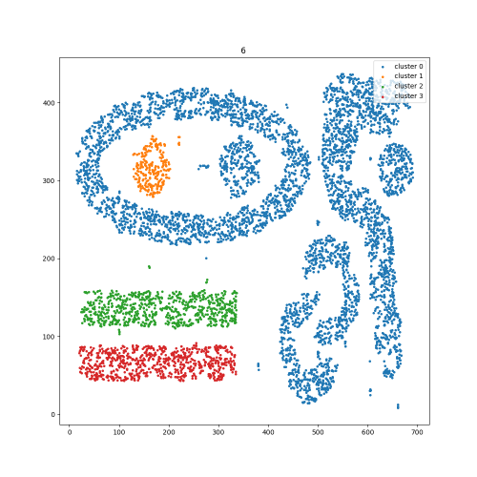}
		\caption{}
		\label{fig:A}
	\end{subfigure}
	% \hfill % 这个命令在子图之间添加了一些水平空间
	\begin{subfigure}[b]{0.49\linewidth}
		\includegraphics[width=\textwidth]{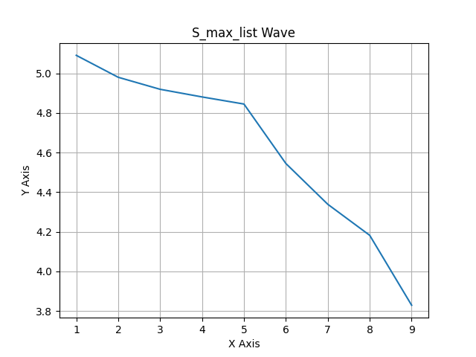}
		\caption{}
		\label{fig:A}
	\end{subfigure}
	
	\caption{The Adaptive Principle of GBCT}
	\label{t10}
\end{figure}

\begin{figure}[htp]
	\centering
	% 第一行
	\begin{subfigure}[b]{0.32\linewidth}
		\includegraphics[width=\textwidth]{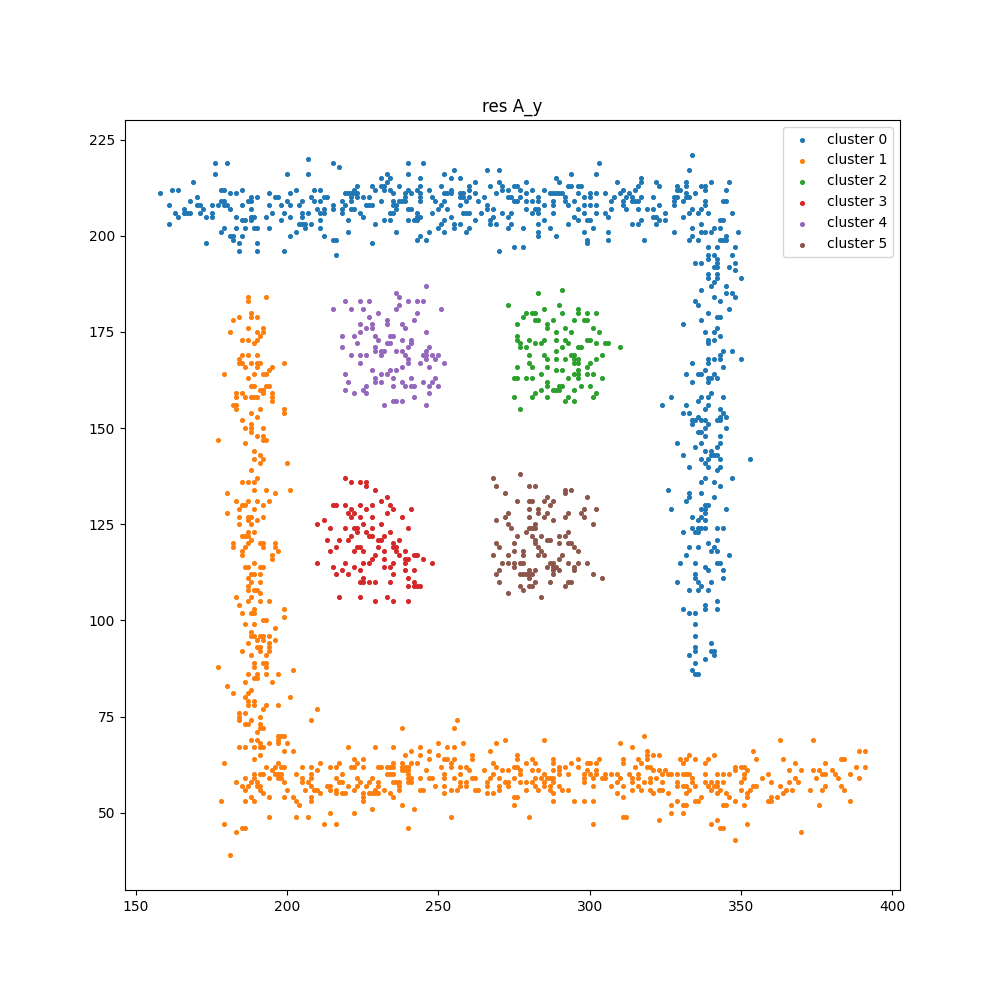 }
		\caption{}
		\label{fig:A}
	\end{subfigure}
	% \hfill % 这个命令在子图之间添加了一些水平空间
	% 重复上面的代码块，更改图片路径、caption和label来添加其他图片
	\begin{subfigure}[b]{0.32\linewidth}
		\includegraphics[width=\textwidth]{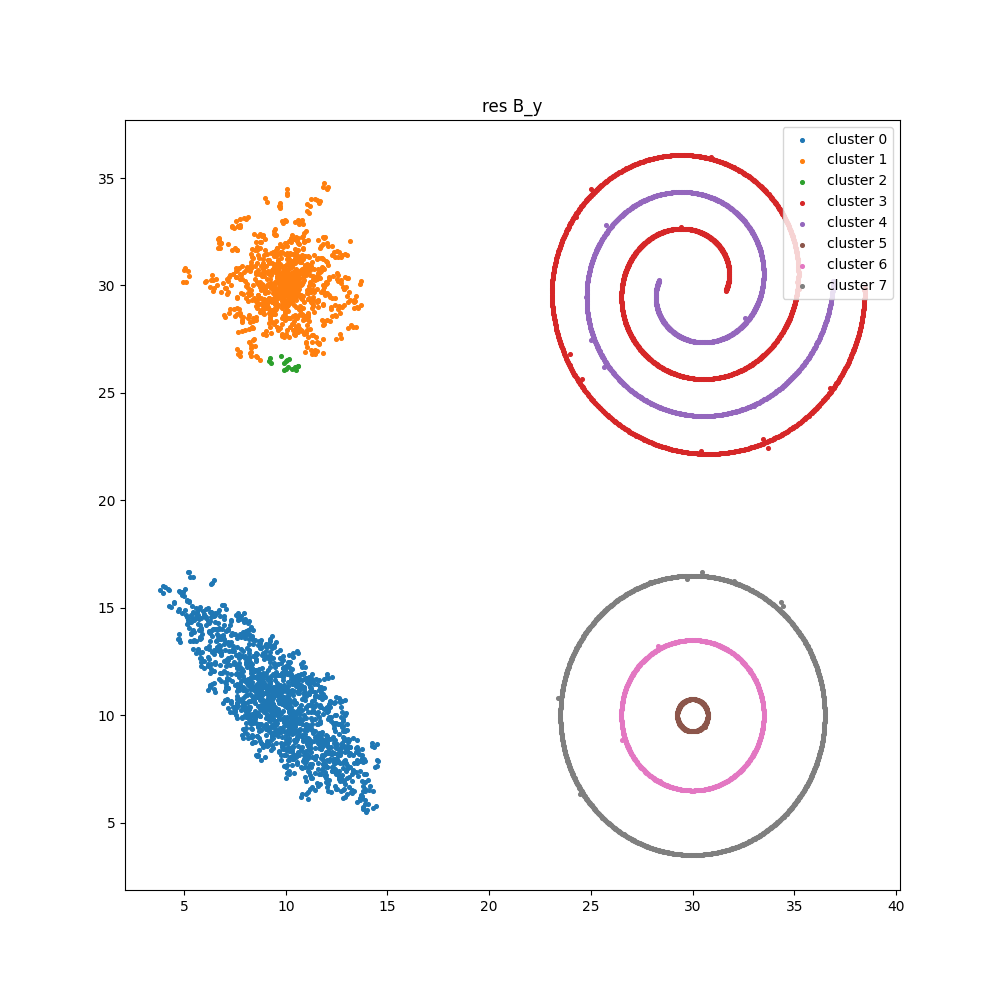 }
		\caption{}
		\label{fig:A}
	\end{subfigure}
	% \hfill % 这个命令在子图之间添加了一些水平空间
	\begin{subfigure}[b]{0.32\linewidth}
		\includegraphics[width=\textwidth]{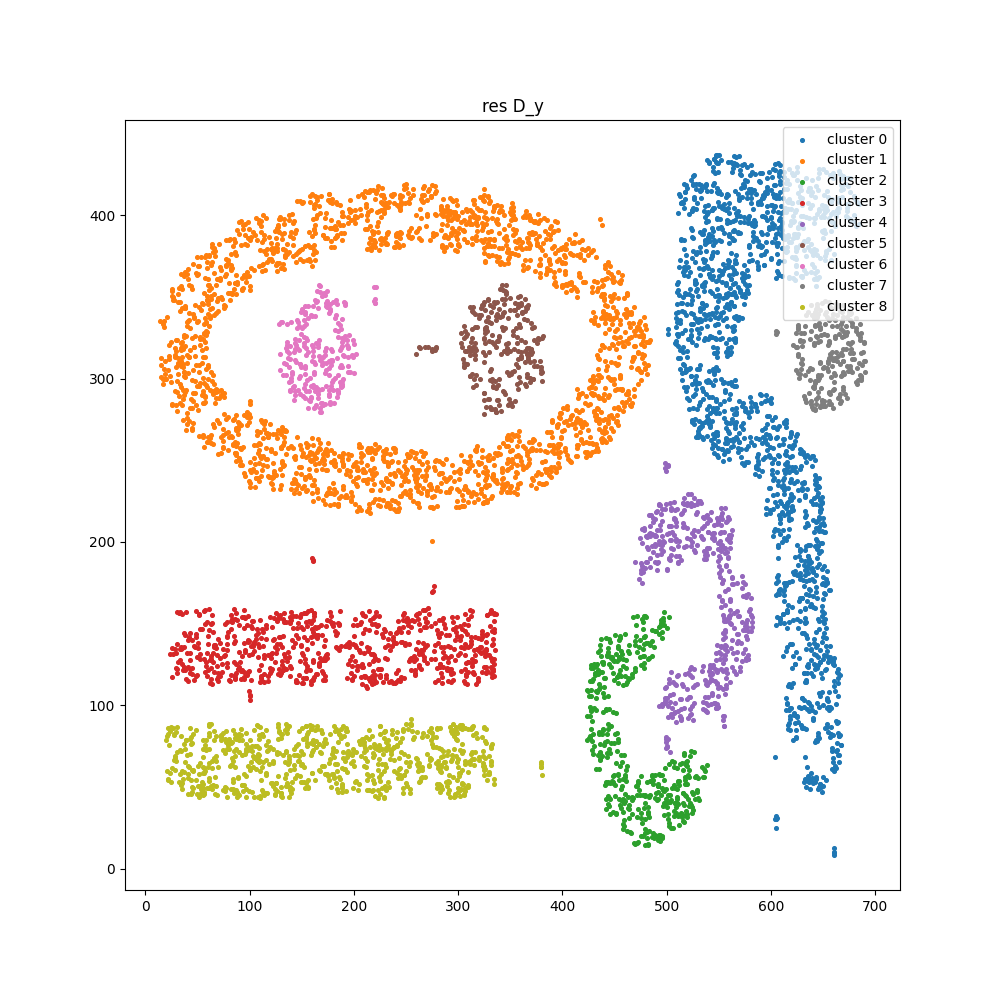 }
		\caption{}
		\label{fig:A}
	\end{subfigure}
	% \hfill % 这个命令在子图之间添加了一些水平空间
	
	% 第一行
	\begin{subfigure}[b]{0.32\linewidth}
		\includegraphics[width=\textwidth]{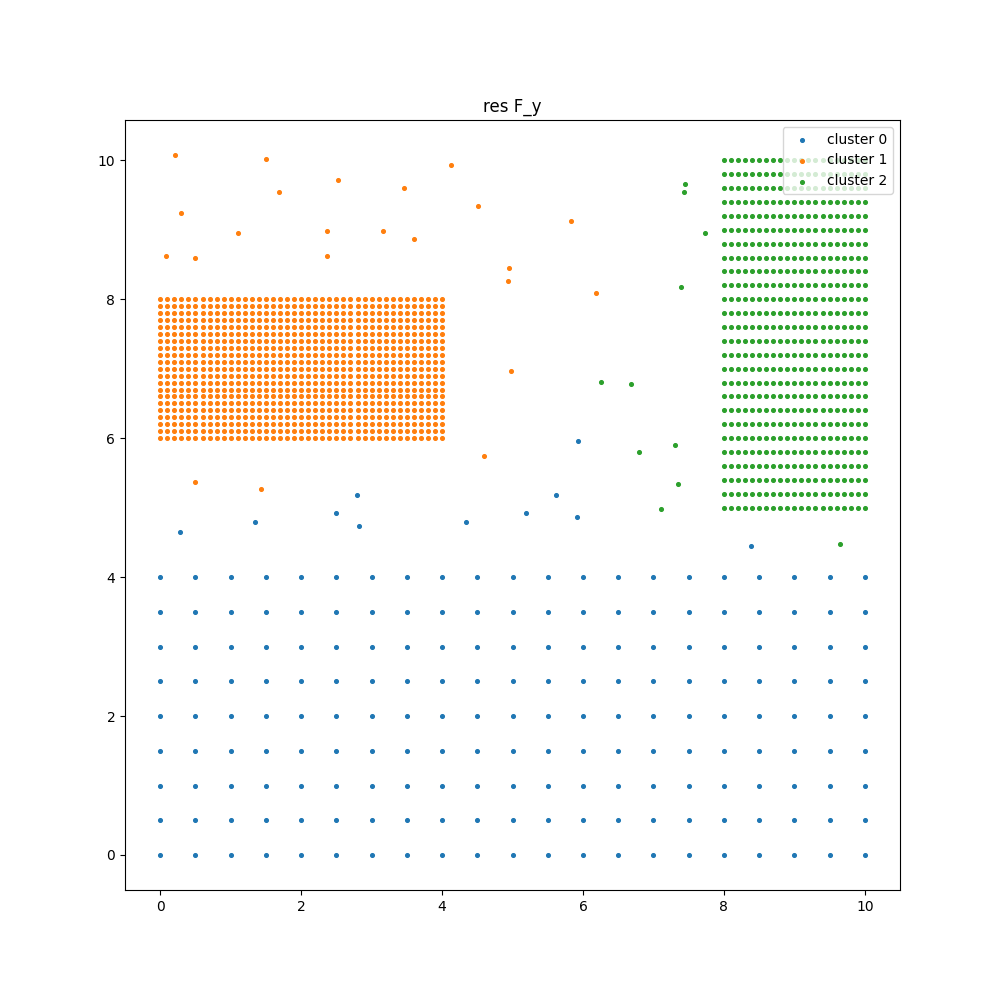 }
		\caption{}
		\label{fig:A}
	\end{subfigure}
	% \hfill % 这个命令在子图之间添加了一些水平空间
	% 重复上面的代码块，更改图片路径、caption和label来添加其他图片
	\begin{subfigure}[b]{0.32\linewidth}
		\includegraphics[width=\textwidth]{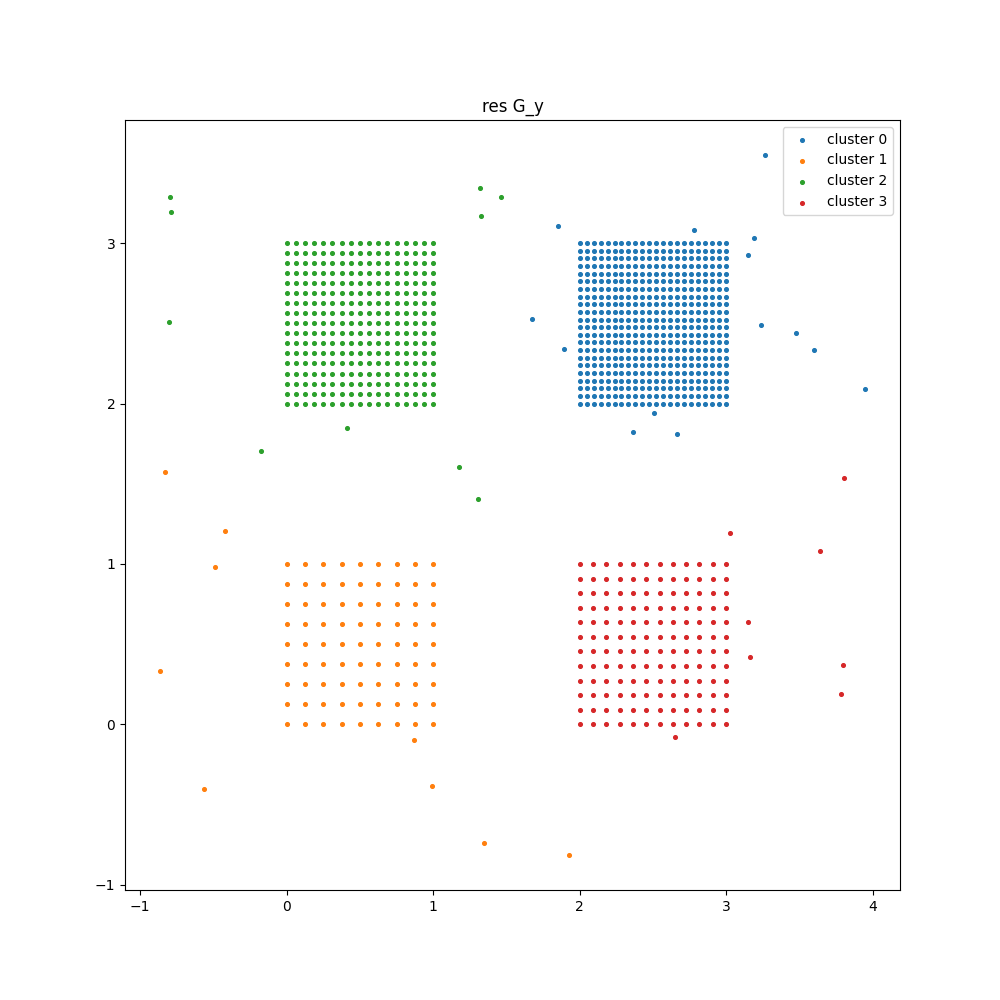 }
		\caption{}
		\label{fig:A}
	\end{subfigure}
	% \hfill % 这个命令在子图之间添加了一些水平空间
	\begin{subfigure}[b]{0.32\linewidth}
		\includegraphics[width=\textwidth]{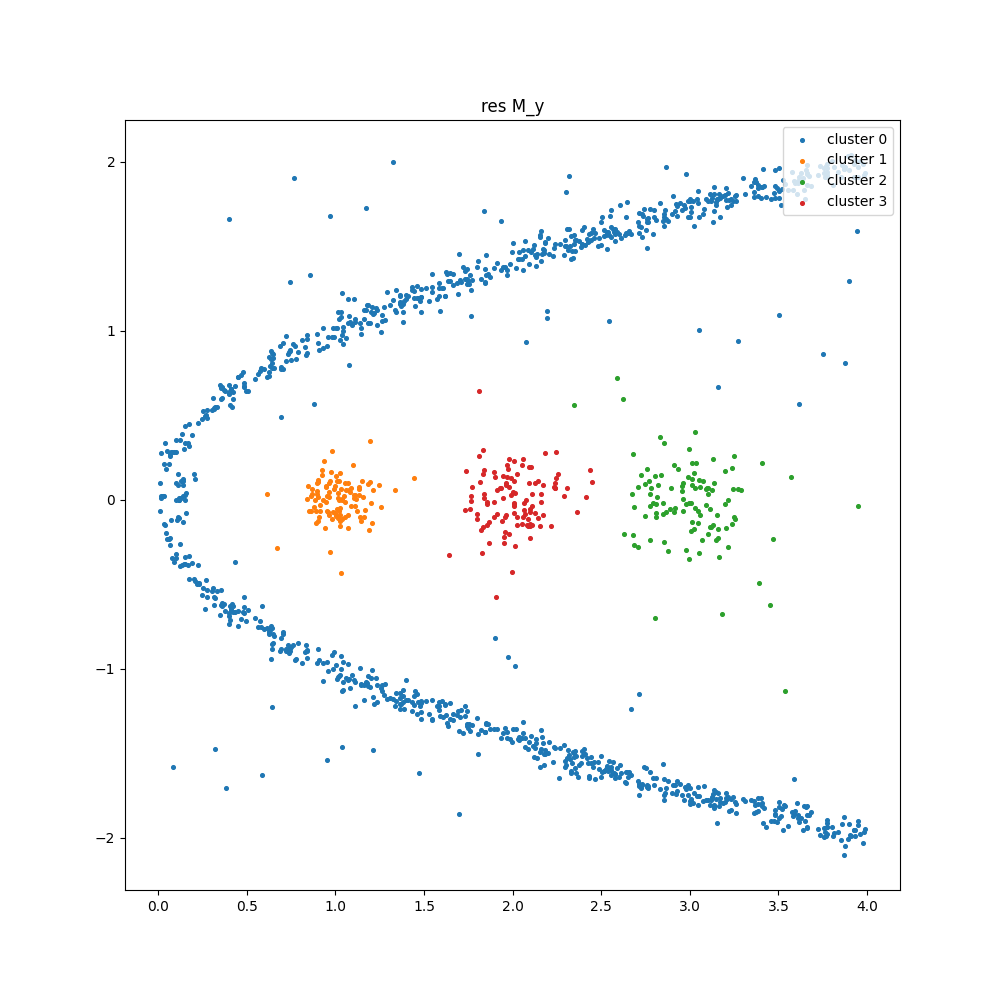}
		\caption{}
		\label{fig:A}
	\end{subfigure}
 
	\caption{The Performance of Adaptive GBCT }
	\label{t11}
\end{figure}

\autoref{t10} illustrates the principle of the adaptive GBCT. In \autoref{t10} (a), the red box indicates the distance within clusters, while in \autoref{t10} (b), the red box shows the distance between clusters, which is significantly greater than the intra-cluster distance seen in \autoref{t10} (a). \autoref{t10} (c) indicates that different clusters have been merged, leading to overfitting. \autoref{t10} (d) is a line graph showing the nearest inter-cluster granular-ball distance across the entire dataset as a function of the number of merging rounds. A notable decrease is observed in the fifth round for the first time, indicating the optimal clustering result at this point. To verify the effectiveness of the adaptive GBCT, we conducted validations on six datasets, as shown in \autoref{t11} .

\section{Conclusions and Future Work}\label{s5}

In this paper, we propose a frontier adaptive granular-ball coverage and representation clustering method, called granular-ball Clustering (GBCT). The framework aims to solve the efficiency bottleneck, robustness challenge and how to improve the clustering effect faced by traditional unsupervised learning at the fine-grained level. Firstly, GBCT achieves efficient and flexible representation and coverage of the sample space by designing a central consistency metric based on granular-balls. It not only replaces the sample point set with a smaller number of granular-balls, but also effectively resisters the interference of fine-grained noise through the coarse-grained characteristics of granular-balls, which enhances the robustness of the algorithm. In addition, GBCT shows a high degree of parameter simplicity in the clustering process, and only needs to specify a single parameter, the number of clusters K, which can flexibly meet the clustering needs of various complex data sets, demonstrating its wide applicability and strong flexibility. The experimental results show that compared with several existing basic clustering methods, the GBCT method shows significant advantages in clustering performance, does not need parameter tuning, and can deal with a wider range of complex data types.

Looking into the future, as a basic algorithm framework with wide application potential, the research on performance optimization and cross-domain integration of GBCT has important academic value and practical significance. On the one hand, we can explore the integration of advanced techniques such as kernel learning methods and non-Euclidean distance metrics into GBCT to enrich its similarity evaluation mechanism and further improve the clustering accuracy and depth. On the other hand, in view of the unique advantages of GBCT in multi-granularity coverage and representation, we can try to combine it with deep learning, traditional clustering algorithms, classification technology, image processing and other fields, and explore more possibilities in complex data processing and analysis through cross-border integration, so as to promote technical breakthroughs and application innovation in related fields.

\section*{Acknowledgements}
This work was supported in part by the National Natural Science Foundation of China under Grant Nos. 62222601, 62221005 and 62176033, Key Cooperation Project of Chongqing Municipal Education Commission under Grant No. HZ2021008, and Natural Science Foundation of Chongqing under Grant No.cstc2019jcyj-cxttX0002 and CSTB2023NSCQ-JQX0034.

\ifCLASSOPTIONcaptionsoff
  \newpage
\fi
\bibliographystyle{IEEEtran}
\bibliography{IEEEabrv,sample-base}

\begin{IEEEbiography}[{\includegraphics[width=1in,height=1.25in,clip,keepaspectratio]{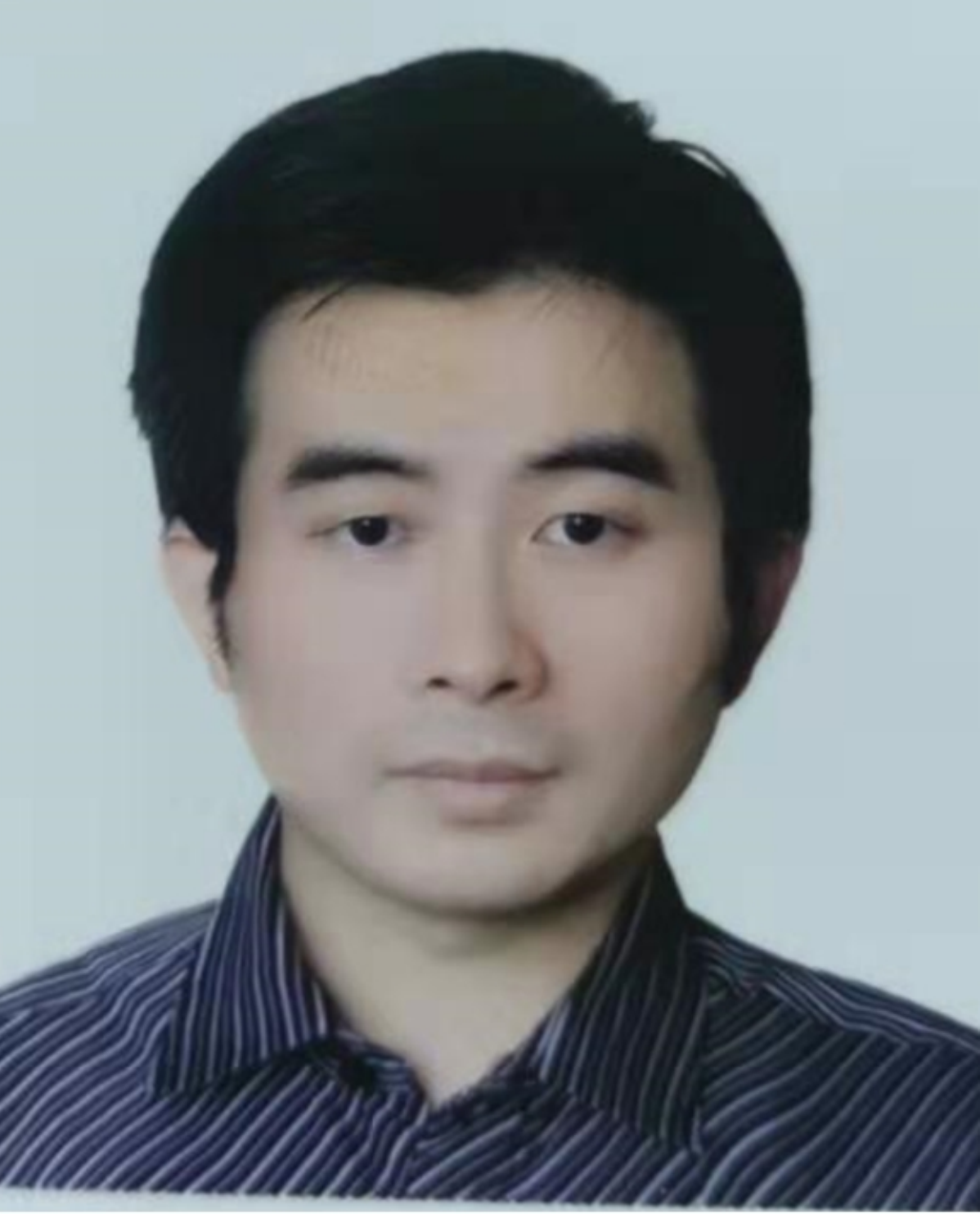}}]{Shuyin Xia} received his B.S. degree and M.S. degree in Computer science in 2008 and 2012, respectively, both from Chongqing University of Technology in China. He received his Ph.D. degree from College of Computer Science in Chongqing University in China. He is an IEEE Member. Since 2015, he has been working at the Chongqing University of Posts and Telecommunications, Chongqing, China, where he is currently an associate professor and a Ph.D. supervisor, the executive deputy director of CQUPT - Chongqing Municipal Public Security Bureau - Qihoo 360 Big Data and Network Security Joint Lab. Dr. Xia is the director of Chongqing Artificial Intelligence Association. His research results have expounded at many prestigious journals, such IEEE-TKDE and IS. His research interests include data mining, granular computing, fuzzy rough sets, classifiers and label noise detection.
\end{IEEEbiography}
\begin{IEEEbiography}[{\includegraphics[width=1in,height=1.25in,clip,keepaspectratio]{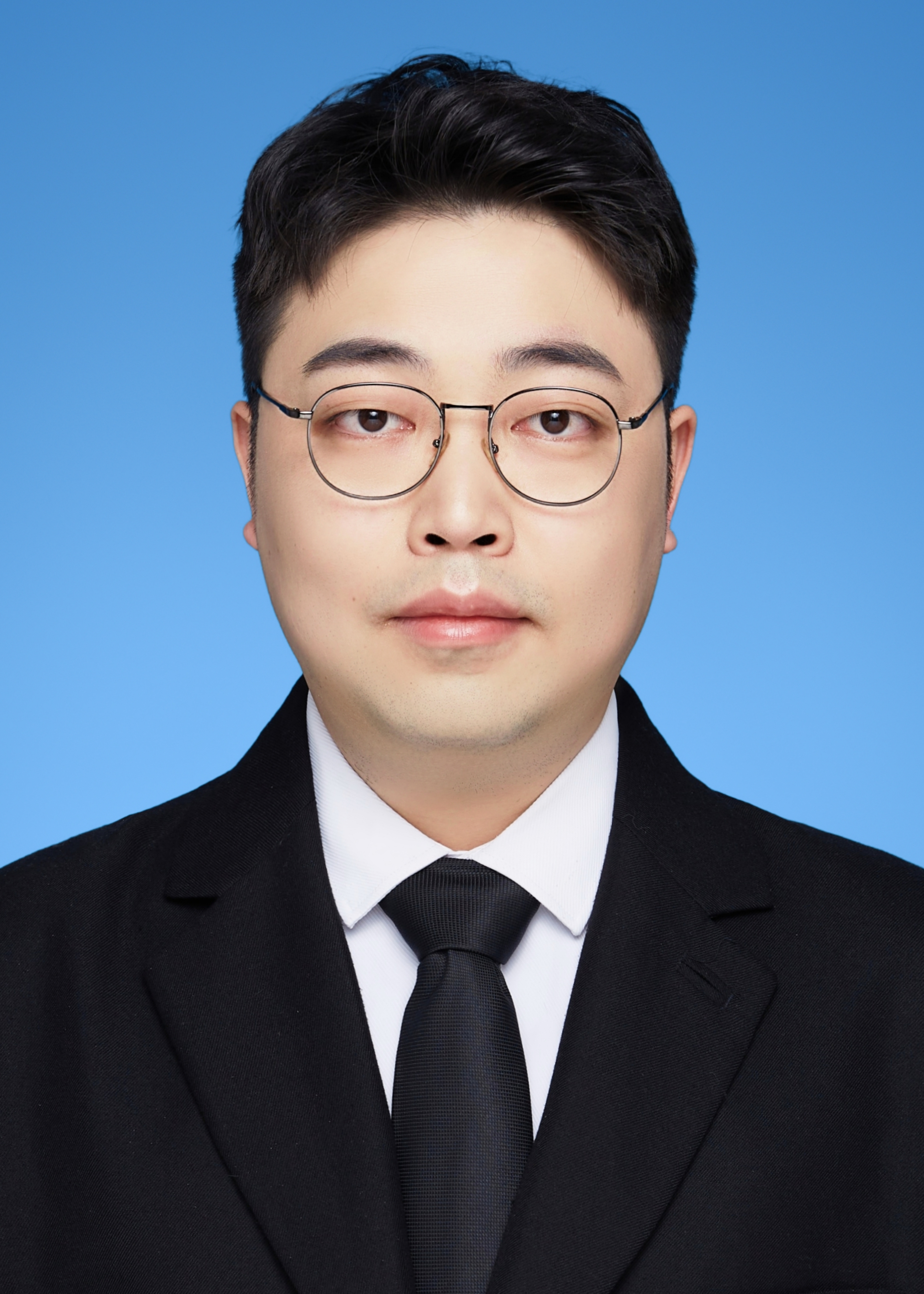}}]{Jiang Xie*}
	received the M.S. degrees and Ph.D. degrees in computer science from Chongqing University, in 2015 and 2019, respectively. He is now a lecturer in the College of Computer Science and Technology at Chongqing University of Posts and Telecommunications. His research interests include clustering analysis and data mining. 
\end{IEEEbiography}
\begin{IEEEbiography}[{\includegraphics[width=1in,height=1.25in,clip,keepaspectratio]{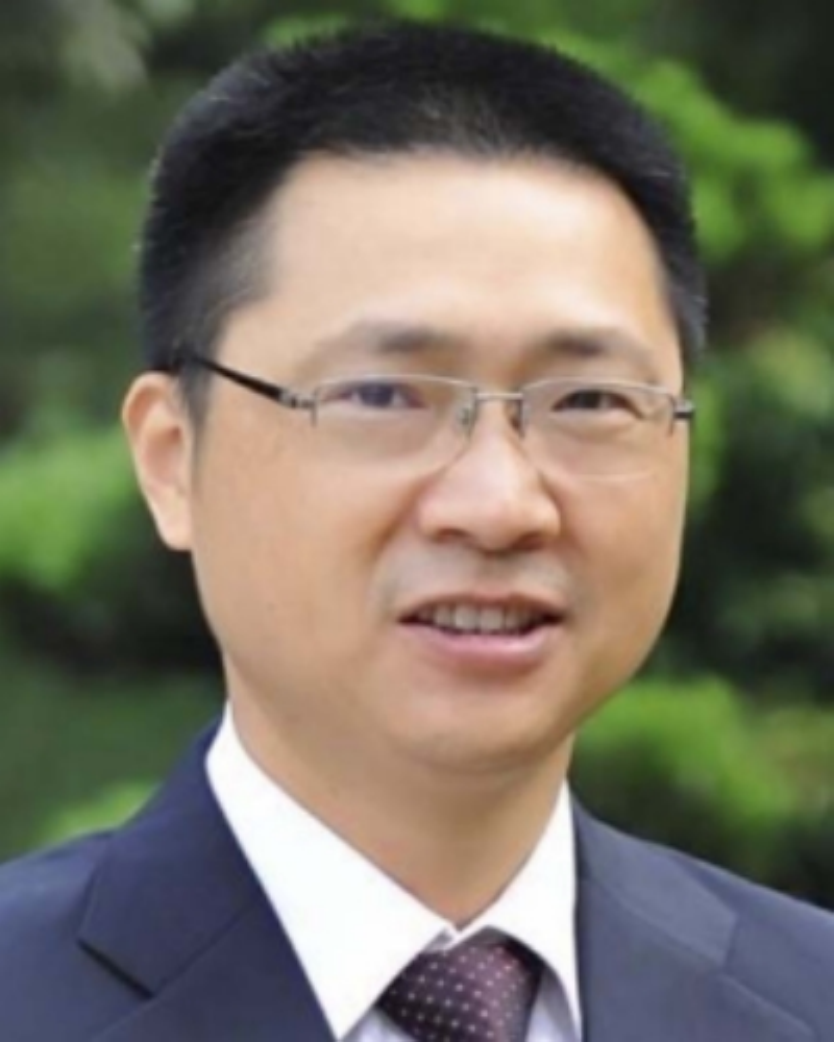}}]{Guoyin Wang(Senior Member, IEEE)} received a B.E. degree in computer software in 1992, a M.S. degree in computer software in 1994, and a Ph.D. degree in computer organization and architecture in 1996, all from Xi’an Jiaotong University in Xi’an, China. His research interests include data mining, machine learning, rough sets, granular computing, cognitive computing, and so forth. He has published over 300 papers in prestigious journals and conferences, including IEEE T-PAMI, T-KDE, T-IP, T-NNLS, and T-CYB. He has worked at the University of North T exas, USA, and the University of Regina, Canada, as a Visiting Scholar. Since 1996, he has been working at the Chongqing University of Posts and Telecommunications in Chongqing, China, where he is currently a Professor and a Ph.D. supervisor, the Director of the Chongqing Key Laboratory of Computational Intelligence, and the Vice President of the Chongqing University of Posts and Telecommunications. He is the Steering Committee Chair of the International Rough Set Society (IRSS), a Vice-President of the Chinese Association for Artificial Intelligence (CAAI), and a council member of the China Computer Federation (CCF).
\end{IEEEbiography}
\begin{IEEEbiography}
	[{\includegraphics[width=1in,height=1.25in,clip,keepaspectratio]{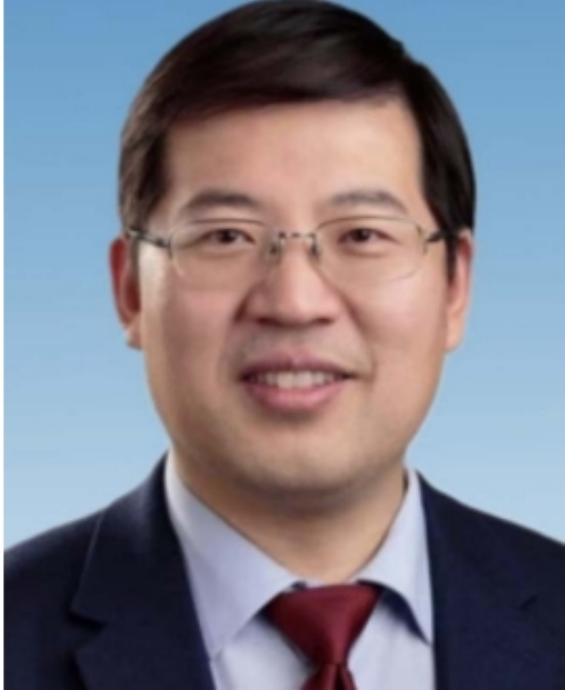}}]{Xinbo Gao} (Senior Member, IEEE) received the B.Eng., M.Sc., and Ph.D. degrees in signal and information processing from Xidian University, Xi'an, China, in 1994, 1997, and 1999, respectively. From 1997 to 1998, he was a Research Fellow with the Department of Computer Science, Shizuoka University, Shizuoka, Japan. From 2000 to 2001, he was a Post-Doctoral Research Fellow with the Department of Information Engineering, Chinese University of Hong Kong, Hong Kong. From 2001 to 2020, he has been with the School of Electronic Engineering, Xidian University. He is currently the President of the Chongqing University of Posts and Telecommunications, Chongqing, China. He has authored or co-authored six books and approximately 200 technical articles in prestigious journals and conferences, including the IEEE TRANSACTIONS ON PATTERN ANALYS IS AND MACHINE INTELLIGENCE (TPAMI), the IEEE TRANSACTIONS ON KNOWLEDGE AND DATA ENGINEERING (TIP), the IEEE TRANSACTIONS ON NEURAL NETWORKS AND LEARNING SYSTEMS (TNNLS), the IEEE TRANSACTIONS ON MEDICAL IMAGING (TMI), Conference and Workshop on Neural Information Processing Systems (NIPS), IEEE Conference on Computer Vision and Pattern Recognition (CVPR), IEEE International Conference on Computer Vision (ICCV), AAAI Conference on Artificial Intelligence (AAAI), and International Joint Conference on Artificial Intelligence (IJCAI). 
\end{IEEEbiography}

\end{document}